\documentclass{article}

\usepackage[preprint,nonatbib]{neurips_2019}
\usepackage[utf8]{inputenc} % allow utf-8 input
\usepackage[T1]{fontenc}    % use 8-bit T1 fonts
\usepackage{hyperref}       % hyperlinks
\usepackage{url}            % simple URL typesetting
\usepackage{booktabs}       % professional-quality tables
\usepackage{amsfonts}       % blackboard math symbols
\usepackage{nicefrac}       % compact symbols for 1/2, etc.
\usepackage{microtype}      % microtypography
\usepackage{graphicx}
\usepackage{color}
\usepackage{enumitem}

\title{Deep Model Transferability from Attribution Maps}
\author{
Jie Song$^{1,3}$, Yixin Chen$^1$, Xinchao Wang$^2$, Chengchao Shen$^1$, Mingli Song$^{1,3}$\\
$^1$Zhejiang University, $^2$Stevens Institute of Technology\\
$^3$Alibaba-Zhejiang University Joint Institute of Frontier Technologies\\
\texttt{\{sjie,chenyix,chengchaoshen,brooksong\}@zju.edu.cn}\\
\texttt{xinchao.wang@stevens.edu}\\
}

\begin{document}

\maketitle

\begin{abstract}
Exploring the transferability between heterogeneous tasks sheds light on their intrinsic interconnections, and consequently enables knowledge transfer from one task to another so as to reduce the training effort of the latter. In this paper, we propose an embarrassingly simple yet very efficacious approach to estimating the transferability of deep networks, especially those handling vision tasks. Unlike the seminal work of \emph{taskonomy} that relies on a large number of annotations as supervision and is thus computationally cumbersome, the proposed approach requires no human annotations and imposes no constraints on the architectures of the networks. This is achieved, specifically, via projecting deep networks into a \emph{model space}, wherein each network is treated as a point and the distances between two points are measured by deviations of their produced attribution maps. The proposed approach is several-magnitude times faster than taskonomy, and meanwhile preserves a task-wise topological structure highly similar to the one obtained by taskonomy. Code is available at \url{https://github.com/zju-vipa/TransferbilityFromAttributionMaps}.

\end{abstract}

\section{Introduction}
Deep learning has brought about unprecedented advances in many if not all the major artificial intelligence tasks, especially computer vision ones. The state-of-the-art performances, however, come at the costs of the often burdensome training process that requires an enormous number of human annotations and GPU hours, as well as the partially interpretable and thus the only intermittently predictable black-box behaviors. Understanding the intrinsic relationships between such deep-learning tasks, if any, may on the one hand elucidate the rationale of the encouraging results achieved by deep learning, and on the other hand allows for more predictable and explainable transfer learning from one task to another, so that the training effort can be significantly reduced.

The seminal work of \emph{taskonomy}~\cite{Zamir_2018_CVPR} made the pioneering attempt towards disentangling the relationships between visual tasks through a computational approach. This is accomplished by training first all the task models and then all the feasible transfers among models, in a fully supervised manner. Based on the obtained transfer performances, an affinity matrix of transferability is derived, upon which an Integer Program can be further imposed to compute the final budget-constrained task-transferability graph. Despite the intriguing results achieved, the training cost, especially that for the {combinatorial}-based transferability learning, makes taskonomy prohibitively expensive to estimate. Even for the first-order transferability estimation, the training costs grow quadratically with respect to the number of tasks involved; when adding a new task to the graph, the transferability has to be explicitly trained between the new task and all those in the task dictionary.

In this paper, we propose an embarrassingly simple yet competent approach to estimating the transferability between different tasks, with a focus on the computer vision ones. Unlike taskonomy that relies on training the task-specific models and their transferability using human annotations, in our approach we assume {no} labelled data are available, and we are given only the pre-trained deep networks, which can be nowadays found effortless online. Moreover, we do not impose constraints on the architectures of the deep networks, such as networks handling different tasks sharing the same architectures.

At the heart of our approach is to {project} pre-trained deep networks into a common space, termed \emph{model space}. The model space accepts networks of heterogeneous architectures and handling different tasks, and transforms each network into a point. The distance between two points in the model space is then taken to be the measure of their relatedness and the consequent transferability. Such construction of the model space enables prompt model insertion or deletion, as updating the transferability graph boils down to computing nearest neighbors in the model space, which is therefore much lighter than taskonomy that requires the pair-wise re-training for each newly added task.

The projection to the model space is attained by feeding unlabelled images, which can be obtained handily online, into a network and then computing the corresponding \emph{attribution maps}. An attribution map signals pixels in the input image highly relevant to the downstream tasks or hidden representations, and therefore highlights the ``attention'' of a network over a specific task. In other words, the model space can be thought as a space defined on top of attribution maps, where the affinity between points or networks is evaluated using the distance between their produced attribution maps, which again, requires no supervision and can be computed really fast.

The intuition behind adopting attribution maps for network-affinity estimation is rather straightforward:
models focusing on similar regions of input images are expected to produce correlated representations,
and thus potentially give rise to favorable transfer-learning results.
This assumption is inspired by the work of~\cite{Zagoruyko2017PayingMA},
which utilizes the attention of a teacher model to guide the learning of a student and produces encouraging results. Despite its very simple nature, the proposed approach yields truly promising results: it leads to a speedup factor of several magnitudes of times and meanwhile maintains a highly similar transferability topology, as compared to taskonomy.
In addition, experiments on vision tasks beyond those involved in taskonomy also produce intuitively plausible results, validating the proposed approach and providing us with insights on their transferability.

Our contribution is therefore a lightweight and effective approach towards estimating transferability between deep visual models, achieved via projecting each model into a common space and approximating their affinity using attribution maps. It requires no human annotations and is readily applicable to pre-trained networks specializing in various tasks and of heterogeneous architectures. Running at a speed several magnitudes faster than taskonomy and producing competitively similar results, the proposed model may serve as a competent transferability estimator and an effectual substitute for taskonomy, especially when human annotations are unavailable, when the model library is large in size, or when frequent model insertion or update takes place.

\section{Related Work}
We briefly review here some topics that are most related to the proposed work, including model reusing, transfer learning, and attribution methods for deep models.
\paragraph{Model Reusing.}
Reusing pre-trained models has been an active research topic in recent years.
Hinton \textit{et al.}~\cite{hinton2015distilling} firstly propose the concept of ``knowledge distillation'' where the trained cumbersome teacher models are reused to produce soft labels for training a lightweight student model. Following their teacher-student scheme, some more advanced methods~\cite{romero2014fitnets,Zagoruyko2017PayingMA,furlanello2018born,lan2018knowledge} are proposed to fully exploit the knowledge encoded in the trained teacher model. However, in these works all the teachers and the student are trained for the same task. To reuse models of different tasks, Rusu \textit{et al.}~\cite{rusu2016progressive} propose the progressive neural net to extract useful features from multiple teachers for a new task. Parisotto \textit{et al.}~\cite{parisotto2015actor} propose ``Actor-Mimic'' to use the guidance from several expert teachers of distinct tasks.
However, none of these works explore the relatedness among different tasks. In this paper, by explicitly modeling the model transferability, we provide an effective method to pick a  trained model most beneficial for solving the target task.
\paragraph{Transfer Learning.}
Another way of reusing trained models is to transfer the trained model to another task by reusing the features extracted from certain layers. Razavian \textit{et al.}~\cite{Razavian:2014:CFO:2679599.2679731} demonstrated that features extracted from deep neural networks could be used as generic image representations to tackle the diverse range of visual tasks. Yosinski \textit{et al.}~\cite{NIPS2014_5347} investigated the transferability of deep features extracted from every layer of a deep neural network. Azizpour \textit{et al.}~\cite{7328311} investigated several factors affecting the transferability of deep features. Recently, the effects of pre-training datasets for transfer learning are studied~\cite{47104,DBLP:journals/corr/abs-1811-08883,DBLP:journals/corr/HuhAE16,5995347,Richter_2017_ICCV}.  None of these works, however, explicitly quantify the relatedness among different tasks or trained models to provide a principled way for model selection. Zamir \textit{et al.}~\cite{Zamir_2018_CVPR} proposed a fully computational approach,  known as taskonomy, to address this challenging problem. However, taskonomy requires labeled data and is computationally expensive, which limited its applications in large-scale real-world problems. Recently, Dwivedi and Roig~\cite{Dwivedi_2019_CVPR} proposed to use representation similarity analysis to approximate the task taxonomy.
In this paper, we introduce a model space for modeling task transferability and propose to measure the transferability via attribution maps, which, unlike taskonomy, requires no human annotations and works directly on pre-trained models. We believe our method is a good complement to existing works.

\paragraph{Attribution Methods for Deep Models.}
Attribution refers to assigning importance scores to the inputs for a specified output. Existing attribution methods can be mainly divided into two groups, including perturbation-~\cite{Zeiler2014VisualizingAU,Zhou2015PredictingEO,Zintgraf2017VisualizingDN} and gradient-based~\cite{Simonyan2013DeepIC,Bach2015OnPE,Shrikumar2016NotJA,Sundararajan2017AxiomaticAF,Shrikumar2017LearningIF,Montavon2017ExplainingNC,Ancona2018TowardsBU} methods. Perturbation-based methods compute the attribution of an input feature by making perturbations,~\textit{e.g.}, removing, masking or altering, to individual inputs or neurons and observe the impact on later neurons. However, such methods are computationally inefficient as each perturbation requires a separate forward propagation through the network. Gradient-based methods, on the other hand, estimate the attributions for all input features in one or few forward and backward passes throughout the network, which renders them generally more efficient. Simonyan \textit{et al.}~\cite{Simonyan2013DeepIC} construct attributions by taking the absolute value of the partial derivative of the target output with respect to the input features. Later, Layer-wise Relevance Propagation ($\epsilon$-LRP)~\cite{Bach2015OnPE}, gradient*input~\cite{Shrikumar2016NotJA}, integrated gradients~\cite{Sundararajan2017AxiomaticAF} and deepLIFT~\cite{Shrikumar2017LearningIF} are proposed to aid understanding the information flow of deep neural networks.  In this paper, we directly adopt some of these off-the-shelf methods to produce the attribution maps. Devising more suitable attribution method for our problem is left to future work.

\begin{figure}
  \centering
  \includegraphics[scale=0.43]{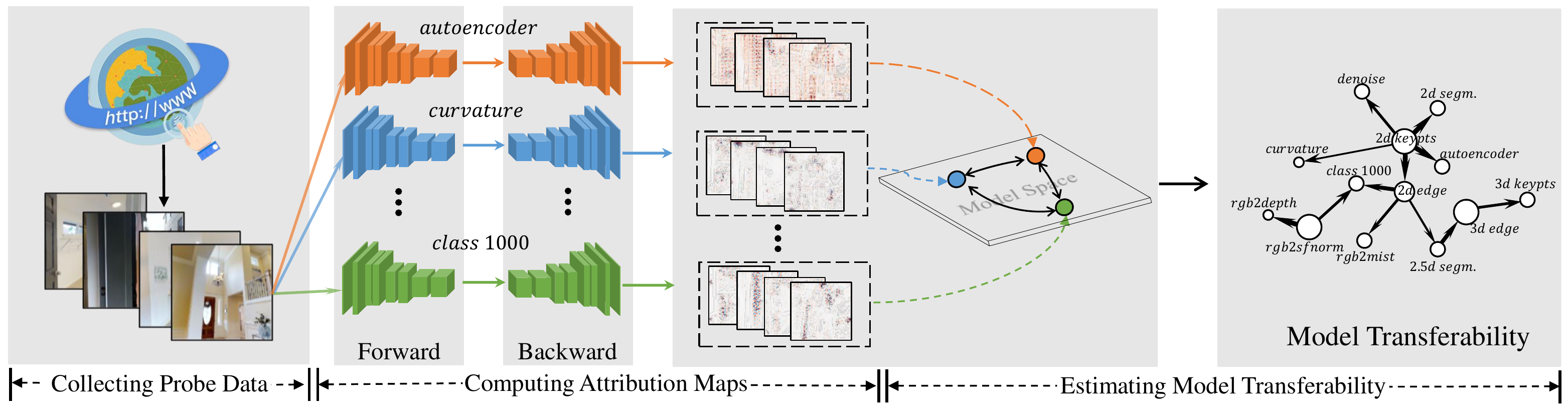}\\
  \caption{An illustrative diagram of the workflow of the proposed method. It mainly consists of three steps: collecting probe data, computing attribution maps, and estimating model transferability.} \label{fig:framework}
\end{figure}

\section{Estimating Model Transferability from Attribution Maps}
We provide in this section the details of the proposed transferability estimator.
We start by giving the problem setup and an overview of the method, followed by describing its three steps, and finally show the efficiency analysis.

\subsection{Problem Setup}
Assume we are given a set of pre-trained deep models
$\mathcal{M}=\{m_{1}, m_{2}, ..., m_{N}\}$, where $N$ is the total number of models involved.
No constraints are imposed on the architectures of these models. We use  $t_{i}$ to denote the task handled by model $m_{i}$, and use
$\mathcal{T}=\{t_{1}, t_{2}, ..., t_{N}\}$ to denote the \emph{task dictionary}, \textit{i.e.}, the set of all the
tasks involved in $\mathcal{M}$.
Furthermore, we assume that no labeled annotations are available.
Our goal is to efficiently quantify the  transferability between
different tasks in $\mathcal{T}$, so that given a target task,
we can read out from the learned transferability matrix
the source task that potentially yields the highest transfer performance.

\subsection{Overview}
The core idea of our method is to embed the pre-trained deep models into the model space, wherein models are represented by points and model transferability is measured by the distance between corresponding points. To this end, we utilize the attribution maps to construct such a model space. The assumption is that related models should produce similar attribution maps for the same input image. The workflow of our method consists of three steps, as shown in Figure~\ref{fig:framework}. First, we collect an unlabeled \textit{probe dataset}, which will be used to construct the model space, from a randomly selected data distribution. Second, for each trained model, we adopt off-the-shelf attribution methods to compute the attribution maps of all images in the constructed probe dataset. Finally, for each model, all its attribution maps are collectively viewed as a single point in the model space, based on which the model transferability is estimated. In what follows, we provide details for each of the three steps.

\subsection{Key Steps}
\paragraph{Step 1: Building the Probe Dataset.}
As deep models handling different tasks or even the same one may be of
heterogeneous architectures or trained on data from various domains,
it is non-trivial to measure their transferability directly from their outputs or intermediate features. To bypass this problem, we feed the same input images to these models and measure the model transferability by the similarity of their response to the same stimuli.
We term the set of all such input images \textit{probe data},
which is shared by all the tasks involved.

Intuitively,  the probe dataset should be designed not only large in size but also rich
in diversity, as models in $\mathcal{M}$ may be trained on various domains for different tasks.
However, experiments show that the proposed method works
surprisingly well even when the probe data are collected in a single domain and of moderately small size ($\sim 1,000$ images).
The produced transferability relationship is highly similar to the one derived by taskonomy. This property renders the proposed method attractive as little effort is required for collecting the probe data. More details can be found in Section~\ref{section:task_relation}.

\paragraph{Step 2: Computing Attribution Maps.}
Let us denote the collected probe data by $\mathcal{X}=\{X_{1}, X_{2}, ...,X_{N_p}\}$, $X_{i}=\left[x^i_{1}, x^i_{2}, ..., x^i_{WHC}\right] \in \mathbb{R}^{WHC}$, where $W$, $H$ and $C$ respectively denote the width, the height and the channels of the input images, and $N_p$ is the size of the probe data. Note that for brevity the maps are symbolized in vectorization form here. For model $m_i$, it takes an input $\tilde{X}= T_{i}(X)\in \mathbb{R}^{W_{i}H_{i}C_{i}}$ and produces a hidden representation $R=[r_{1}, r_{2}, ..., r_{D}]$. Here, $T_{i}$ serves as a preprocessing function that transforms the images in probe data for model $m_{i}$, as we allow different models to take images of different sizes as input, and
$D$ is the dimension of the representation. For each model $m_{i}$ in $\mathcal{M}$, our goal in this step is to produce an attribution map $A_{j}^{i}=[a_{j1}^{i}, a_{j2}^{i}, ...]\in \mathbb{R}^{WHC}$ for each image $X_{j}$ in the probe data $\mathcal{X}$.

In fact, an attribution map $A^{i,k}_j$ can be computed for each unit $r_{k}$ in $R$. However, as we consider the transferability of $R$, we average the attribution maps of all $r$ in $R$ as the overall attribution map of $R$. Formally, we have $A_{j}^{i}=\frac{1}{D}\sum_{k=1}^{D}A_{j}^{i,k}$.
Specifically, here we adopt three off-the-shelf attribution methods to produce the attribution maps: saliency map~\cite{Simonyan2013DeepIC}, gradient * input~\cite{Shrikumar2016NotJA}, and $\epsilon$-LRP~\cite{Bach2015OnPE}.
Saliency map computes attributions by taking the absolute value of the partial derivative of the target output with respect to the input. Gradient * input refers to a first-order taylor approximation of how the output would change if the input was set to zero. $\epsilon$-LRP, on the other hand, computes the attributions by redistributing the prediction score (output) layer by layer until the input layer is reached.
For all the three attribution methods, the overall attribution map $A_{j}^{i}$ can be computed through one single forward-and-backward propagation~\cite{Ancona2018TowardsBU} in Tensorflow.
The formulations of the three attribution maps are summarized in Table~\ref{table:attributions}. More details can be found from~\cite{Simonyan2013DeepIC,Shrikumar2016NotJA,Bach2015OnPE,Ancona2018TowardsBU}.
\begin{table}[htp]
\caption{Mathematical formulations of saliency map~\cite{Simonyan2013DeepIC}, gradient * input~\cite{Shrikumar2016NotJA} and $\epsilon$-LRP~\cite{Bach2015OnPE}. Note that the superscript $g$ denotes a novel definition of partial derivative~\cite{Ancona2018TowardsBU}.}
\begin{center}
\begin{tabular}{c|c|c|c}
\toprule
\textbf{Method}		&\textbf{Saliency Map}~\cite{Simonyan2013DeepIC}		&\textbf{Gradient * Input}~\cite{Shrikumar2016NotJA}	&\textbf{$\epsilon$-LRP}~\cite{Bach2015OnPE,Ancona2018TowardsBU}\\
\midrule
$\tilde{A}^{i,k}_j$	&$\left[\left|\frac{\partial{r}_{k}}{\partial{x^j_{d}}}\right|\right]_{d=1}^{W_iH_iC_i}$	&$\left[x^j_{d}\cdot\frac{\partial{r}_{k}}{\partial{x^j_{d}}}\right]_{d=1}^{W_iH_iC_i}$	&$\left[x^j_{d}\cdot\frac{\partial^{g}{r}_{k}}{\partial{x^j_{d}}}\right]_{d=1}^{W_iH_iC_i}$, $g=\frac{f(z)}{z}$	\\
\bottomrule
\end{tabular}
\end{center}
\label{table:attributions}
\end{table}

For model $m_{i}$, the produced attribution map $\tilde{A^{i}}$ is of the same size as the input $\tilde X$, \textit{i.e.}, $\tilde{A^{i}} \in \mathbb{R}^{W_{i}H_{i}C_{i}}$.  We do the inverse of $T$ to transform the attribution maps back to the same size as the images in the probe data: $A^{i}=T^{-1}(\tilde{A^{i}}), A^{i}\in\mathbb{R}^{WHC}$. As attribution maps of all models are transformed into the same size, the transferability can be computed based on these maps.

\paragraph{Step 3: Estimating Model Transferability.}
Once step~2 is completed, we have $N_p$ attribution maps $\mathcal{A}^{i} = \{A_{1}^{i}, A_{2}^{i}, ..., A_{N_p}^{i}\}$ for each model $m_{i}$, where $A_{j}^{i}$ denotes the attribution map of $j$-th image $X_{j}$ in $\mathcal{X}$.
The model $m_{i}$ can be viewed as a sample in the model space $\mathbb{R}^{NWHC}$, formed by concatenating all the attribution maps. The distance of two models are taken to be
\begin{equation}
d(m_{i}, m_{j}) = \frac{N_p}{\sum_{k=1}^{N_p}{cos\_sim(A_k^i, A_k^j)}},
\end{equation}
where $cos\_sim(A_k^i, A_k^j)=\frac{A_{k}^{i}\cdot A_{k}^{j}}{\left\|A_k^i\right\|\cdot\left\|A_k^j\right\|}$.
The model transferability map, which measures the pairwise
transferability relationships, can then be derived based on these
distances.
The model transferability, as shown by taskonomy~\cite{Zamir_2018_CVPR}, is inherently asymmetric. In other words, if model $m_i$ ranks first in being transferred to task $t_j$ among all the models (except $m_j$) in $\mathcal{M}$, $m_j$ does not necessarily rank first in being transferred to task $t_i$. Yet, the proposed model space is symmetric in distance, as we have $d(m_{i}, m_{j}) = d(m_{j}, m_{i})$. We argue that the symmetric property of the distance in the model space makes little negative effect on the transferability relationships, as the task transferability rankings of the source tasks are computed by relative comparison of distances. Experiments demonstrate that with the symmetric model space, the proposed method is able to effectively approximate the asymmetric transferability relationships produced by taskonomy.
\subsection{Efficiency Analysis}
\label{section:efficiency}
Here we make a rough comparison between the efficiency of the proposed approach and that of taskonomy.
As we assume task-specific trained models are available, we compare the computation cost of our method with that of only the transfer modeling  in taskonomy. For taskonomy, let us assume the transfer model is trained for $E$ epochs on the training data of size $N$, then for a task dictionary of size $T$, the computation cost can be approximately denoted as $ENT(T-1)$-times forward-and-backward propagation\footnote{Here for simplicity, we ignore the computation-cost difference caused by the model architectures.}. For our method working on the probe dataset, however, only one time of forward-and-backward propagation is required. The overall computation cost for building the model space in our method is about $TM$-times forward-and-backward propagation, where $M$ is the size of the probe dataset and usually $M\ll N$. The proposed method is thus about $\frac{EN(T-1)}{M}$-times more efficient than taskonomy.
This also means the speedup over taskonomy will be even more significant, if more tasks are involved and hence $T$ enlarges.

In our experiments, the proposed method takes about $20$ GPU hours to compute the pairwise transferability relationships on one Quadro P5000 card for $20$ pre-trained taskonomy models, while taskonomy takes thousands of GPU hours on the cloud\footnote{As the hardware configurations are not clear here, we list the GPU hours only for perceptual comparison.} for the same number of tasks.

\section{Experiments}

\subsection{Experimental Settings}
\paragraph{Pre-trained Models.}
Two groups of trained models are adopted to validate the proposed method. In the first group, we adopt 20 trained models of single-image tasks released by taskonomy~\cite{Zamir_2018_CVPR}, of which the task relatedness has been constructed and also released. It is used as the oracle to evaluate the proposed method. Note that all these models adopt an encoder-decoder architecture, where the encoder is used to extract representations and the decoder makes task predictions. For these models, the attribution maps are computed with respect to the output of the encoder.

To further validate the proposed method, we construct a second group of trained models which are collected online. We have managed to obtain 18 trained models in this group: two VGGs~\cite{Simonyan2015VeryDC} (VGG16, VGG19), three ResNets~\cite{He2016DeepRL} (ResNet50, ResNet101, ResNet152), two Inceptions (Inception V3~\cite{Szegedy2016RethinkingTI}, Inception ResNet V2~\cite{szegedy2017inception}), three MobileNets~\cite{Howard2017MobileNetsEC} (MobileNet, 0.5 MobileNet, 0.25 MobileNet), four Inpaintings~\cite{yu2018generative} (ImageNet, CelebA, CelebA-HQ, Places), FCRN~\cite{laina2016deeper}, FCN~\cite{Long2015FullyCN}, PRN~\cite{Feng2018Joint3F} and Tiny Face Detector~\cite{Hu2017FindingTF}. All these models are also viewed in an encoder-decoder architecture. The sub-model which produces the most compact features is viewed as the encoder and the remainder as the decoder. Similar to taskonomy models, the attribution maps are computed with respect to the output of the encoder.
More details of these models can be found in the supplementary material.
\paragraph{Probe Datasets.}
We build three datasets, taskonomy data~\cite{Zamir_2018_CVPR}, indoor scene~\cite{Quattoni2009RecognizingIS}, and COCO~\cite{Lin2014MicrosoftCC}, as the probe data to evaluate our method.
The domain difference between taskonomy data and COCO is much larger than that between taskonomy data and indoor scene. For all the three datasets, we randomly select about $1,000$ images to construct the probe datasets. More details of the three probe datasets are provided in the supplementary material. In Section~\ref{section:task_relation}, we demonstrate the performances of the proposed method evaluated on these three probe datasets.

\subsection{Experiments on  Models in Taskonomy}
\subsubsection{Visualization of Attribution Maps}
\begin{figure}
  \centering
  \includegraphics[scale=0.42]{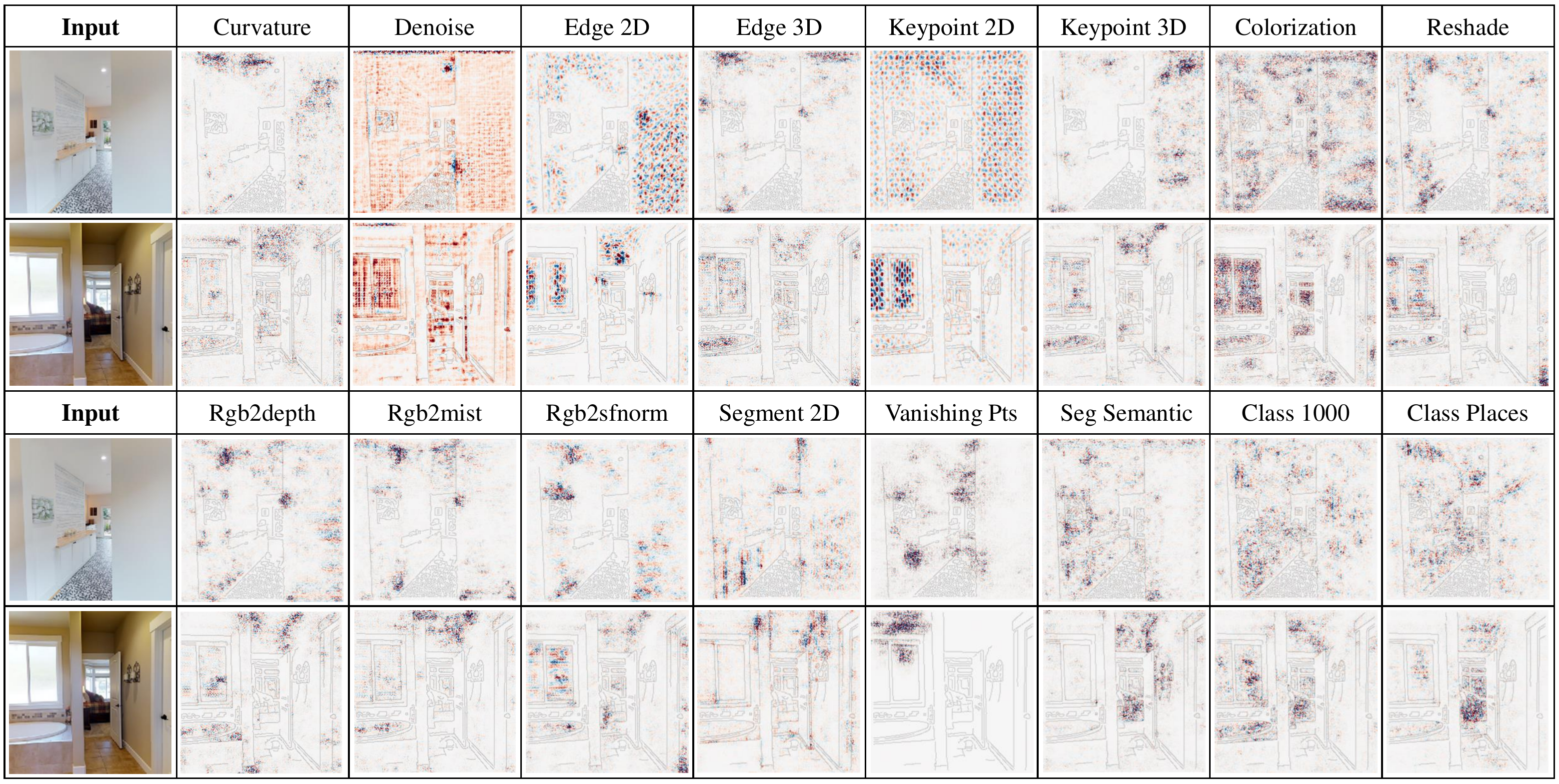}
  \caption{Visualization of attribution maps produced using $\epsilon$-LRP on taskonomy models. Some tasks produce visually similar attribution maps, such as Rgb2depth and Rgb2mnist.}
  \label{fig:vis_attribution}
\end{figure}
We first visualize the attribution maps produced by various trained models for the same input images. Two examples are given in Figure~\ref{fig:vis_attribution}. Attribution maps are produced by $\epsilon$-LRP on taskonomy data. From the two examples, we can see that some tasks produce visually similar attribution maps. For example, $\langle$Rgb2depth, Rgb2mist$\rangle$\footnote{Here we use $\langle\rangle$ to denote a cluster of tasks, of which the attribution maps are highly similar.}, $\langle$Class 1000, Class Places$\rangle$ and $\langle$Denoise, Keypoint 2D$\rangle$. In each cluster, trained models pay their ``attentions'' to the similar regions, thus the ``knowledge'' they learned are intuitively highly correlated~(as seen in Section~\ref{section:cca}) and can be transferred to each other~(as seen in Section~\ref{section:task_relation}). Two examples may produce conclusions where the constructed model transferability deviates from the underlying model relatedness. However, such deviation is alleviated by aggregating the results of more examples drawn from the data distribution. For more visualization examples, please see the supplementary material.

\subsubsection{Rationality of the Assumption}
\label{section:cca}
Here we adopt Singular Vector Canonical Correlation Analysis~(SVCCA)~\cite{raghu2017svcca} to validate the rationality underlying our assumption: if tasks produce similar attribution maps, the representations extracted from corresponding models should be highly correlated, thus they are expected to yield favorable transfer-learning performance to each other. In SVCCA, each neuron is represented by an \textit{activation vector}: {its set of response to a set of inputs} and hence the layer can be represented by the subspace spanned by the activation vectors of all the neurons in this layer. SVCCA first adopts Singular Value Decomposition~(SVD) of each subspace to obtain new subspaces that comprise the most important directions of the original subspaces, and then uses Canonical Correlation Analysis~(CCA) to compute a series of correlation coefficients between the new subspaces. The overall correlation is measured by the average of these correlation coefficients.

\begin{figure}
  \centering
  % Requires \usepackage{graphicx} 25, 49
  \includegraphics[scale=0.15]{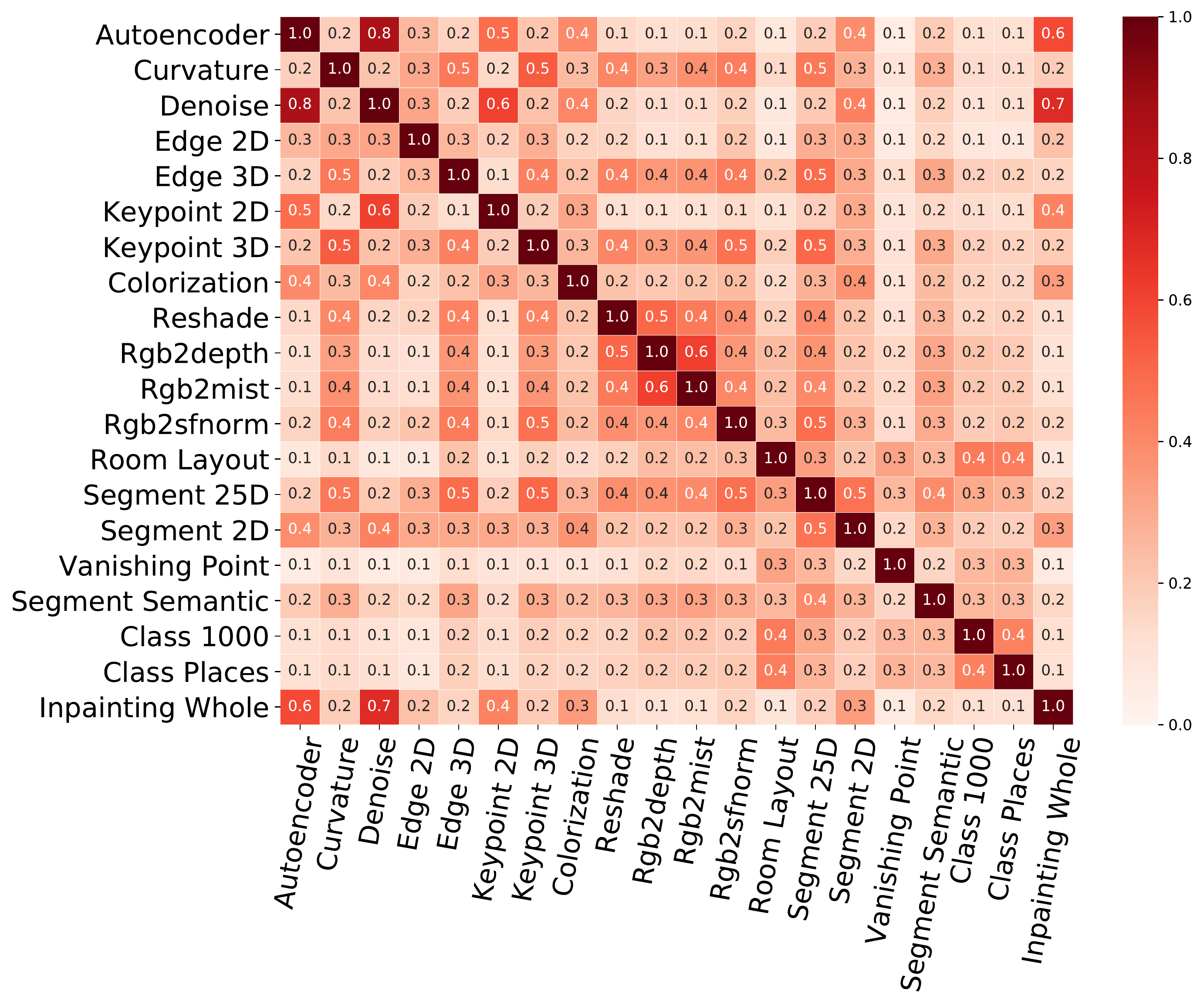}
  \includegraphics[scale=0.15]{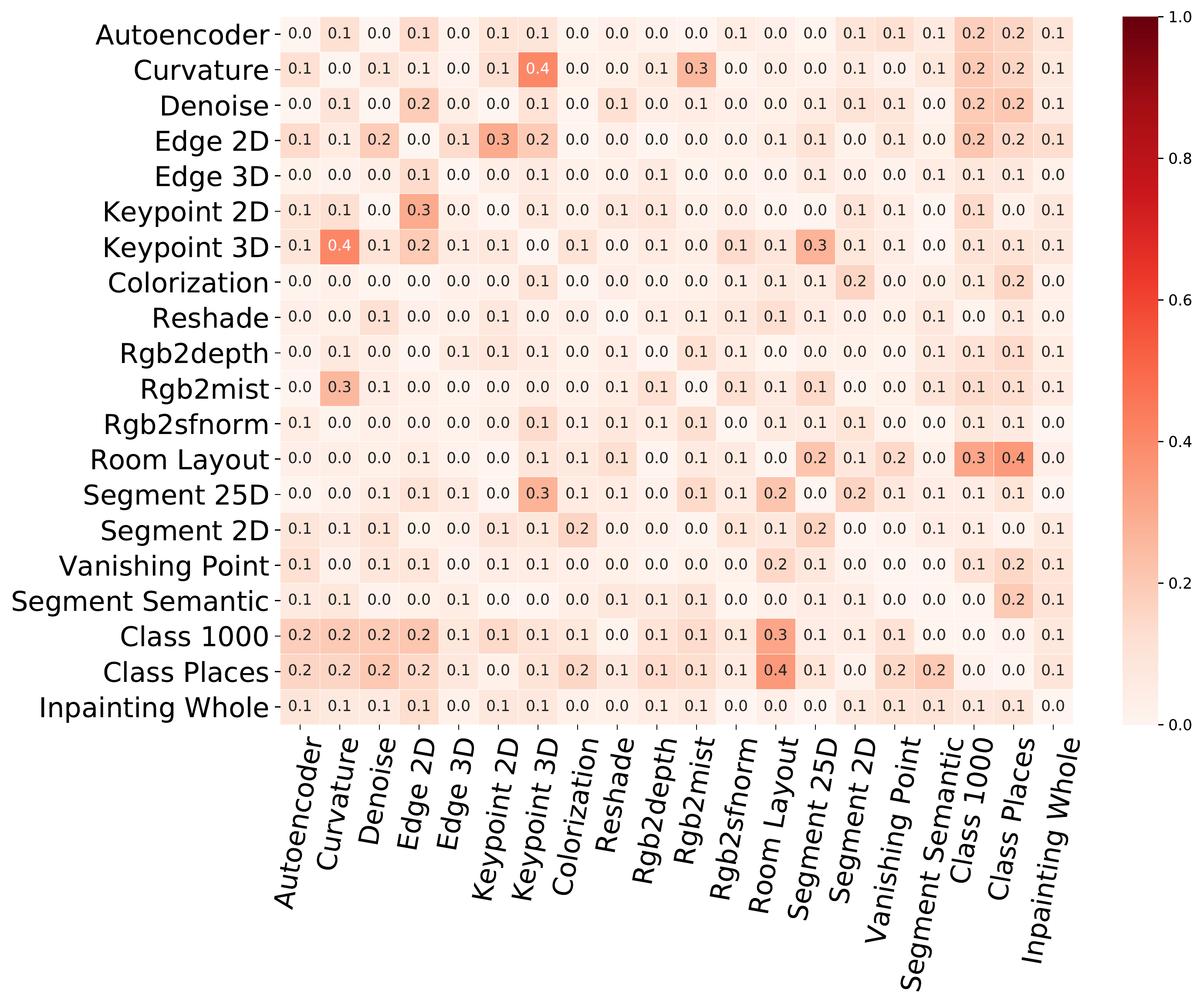}
  \includegraphics[scale=0.32]{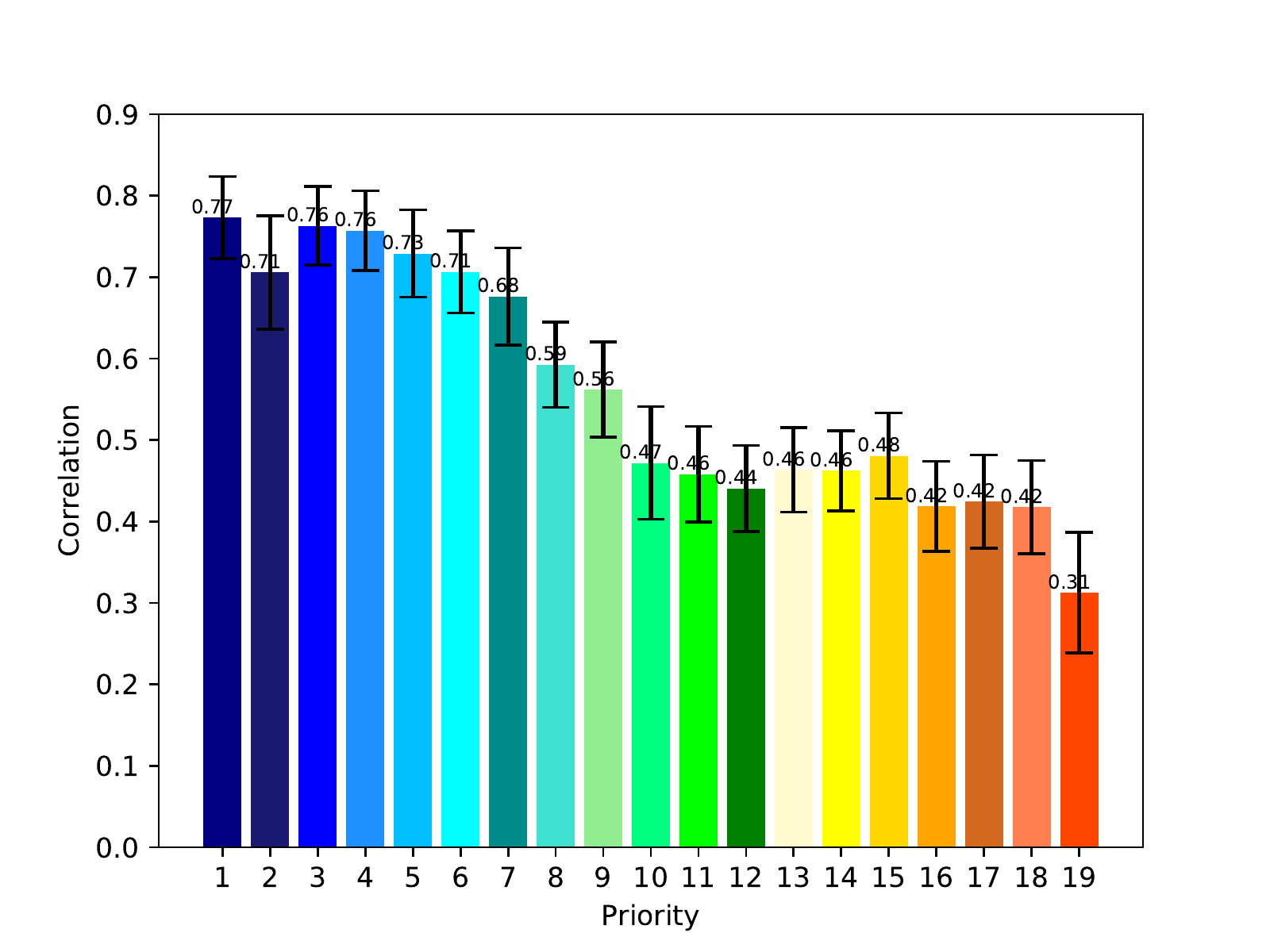}
  \caption{\textbf{Left}: visualization of the correlation matrix from SVCCA. \textbf{Middle}: the difference between correlation matrix from SVCCA and the transferability matrix derived from attribution maps. Both of them are normalized for better visualization. \textbf{Right}: the Correlation-Priority Curve (CPC).}\label{fig:heatmap_cca}
\end{figure}

Experimental results on taskonomy data with $\epsilon$-LRP are shown in Figure~\ref{fig:heatmap_cca}. In the left, the correlation matrix over the pre-trained taskonomy models is visualized. In the middle, we plot the difference between the correlation matrix and the model transferability matrix derived from attribution maps in the proposed method. It can be seen that the values in the difference matrix are in general small,  implying that the correlation matrix is highly similar to the model transferability matrix. To further quantify the overall similarity between these two matrices, we compute their Pearson correlation ($\rho_p=0.939$) and Spearman correlation ($\rho_s=0.660$). All these results show that the similarity of attribution maps is a good indicator of the correlation between representations.

In addition, we can see that some tasks, like Edge3d and Colorization, tend to be more correlated to other tasks, as the colors of the corresponding row or column are darker than those of others, while some other tasks are not, like Vanishing Point. In taskonomy, the priorities\footnote{The priority of a task $i$ refers to the average ranking when transferred to other tasks: $p_i=\frac{1}{N}\sum_j^N{r^i_j}$, where $r^i_j$ denotes the ranking of task $i$ when transferred to task $j$. A smaller value of $p$ denotes a higher priority.} of Edge3d, Colorization and Vanishing Point are $5.4$, $5.8$ and $14.2$, respectively. It indicates that more correlated representations tend to be more suitable for transferring learning to each other. To make this clearer, we depict the Correlation-Priority Curve (CPC) in the right of Figure~\ref{fig:heatmap_cca}. In this figure, for each priority $p$ shown on the abscissa, the correlation shown on the ordinate is computed as $correlation(p) = \frac{1}{N}\sum_{i\ne j} \mathbb{I}(r_j^i=p)\rho_{i,j}$, where $\mathbb{I}$ is the indicator function and $\rho_{i, j}$ is the correlation between representations extracted from two models $m_i$ and $m_j$. It can be seen that as the priority becomes lower, the average correlation becomes weaker. All these results verify the rationality underlying the assumption.

\subsubsection{Deep Model Transferability}
\label{section:task_relation}
\begin{figure}
  \centering
  \includegraphics[scale=0.12]{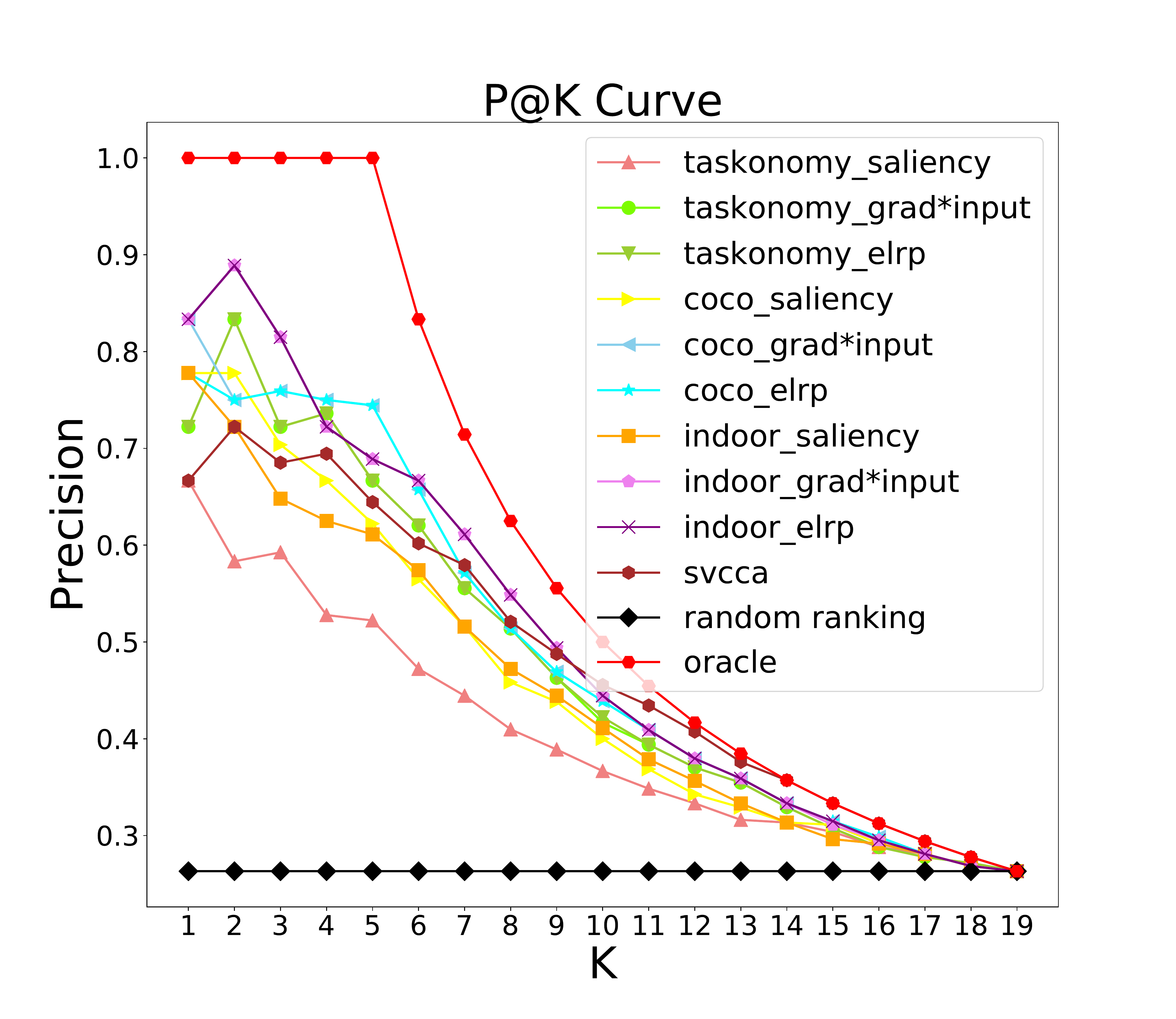}
  \includegraphics[scale=0.12]{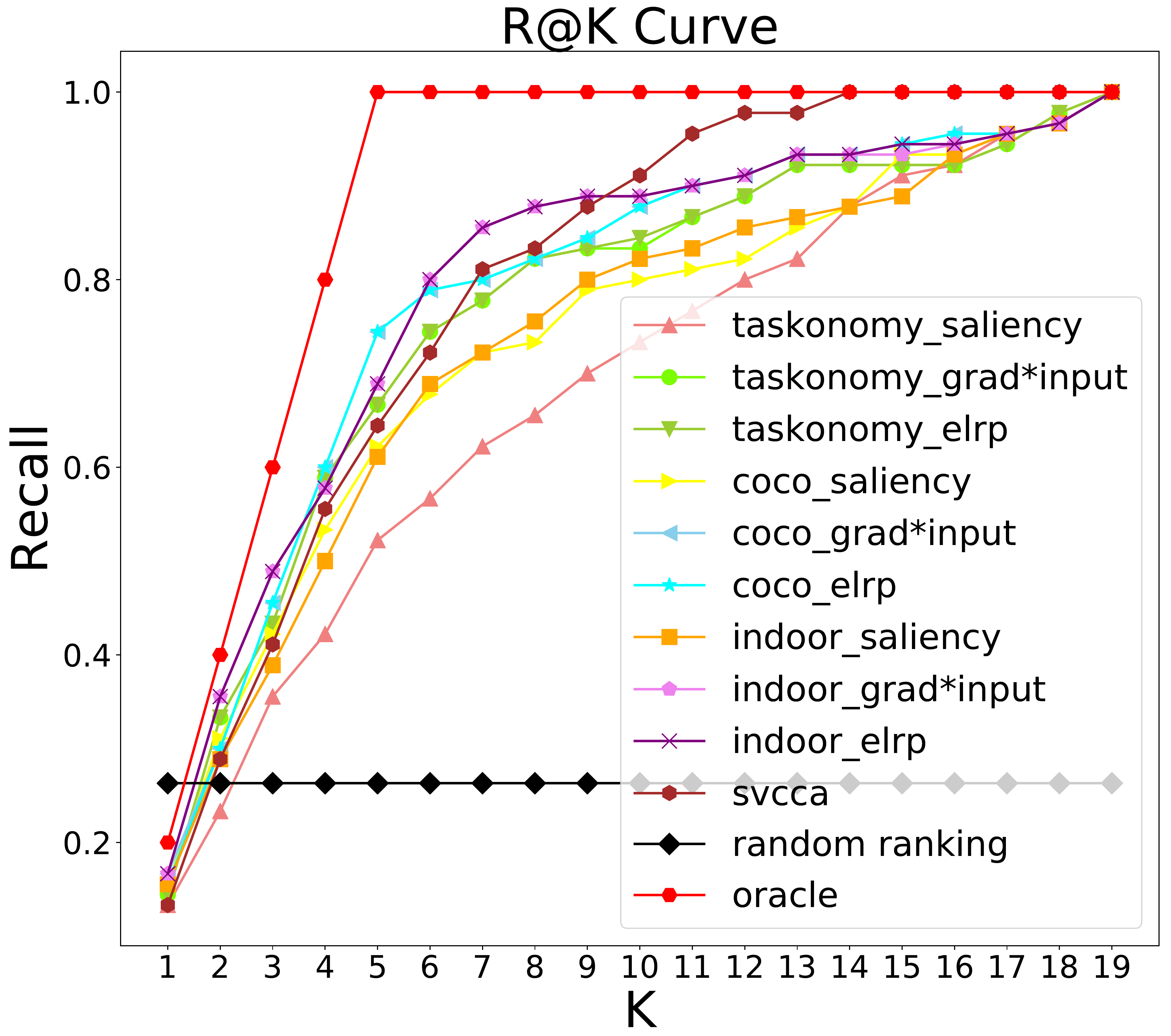}
  \includegraphics[scale=0.26]{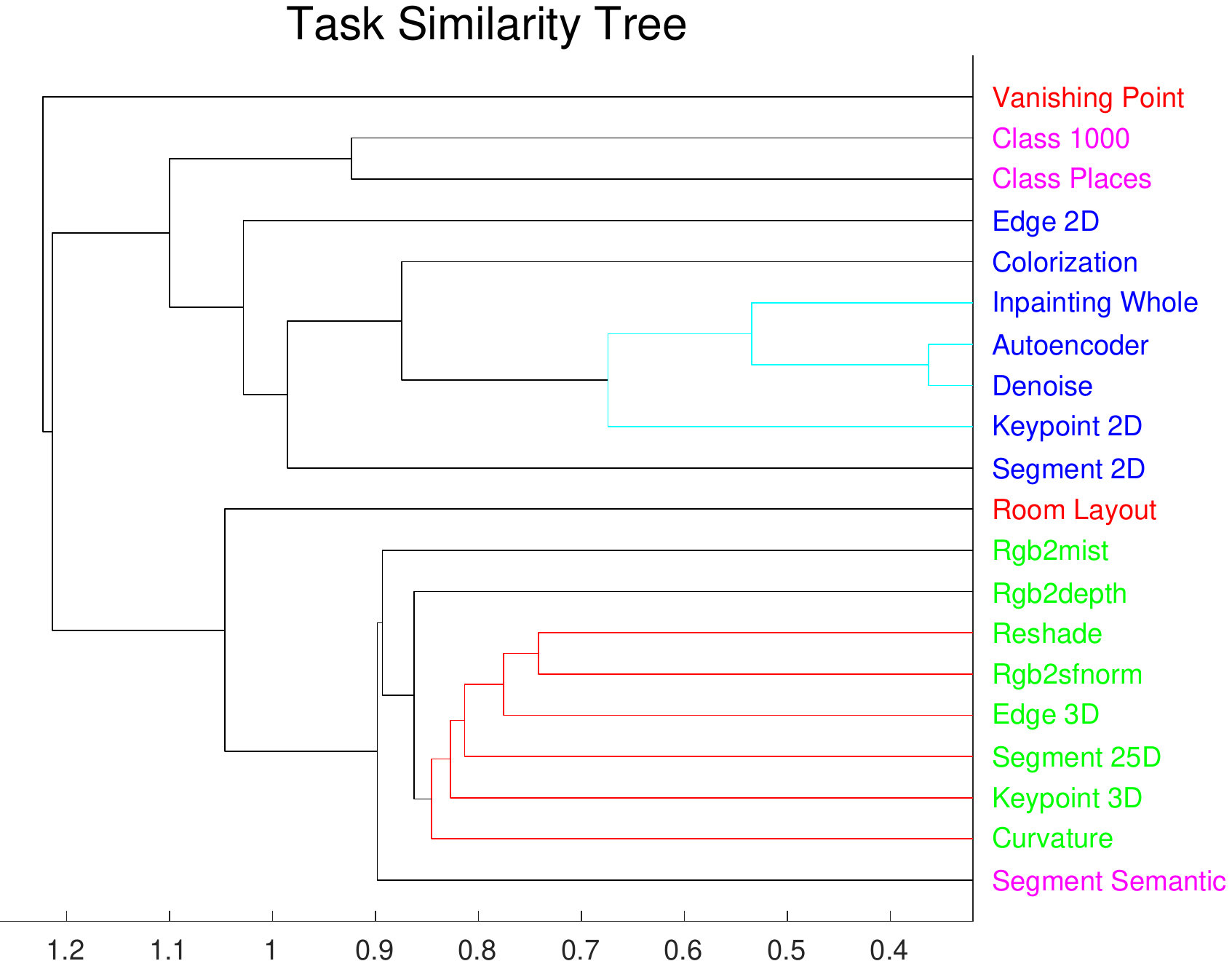}
  \caption{From left to right: P@K curve, R@K curve and task similarity tree constructed by $\epsilon$-LRP. Results of SVCCA are produced using validation data from taskonomy.}
  %``taskonomy\_saliency'' means that we adopt the data from taskonomy as the probe data and saliency for attribution.}
  \label{fig:task_relationships}
\end{figure}

We adopt two evaluation metrics, P@K and R@K\footnote{P: precision, R: recall, @K: only the top-K results are examined.}, which are widely used in the information retrieval field, to compare the model transferability constructed from our method with that from tasknomy. Each target task is viewed as a query, and its top-$5$ source tasks that produce the best transferring performances in taskonomy are regarded as relevant to the query. To better understand the results, we introduce one baseline using random ranking, and the oracle, the ideal method which always produces the perfect results. Additionally, we also evaluate SVCCA for computing the model transferability relationships.
The experimental results are depicted in Figure~\ref{fig:task_relationships}. Based on the results, we can make the following conclusions.
\begin{itemize}[leftmargin=15pt]
\item The topology structure of the model transferability derived from the proposed method is similar to that of oracle. For example, when only top-$3$ predictions are examined, the precision can be about $85\%$ on COCO with $\epsilon$-LRP. To see this clearer, we also depict the task similarity tree constructed by agglomerative hierarchical clustering in Figure~\ref{fig:task_relationships}. This tree is again highly similar to that of taskonomy where {\color{green}3D}, {\color{blue}2D}, {\color{red}geometric}, and {\color{magenta}semantic} tasks cluster together.

\item $\epsilon$-LRP and gradient* input generally produce better performance than saliency.
    This phenomenon  can be in part explained by the fact that
    saliency generates attributions entirely based on gradients
    that denote the direction for optimization.
    However, the gradients are not able to fully reflect the relevance between the inputs and the outputs of the deep model, thus leading to inferior results.
    It also implies the attribution method can affect the performance of our method. Devising better attribution methods may further promote the accuracy of our method, which is left as future work.

\item The proposed method works quite well on the probe data from different domains, such as indoor scene and COCO. It implies that the proposed method is robust to different choices of the probe data to some degree, which makes the data collection effortless. Furthermore, it can be seen that the probe data from indoor scene and COCO surprisingly better predict the taskonomy transferability than the probe data from taskonomy data. We conjecture that more complex textures disentangle the attributions better, thus the probe data from COCO and indoor scene which are generally more complex in texture yield superior results to taskonomy as probe data. However, more research is necessary to discover if the explanation holds in general.

\item SVCCA also works well in estimating the transferability of taskonomy models. However, the proposed method yields superior or comparable performance to SVCCA when using gradient * input and $\epsilon$-LRP for attribution. What's more, as the proposed method measures transferability by computing distances, it is several times more efficient than SVCCA, especially when the hidden representation is large in dimension or a new task is added into a large task dictionary.
\end{itemize}
With all these observations and the fact that the proposed method is significantly more efficient than taskonomy, the proposed method is indeed
an effectual substitute for taskonomy, especially when human annotations are unavailable, when the model library is large in size, or when frequent model insertion and update takes place.

\subsection{Experiments on Models beyond Taskonomy}
\label{section:public}
\begin{figure}[t]
\centering
  \includegraphics[scale=0.11]{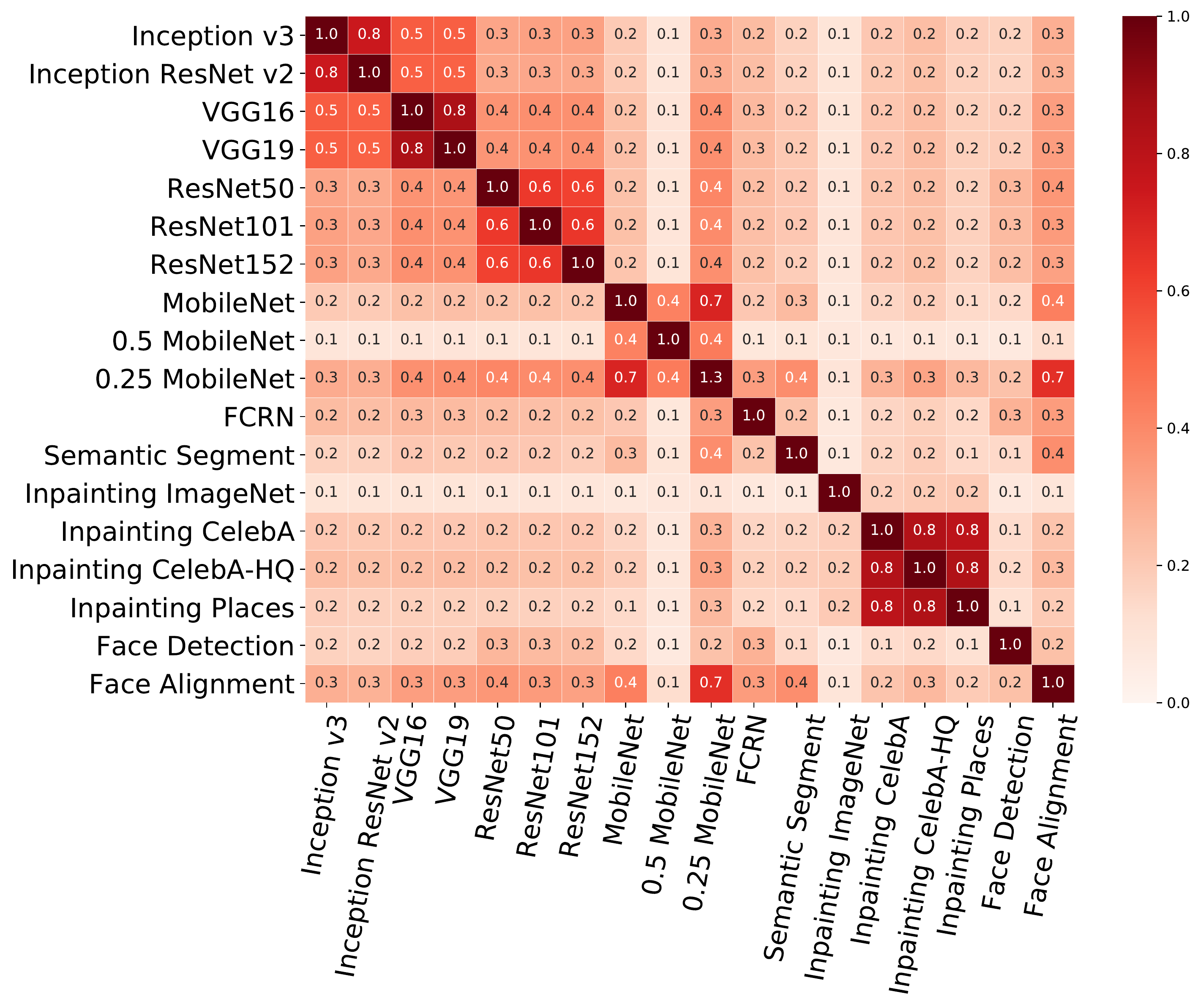}
  \includegraphics[scale=0.11]{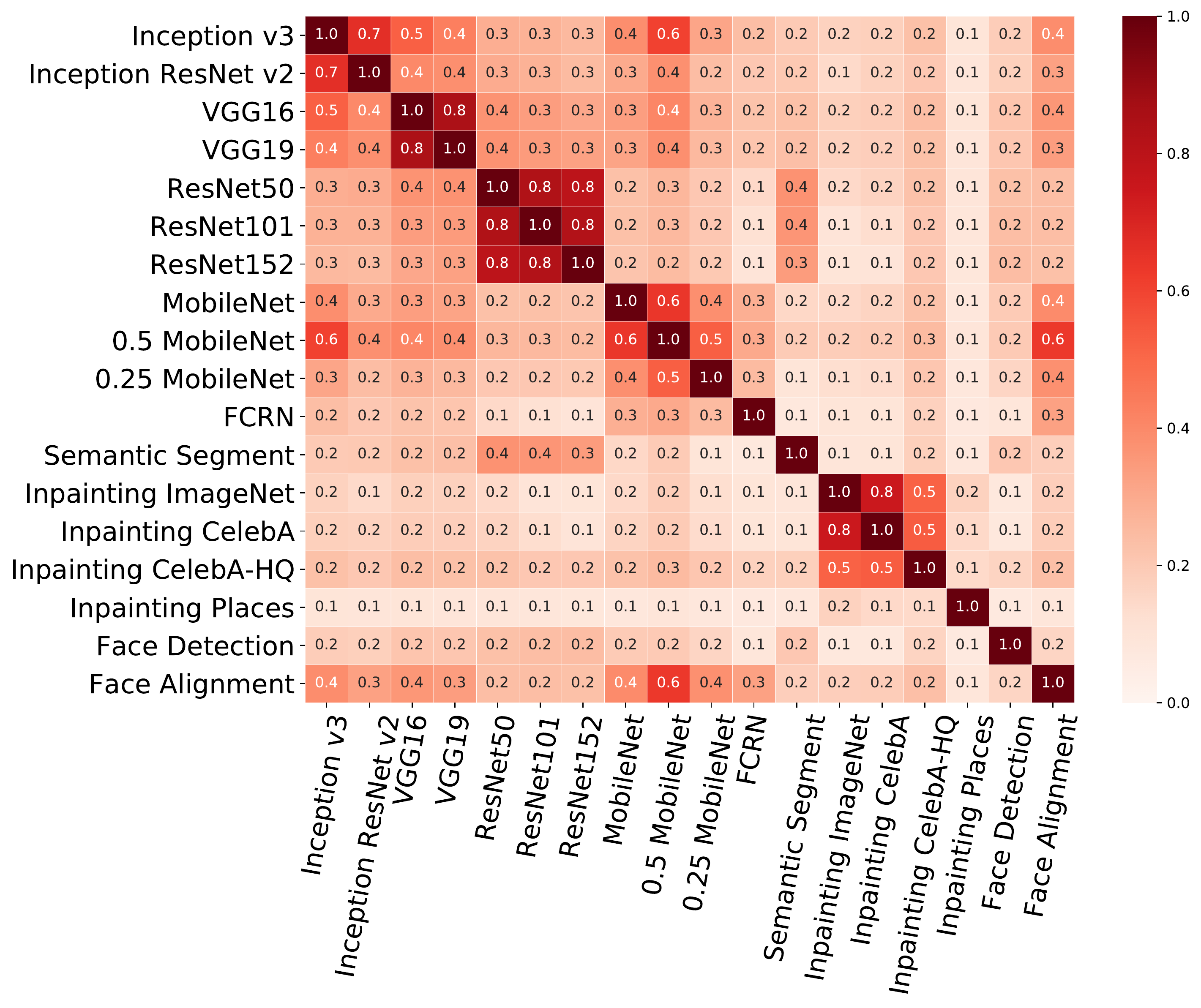}
  \includegraphics[scale=0.20]{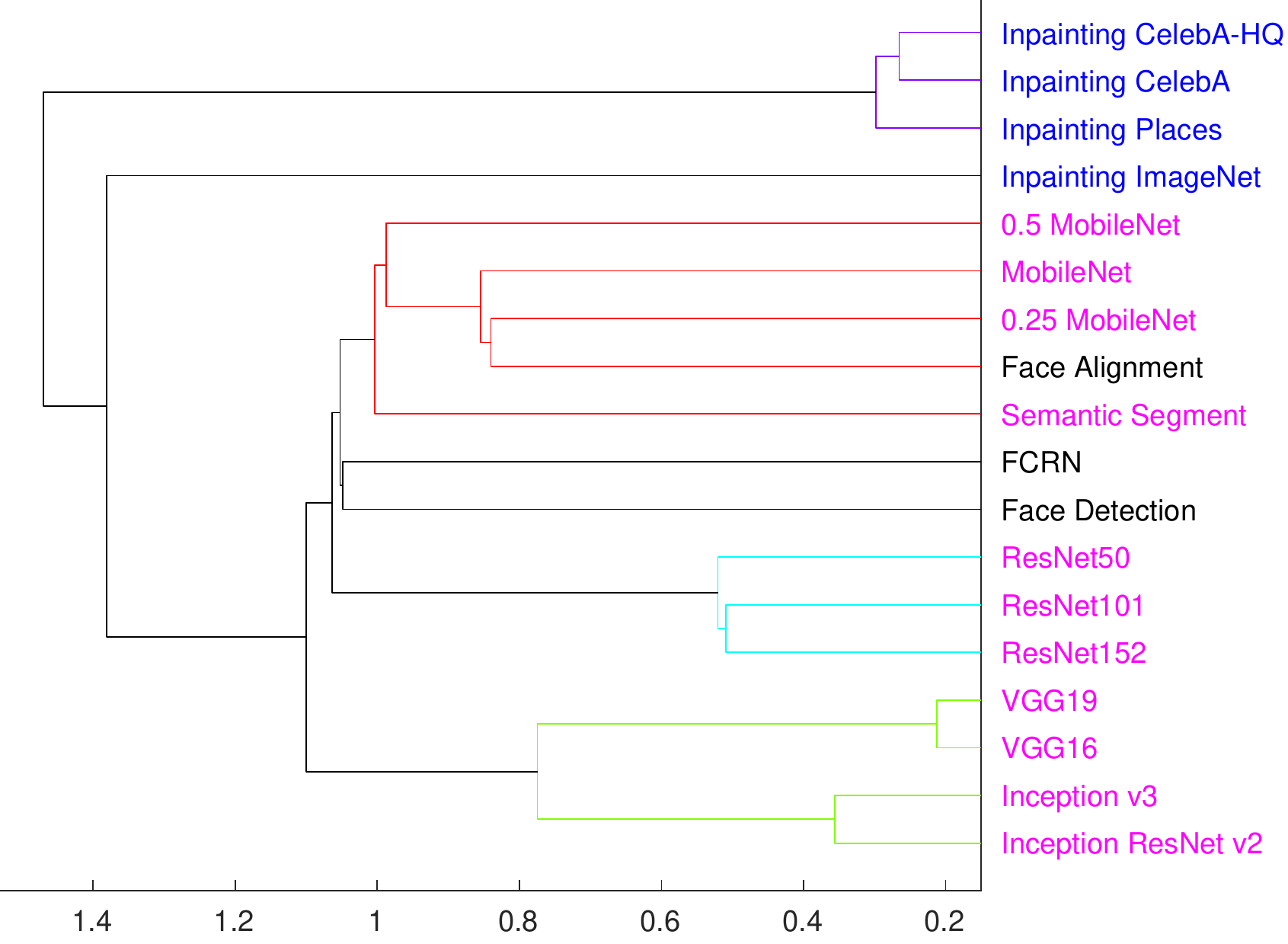}
  \includegraphics[scale=0.20]{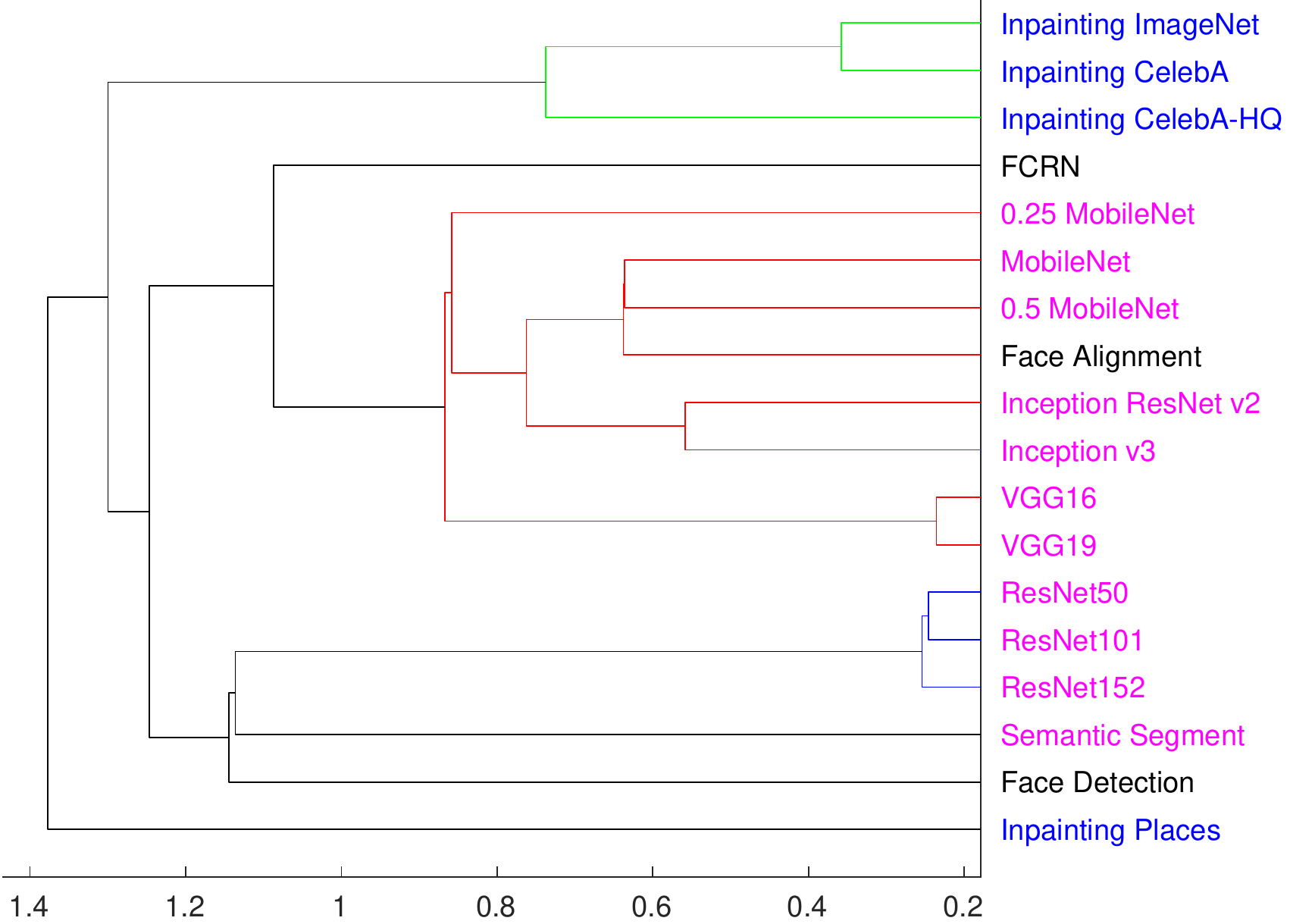}\\
  \caption{Results on collected models beyond taskonomy. From left to right: affinity matrix from SVCCA, affinity matrix from attribution maps, task similarity tree from SVCCA, and task similarity tree from attribution maps.}\label{fig:online_tree}
\end{figure}

To give a more comprehensive view of the proposed method, we also conduct experiments on the online collected pre-trained models beyond taskonomy.
Results are shown in Figure~\ref{fig:online_tree}. The left two subfigures show the correlation matrix from SVCCA and the model transferability matrix produced by our method. The right two subfigures depict the task similarity trees produced by SVCCA and the proposed method. The {\color{magenta}classification} and {\color{blue}inpainting} models are listed in different colors. We have the following observations.
\begin{itemize}[leftmargin=15pt]
\item The proposed method produces an affinity matrix and a task similarity tree alike those derived from SVCCA, although the collected models are heterogeneous in architectures, tasks, and input size. These results further validate that models producing similar attribution maps also produce highly correlated representations.
\item All the ImageNet-trained classification models, despite their different architectures, tend to cluster together. Furthermore, the same-task trained models with the similar architectures tend to be more related than with dissimilar architectures. For example, ResNet50 is more related to ResNet101 and ResNet152 than VGG, MobileNet and Inception models, indicating that the architecture plays a certain role in regularization for solving the tasks.
\item The inpainting models, albeit trained on data from different data domain, also tend to cluster together. It implies that different models of the same task, albeit trained on data from different data domain, tend to play similar role in transfer learning. However, more research is necessary to verify if this observation holds in general.
\end{itemize}
We also merge the two groups into one to further evaluate the proposed method, of which the results are provided in the supplementary material,
providing us with more insights on model transferability.

\section{Conclusion}
We introduce in this paper an embarrassingly simple yet efficacious approach towards estimating the transferability between deep models, without using any human annotation. Specifically, we project the pre-trained models of interest into a model space, wherein each model is treated as a point and the distance between two points are used to approximate their transferability. The projection to the model space is achieved by computing the attribution maps from the unlabelled probe dataset. The proposed approach imposes no constraints on the architectures on the models, and turns out to be robust to the selection of the probe data. Despite the lightweight construction, it yields a transferability map highly similar to the one obtained by taskonomy yet runs at a speed several magnitudes faster, and therefore may serve as a compact and express transferability estimation, especially when no annotations are available, the model library is large in size, or frequent model insertion or update takes place.

\subsubsection*{Acknowledgments}
This work is supported by  National Key Research and Development Program (2016YFB1200203), National Natural Science Foundation of China (61572428,U1509206), Key Research and Development Program of Zhejiang Province (2018C01004), and the Program of International Science and Technology Cooperation (2013DFG12840).

{\small
\bibliographystyle{plain}
\bibliography{neurips_2019}
}

\end{document}

\maketitle
\section{Pre-trained Models}
We adopt two groups of pre-trained models to validate the proposed method. The first group consists of the pre-trained taskonomy models~\cite{Zamir_2018_CVPR}. All the taskonomy models adopt an encoder-decoder architecture. For all tasks, the encoders are identical and implemented by the modified ResNet-50. The architectures of the decoders depend on the tasks. We adopt 20 trained models of single-image tasks released by taskonomy\footnote{\url{https://github.com/StanfordVL/taskonomy/tree/master/taskbank}}: Autoencoder, Curvature, Denoise, Edge 2D, Edge 3D, keypoint 2D, Keypoint 3D, Colorization, Reshade, Rgb2depth, Rgb2minst, Rgb2sfnorm, RoomLayout, Segment 25D, Segment 2D, VanishingPoint, SegmentSemantic, Class 1000, Class Places and Inpainting Whole. Please refer to~\cite{Zamir_2018_CVPR} for more details about these models.
\begin{table}[h]
\scriptsize
%\footnotesize
\caption{Details of the collected models. ``Layer Name'' refers to the layers w.r.t. which the attribution maps are computed. ``Pre-Logits'' indicates the layer previous to the one which produces logits.}
\label{table:collected_models}
\centering
\begin{threeparttable}
\begin{tabular}{cccccc }
\toprule
\textbf{Model}		&\textbf{Task}	&\textbf{Training Data} &\textbf{Input Size} &\textbf{Pre-trained} &\textbf{Layer Name}\\
\midrule
Inception V3~\cite{Szegedy2016RethinkingTI}  & Classification & ILSVRC2012~\cite{ILSVRC15} &$299\times299\times3$   &\XSolid &Pre-Logits\\
Inception-ResNet V2~\cite{szegedy2017inception}  & Classification & ILSVRC2012~\cite{ILSVRC15} &$299\times299\times3$   &\XSolid &Pre-Logits\\
VGG 16~\cite{Simonyan2015VeryDC}     & Classification & ILSVRC2012~\cite{ILSVRC15}   &$224\times224\times3$    &\XSolid &Pre-Logits\\
VGG 19~\cite{Simonyan2015VeryDC}     & Classification & ILSVRC2012~\cite{ILSVRC15}   &$224\times224\times3$    &\XSolid &Pre-Logits\\
ResNet V1 50~\cite{He2016DeepRL}     & Classification & ILSVRC2012~\cite{ILSVRC15}   &$224\times224\times3$ &\XSolid &Pre-Logits\\
ResNet V1 101~\cite{He2016DeepRL}     & Classification & ILSVRC2012~\cite{ILSVRC15}   &$224\times224\times3$ &\XSolid &Pre-Logits\\
ResNet V1 152~\cite{He2016DeepRL}     & Classification & ILSVRC2012~\cite{ILSVRC15}   &$224\times224\times3$ &\XSolid &Pre-Logits\\
Mobilenet V1~\cite{Howard2017MobileNetsEC}     & Classification & ILSVRC2012~\cite{ILSVRC15}   &$224\times224\times3$ &\XSolid &Pre-Logits\\
$50\%$ Mobilenet V1~\cite{Howard2017MobileNetsEC}     & Classification & ILSVRC2012~\cite{ILSVRC15}   &$160\times160\times3$ &\XSolid &Pre-Logits\\
$25\%$ Mobilenet V1~\cite{Howard2017MobileNetsEC}     & Classification & ILSVRC2012~\cite{ILSVRC15}   &$128\times128\times3$ &\XSolid &Pre-Logits\\
\midrule
Generative Inpainting~\cite{yu2018generative}   & Inpainting  &Places2~\cite{zhou2018places} &$512\times680\times3$ &\XSolid &``allconv12''\dag\\
Generative Inpainting~\cite{yu2018generative}   & Inpainting  &CelebA~\cite{Liu2015DeepLF} &$256\times256\times3$ &\XSolid &``allconv12''\dag\\
Generative Inpainting~\cite{yu2018generative}   & Inpainting  &CelebA-HQ~\cite{Karras2018ProgressiveGO} &$256\times256\times3$ &\XSolid &``allconv12''\dag\\
Generative Inpainting~\cite{yu2018generative}   & Inpainting  &ILSVRC2012~\cite{ILSVRC15}  &$256\times256\times3$ &\XSolid &``allconv12''\dag\\
\midrule
FCRN~\cite{laina2016deeper}   & Depth Estimation  &NYU Depth v2~\cite{silberman2012indoor} &$512\times512\times3$ &\Checkmark &``layer1''\dag\\
PRN~\cite{Feng2018Joint3F}   & Face Alignment  &300W-LP~\cite{Zhu2016FaceAA} &$256\times256\times3$ &\XSolid &``ResBlock10''\dag\\
FCN~\cite{Long2015FullyCN}   & Semantic Segmentation  &PASCAL VOC~\cite{Everingham2009ThePV} &$512\times512\times3$ &\Checkmark &``Conv8''\dag\\
Tiny Face Detector~\cite{Hu2017FindingTF}   & Face Detection  &WIDER FACE~\cite{yang2016wider} &$512\times512\times3$ &\Checkmark &``res4b''\dag\\
%DCGAN~\cite{nazeri2018image}   & Colorization  &Places365~\cite{Zhou2016PlacesAI} &$256\times256\times1$ &\XSolid &\\
\bottomrule
\end{tabular}
\begin{tablenotes}
   \item [$\dag$]: the name given in the source code.% or \item [a]
\end{tablenotes}
\end{threeparttable}
\end{table}

\begin{figure}[t]
\centering
\includegraphics[scale=0.40]{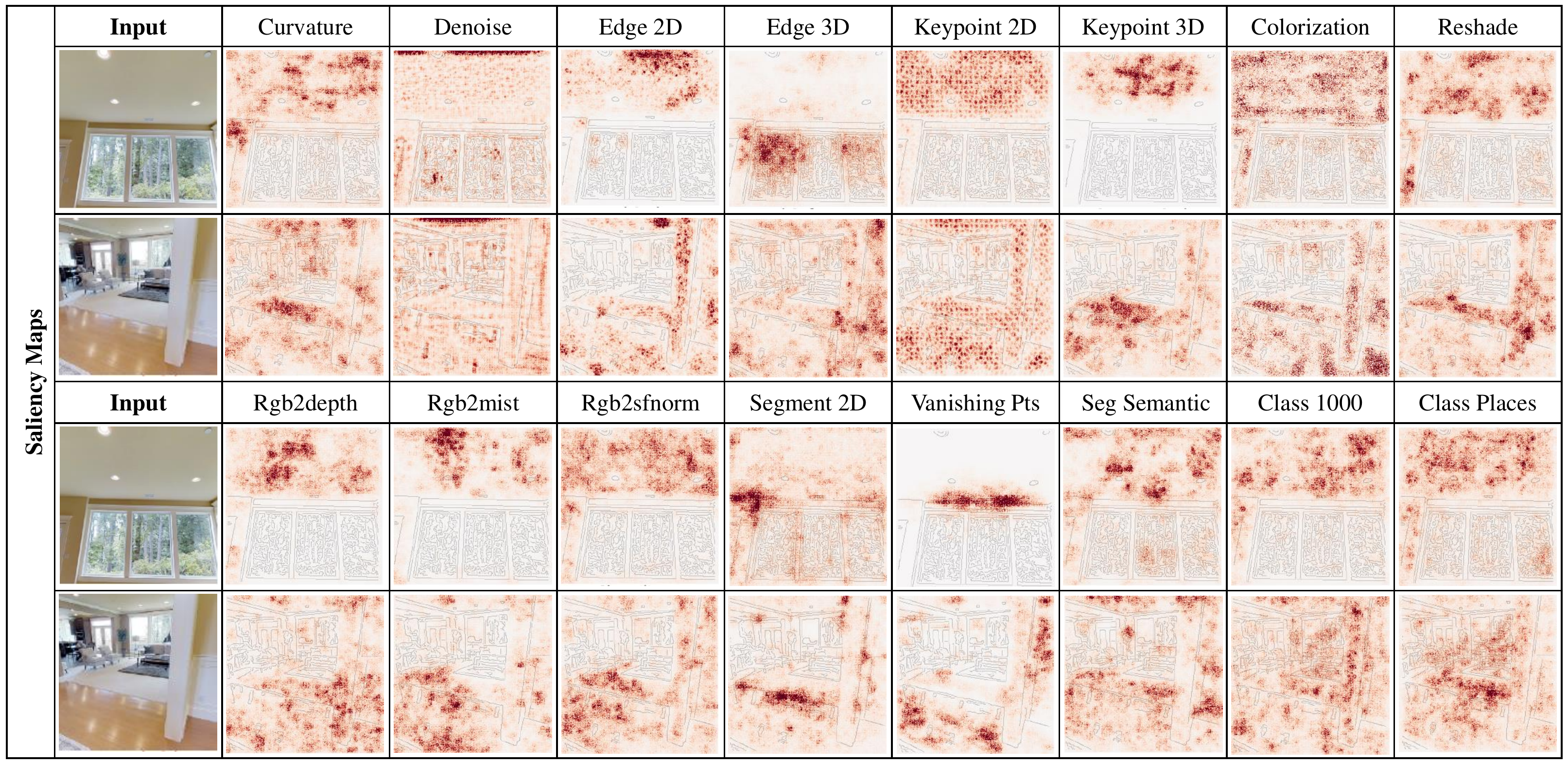}\\
\includegraphics[scale=0.40]{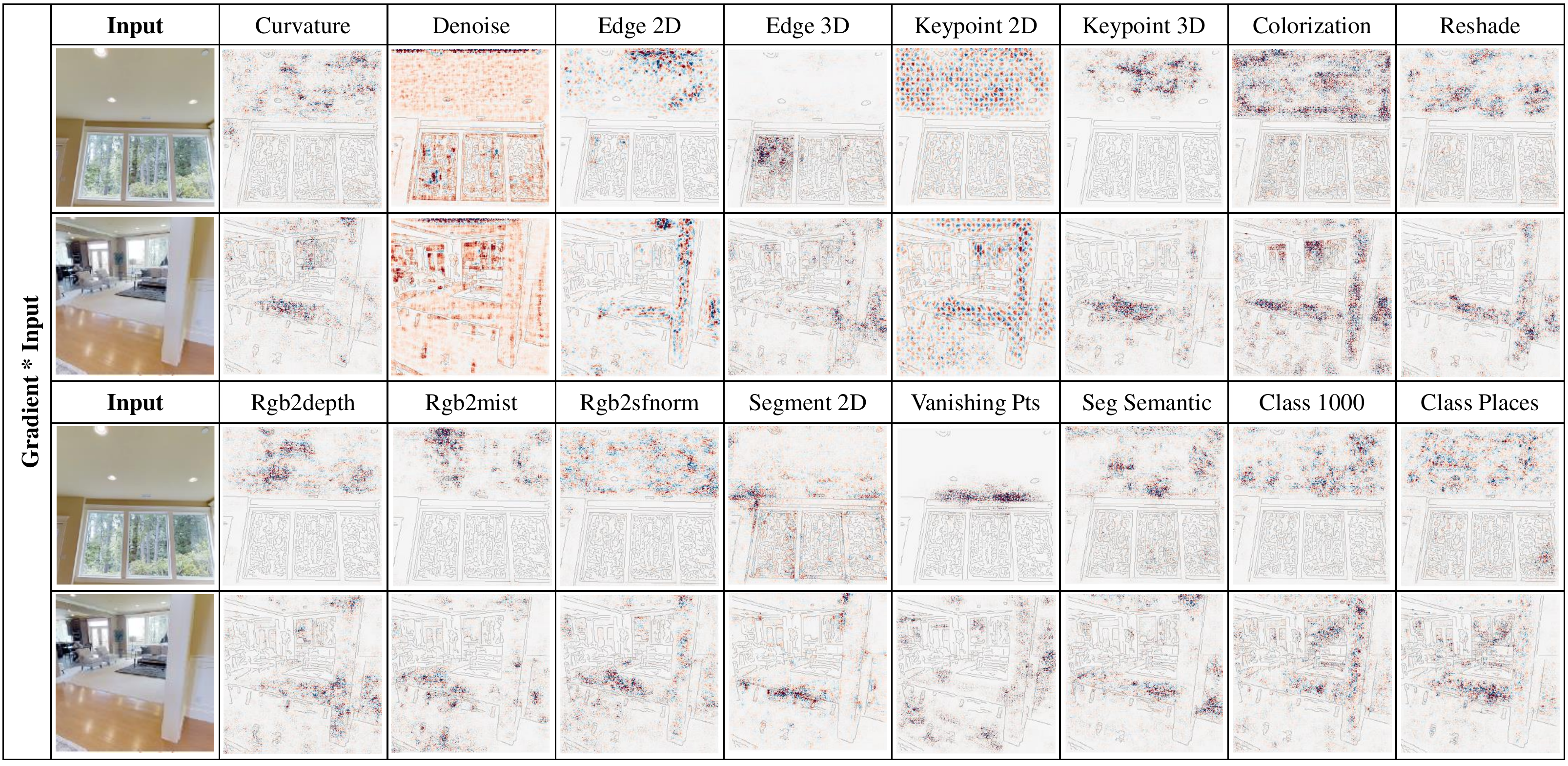}
\centering
\caption{Visualization of attribution maps produced by saliency~\cite{Simonyan2013DeepIC} and gradient*input~\cite{Shrikumar2016NotJA}.}
\label{fig:vis_taskonomy}
\end{figure}

To further validate our method, we also collect another 18 pre-trained models online beyond those involved in taskonomy, including two VGGs~\cite{Simonyan2015VeryDC} (VGG16, VGG19), three ResNets~\cite{He2016DeepRL} (ResNet50, ResNet101, ResNet152), two Inceptions (Inception V3~\cite{Szegedy2016RethinkingTI}, Inception ResNet V2~\cite{szegedy2017inception}), three MobileNets~\cite{Howard2017MobileNetsEC} (MobileNet, 0.5 MobileNet, 0.25 MobileNet), four Inpaintings~\cite{yu2018generative} (ImageNet, CelebA, CelebA-HQ, Places), FCRN~\cite{laina2016deeper}, FCN~\cite{Long2015FullyCN}, PRN~\cite{Feng2018Joint3F} and Tiny Face Detector~\cite{Hu2017FindingTF}. Details of these models are summarized in Table~\ref{table:collected_models}. These models can be further categorized into three groups:
\begin{itemize}[leftmargin=15pt]
	\item Classification models: trained on the same data (ILSVRC2012~\cite{ILSVRC15}), for the same task (1000-way classification), but in different model architectures (Inception, ResNet, VGG, MobileNet). We adopt the pre-trained models released by Tensorflow-Slim\footnote{\url{https://github.com/tensorflow/models/tree/master/research/slim}}~\cite{silberman2016tensorflow} lib.
	\item Inpainting models: in the same model architecture, trained for the same task, but on different datasets (Places2~\cite{zhou2018places}, CelebA~\cite{Liu2015DeepLF}, CelebA-HQ~\cite{Karras2018ProgressiveGO} and ILSVRC2012~\cite{ILSVRC15}). We adopt the pre-trained models released by~\cite{yu2018generative}\footnote{\url{https://github.com/JiahuiYu/generative_inpainting}}.
	\item Other models: models in this group are heterogeneous in architectures, tasks and training data. This group consists of four models, including FCRN~\cite{laina2016deeper}, FCN~\cite{Long2015FullyCN}, PRN~\cite{Feng2018Joint3F} and Tiny Face Detector~\cite{Hu2017FindingTF}. Pre-trained models can be found in their project pages.
\end{itemize}

In our experiments, we also merge the two groups of models to form a more comprehensive group. Experiments on this new group will provide us more insights into the proposed method.

\section{Probe Datasets}
Here we provide more details about the three probe datasets used in the proposed method.
\paragraph{Taskonomy data} On taskonomy data~\cite{Zamir_2018_CVPR}, we construct the probe data by selecting images from the validation data of the ``Tiny'' partition. In the validation set of Tiny partition, images are collected from $5$ different buildings. We randomly select $200$ images from each of these $5$ buildings, constructing a probe dataset consisting of $1,000$ images.
\paragraph{Indoor Scene} Indoor Scene~\cite{Quattoni2009RecognizingIS} is a dataset used for indoor scene recognition. The original database contains $67$ indoor categories, and a total of $15,620$ images. We randomly select 15 images from each of these $67$ categories, constructing a probe dataset consisting of $1,005$ images.
\paragraph{COCO} The COCO~\cite{Lin2014MicrosoftCC} dataset is designed for several purpose such as detection, caption and so on. On this dataset, we randomly select $1,000$ images from the 2014 Val dataset to construct the probe dataset for evaluating the proposed method.

The styles of images in these three datasets are very different. Generally speaking, the textures of images in taskonomy data are simple. However, the textures of images in Indoor Scene and COCO are relatively more complex.

\section{Visualization of Attribution Maps}
\begin{figure}[t]
\centering
\subfigure{
\begin{picture}(0,0)
\put(50,105){\textbf{Taskonomy}~\cite{Zamir_2018_CVPR}}
\put(170,105){\textbf{Indoor Scene}~\cite{Quattoni2009RecognizingIS}}
\put(300,105){\textbf{COCO}~\cite{Lin2014MicrosoftCC}}
\put(10,30){\rotatebox{90}{\textbf{Saliency}~\cite{Simonyan2013DeepIC}}}
\end{picture}
\begin{minipage}[t]{\linewidth}
\centering
\includegraphics[scale=0.15]{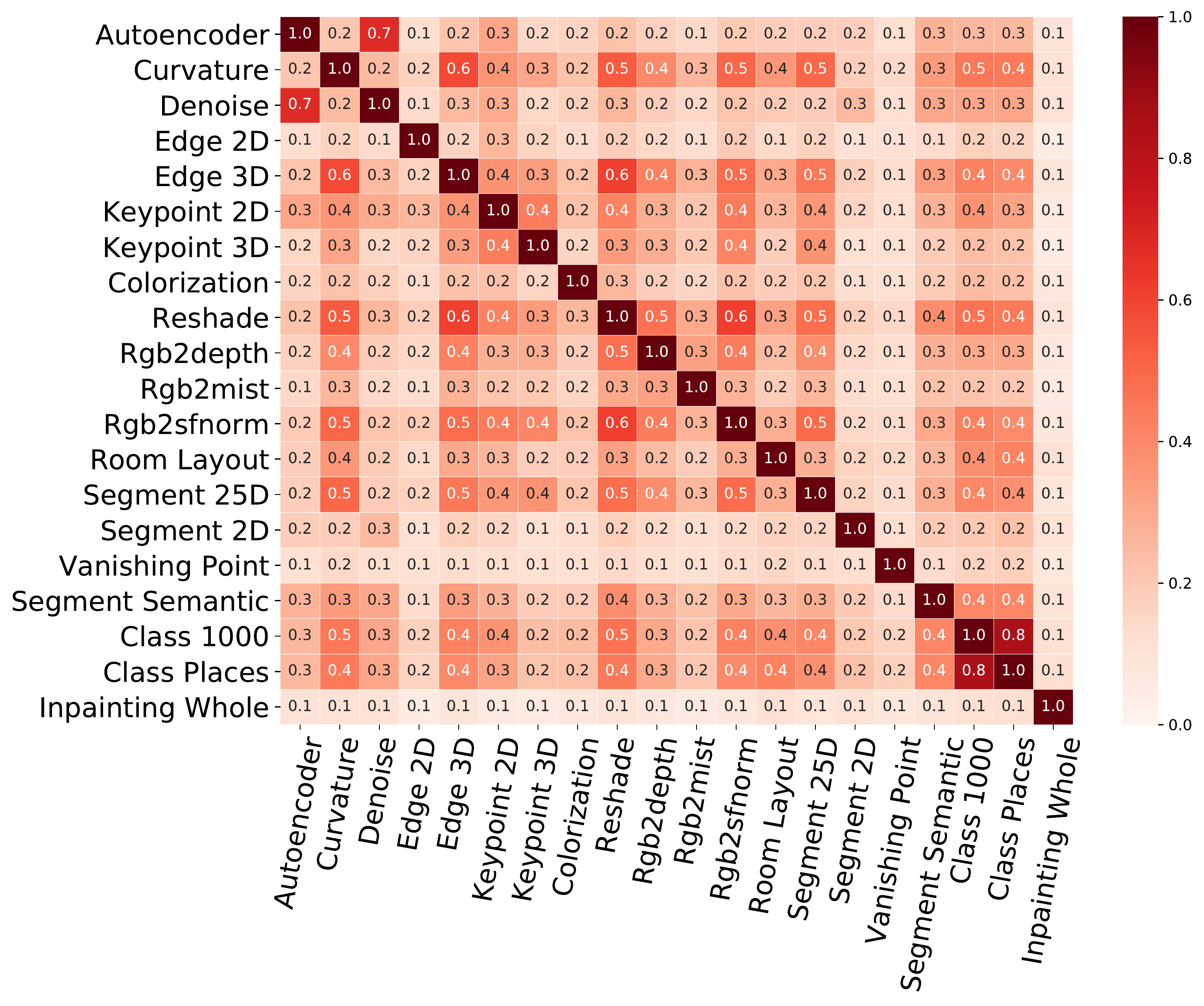}
\includegraphics[scale=0.15]{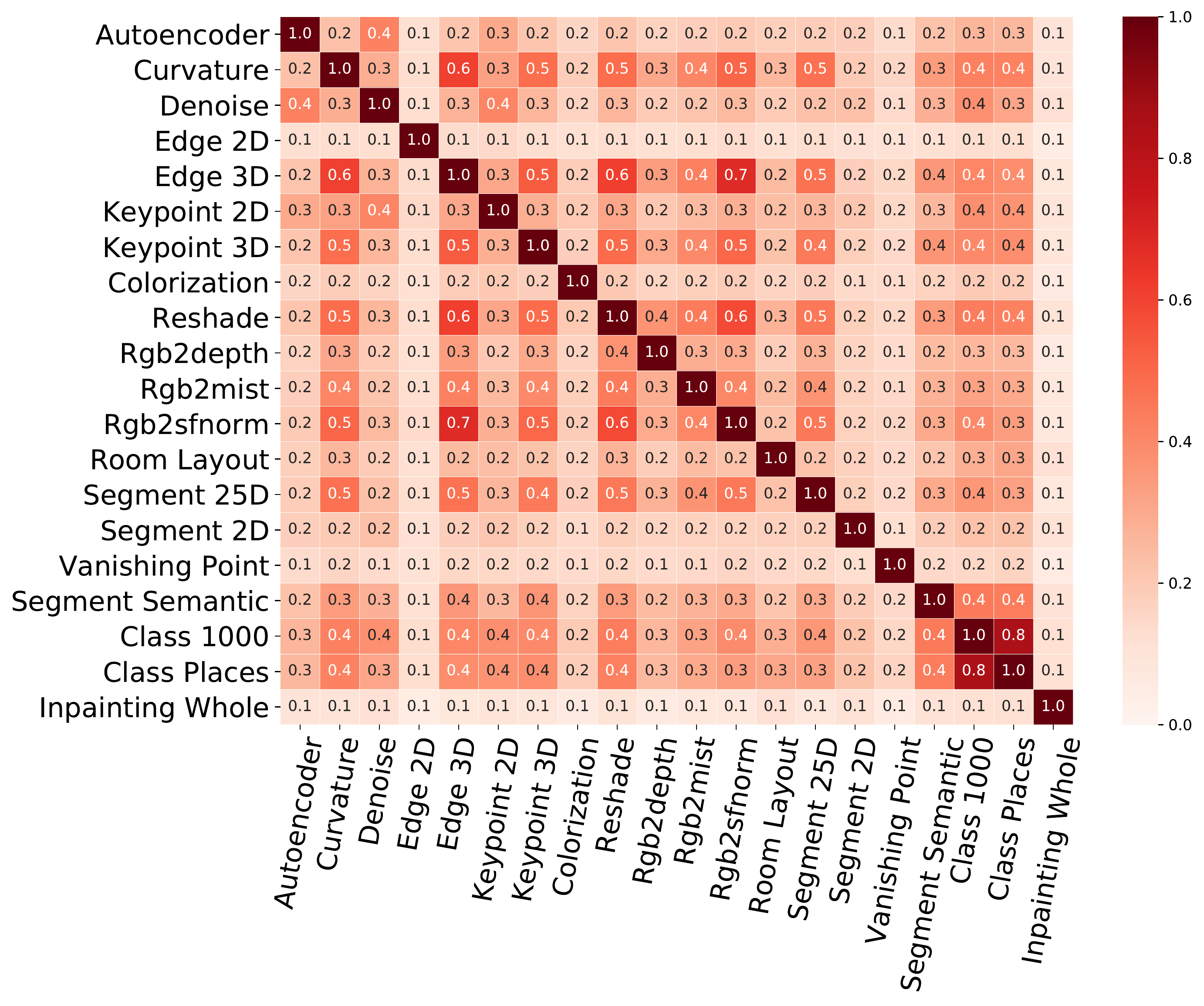}
\includegraphics[scale=0.15]{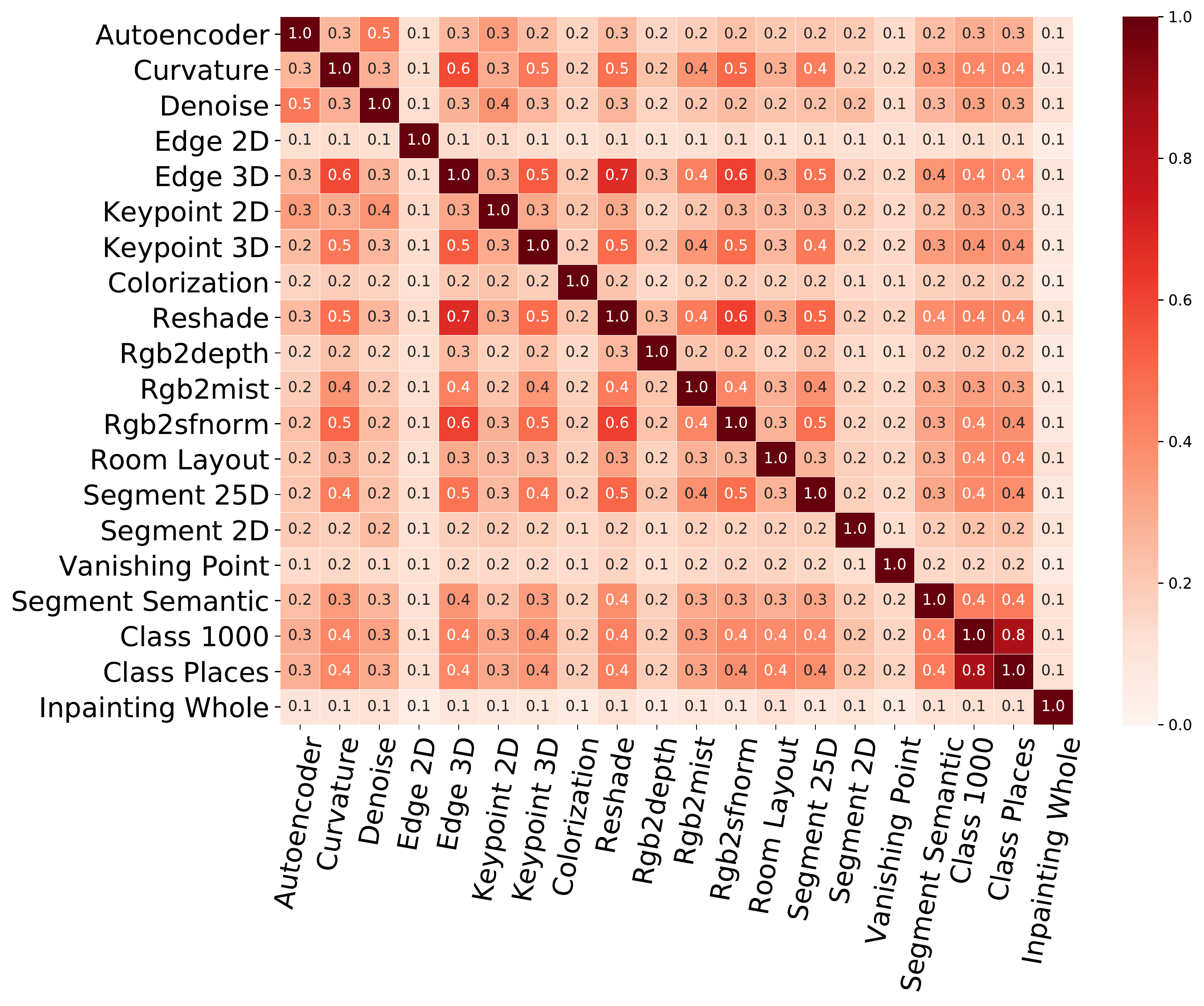}
\end{minipage}%
}%

\subfigure{
\begin{picture}(0,0)
\put(10,8){\rotatebox{90}{\textbf{Gradient*Input}~\cite{Shrikumar2016NotJA}}}
\end{picture}
\begin{minipage}[t]{\linewidth}
\centering
\includegraphics[scale=0.15]{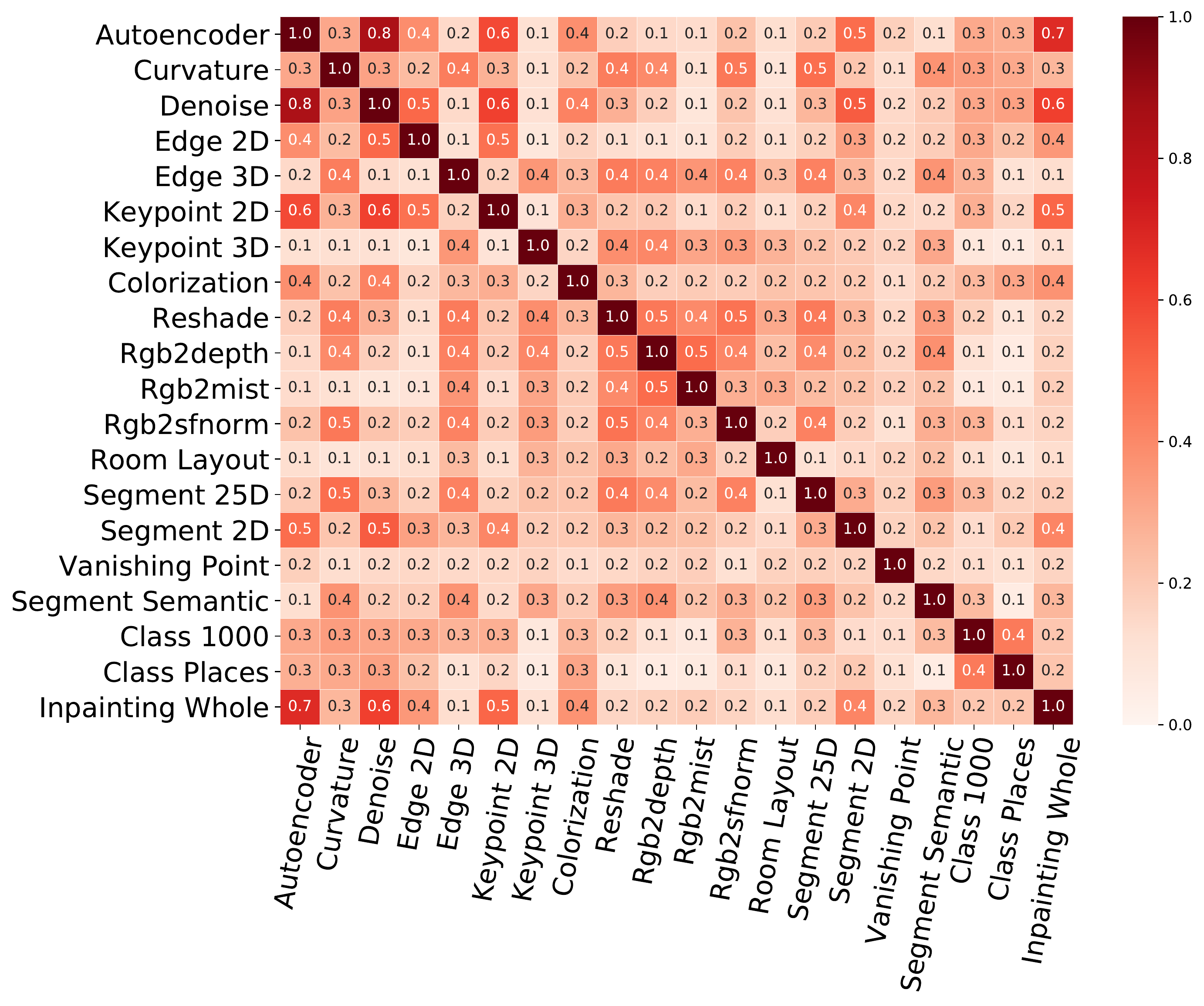}
\includegraphics[scale=0.15]{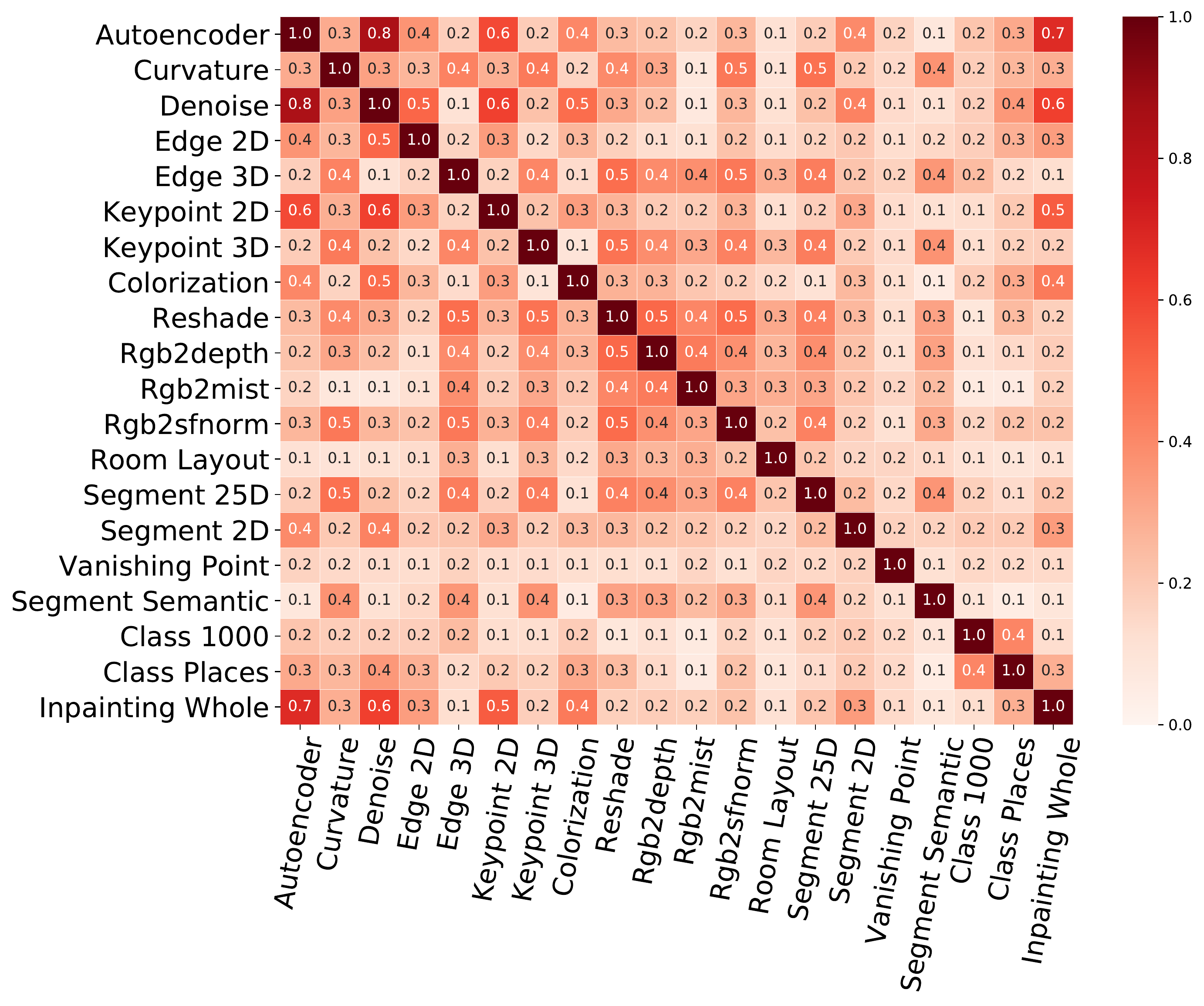}
\includegraphics[scale=0.15]{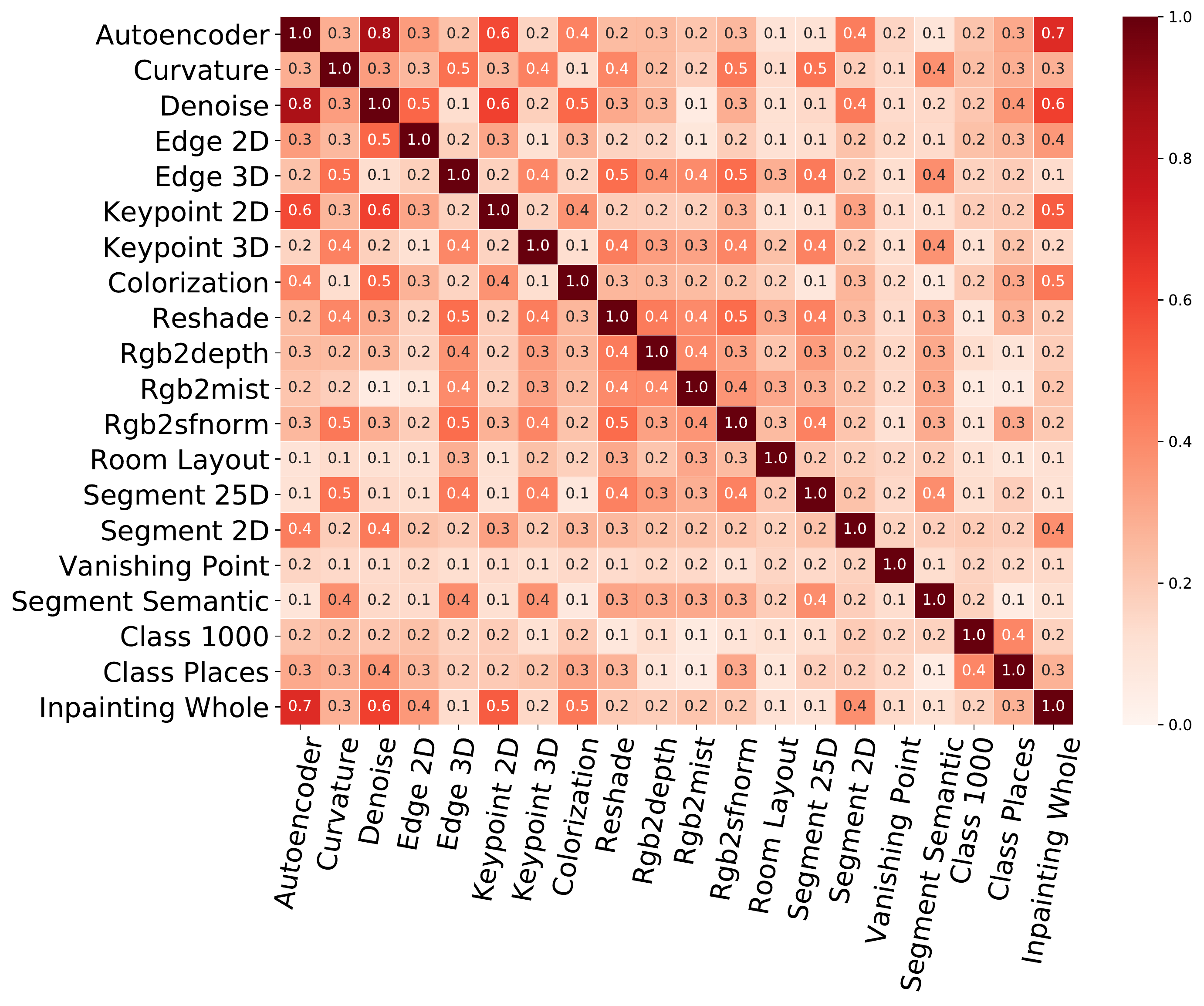}
\end{minipage}%
}%

\subfigure{
\begin{picture}(0,0)
\put(10,30){\rotatebox{90}{$\epsilon$\textbf{-LRP}~\cite{Bach2015OnPE}}}
\end{picture}
\begin{minipage}[t]{\linewidth}
\centering
\includegraphics[scale=0.15]{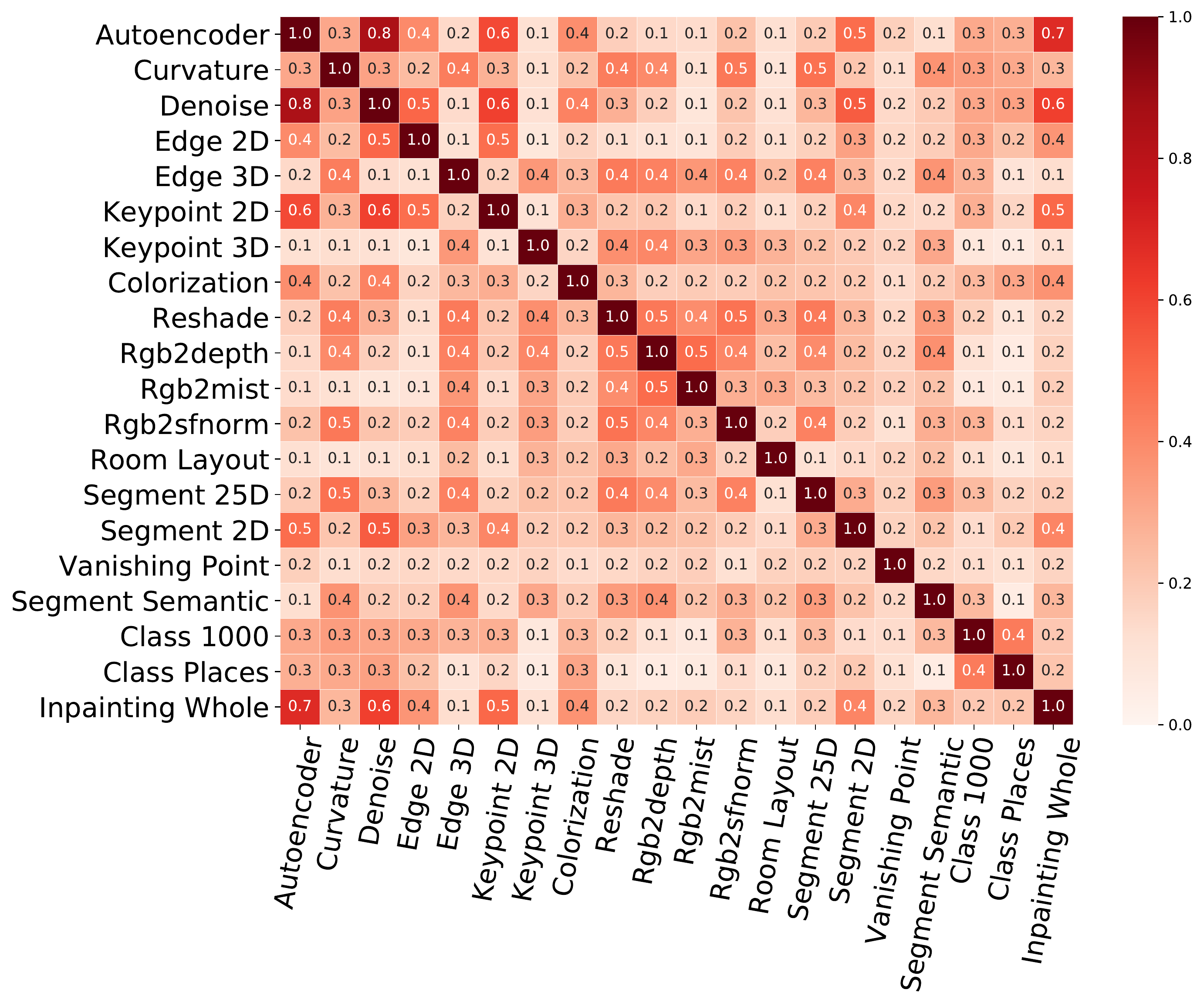}
\includegraphics[scale=0.15]{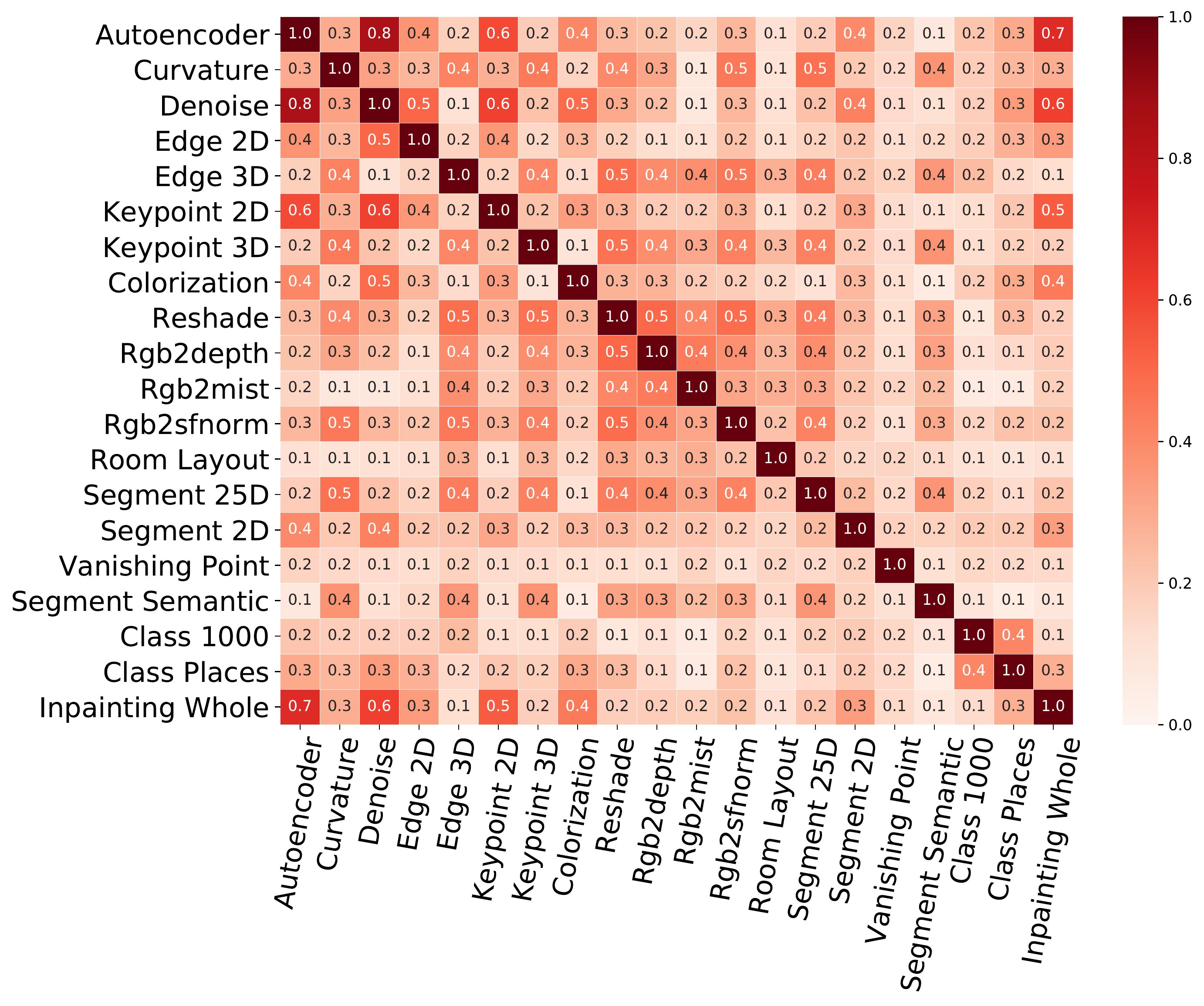}
\includegraphics[scale=0.15]{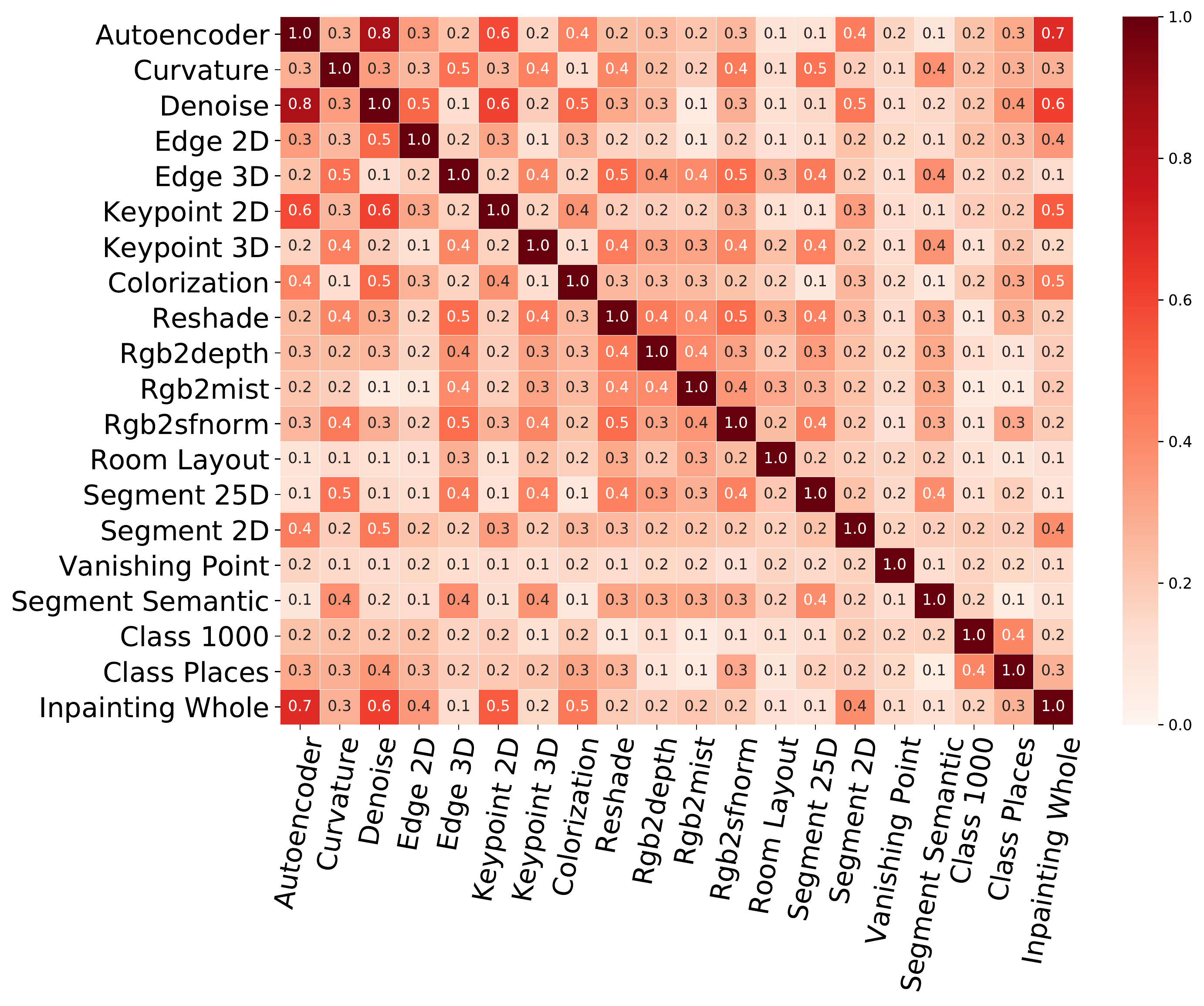}
\end{minipage}%
}%
\centering
\caption{Visualization of transferability matrices produced by saliency~\cite{Simonyan2013DeepIC}, gradient*input~\cite{Shrikumar2016NotJA} and $\epsilon$-LRP~\cite{Bach2015OnPE} on images randomly selected from taxonomy data~\cite{Zamir_2018_CVPR}, indoor scene~\cite{Quattoni2009RecognizingIS}, COCO data~\cite{Lin2014MicrosoftCC}.}
\label{fig:vis_heatmap}
\end{figure}
In this section, we visualize attribution maps of examples from taskonomy data~\cite{Zamir_2018_CVPR} for better understanding of our method. Here attribution maps are produced by saliency maps~\cite{Simonyan2013DeepIC} and gradient * input~\cite{Shrikumar2016NotJA}. Results are visualized in Figure~\ref{fig:vis_taskonomy}. It can be seen that some tasks tend to produce much more similar attribution maps than others. For example, <Rgb2depth, Rgb2mist> and <Class 1000, Class Places>. These producing-similar-attribution tasks are proved to be highly related in the task structure found in taskonomy and thus producing favorable transfer performance to each other. Some examples may produce misleading results. However, the conclusions made by statistically aggregating the results of all the randomly sampled examples become more reliable.
%Additionally, gradient*input~\cite{Shrikumar2016NotJA} and $\epsilon$-LRP~\cite{Bach2015OnPE} seem to produce similar results, while saliency produces somewhat different results. More comparisons and discussion will be provided in the following sections.

\section{Visualization of Deep Model Transferability}

\begin{figure}[t]
\centering
\subfigure{
\begin{picture}(0,0)
\put(40,95){\textbf{Taskonomy}~\cite{Zamir_2018_CVPR}}
\put(160,95){\textbf{Indoor Scene}~\cite{Quattoni2009RecognizingIS}}
\put(300,95){\textbf{COCO}~\cite{Lin2014MicrosoftCC}}
\put(1,15){\rotatebox{90}{\textbf{Saliency}~\cite{Simonyan2013DeepIC}}}
\end{picture}
\begin{minipage}[t]{\linewidth}
\centering
\includegraphics[scale=0.25]{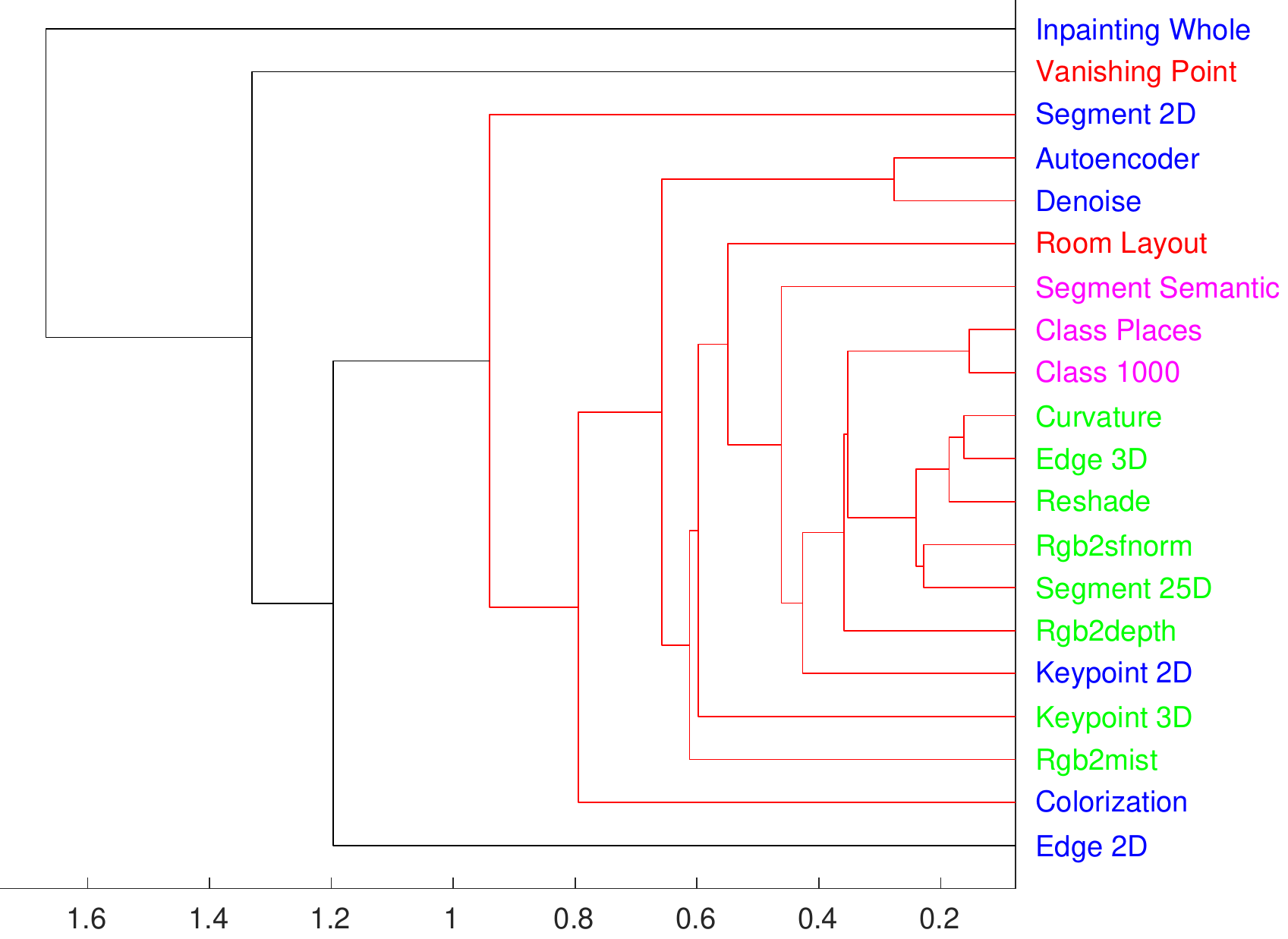}
\includegraphics[scale=0.25]{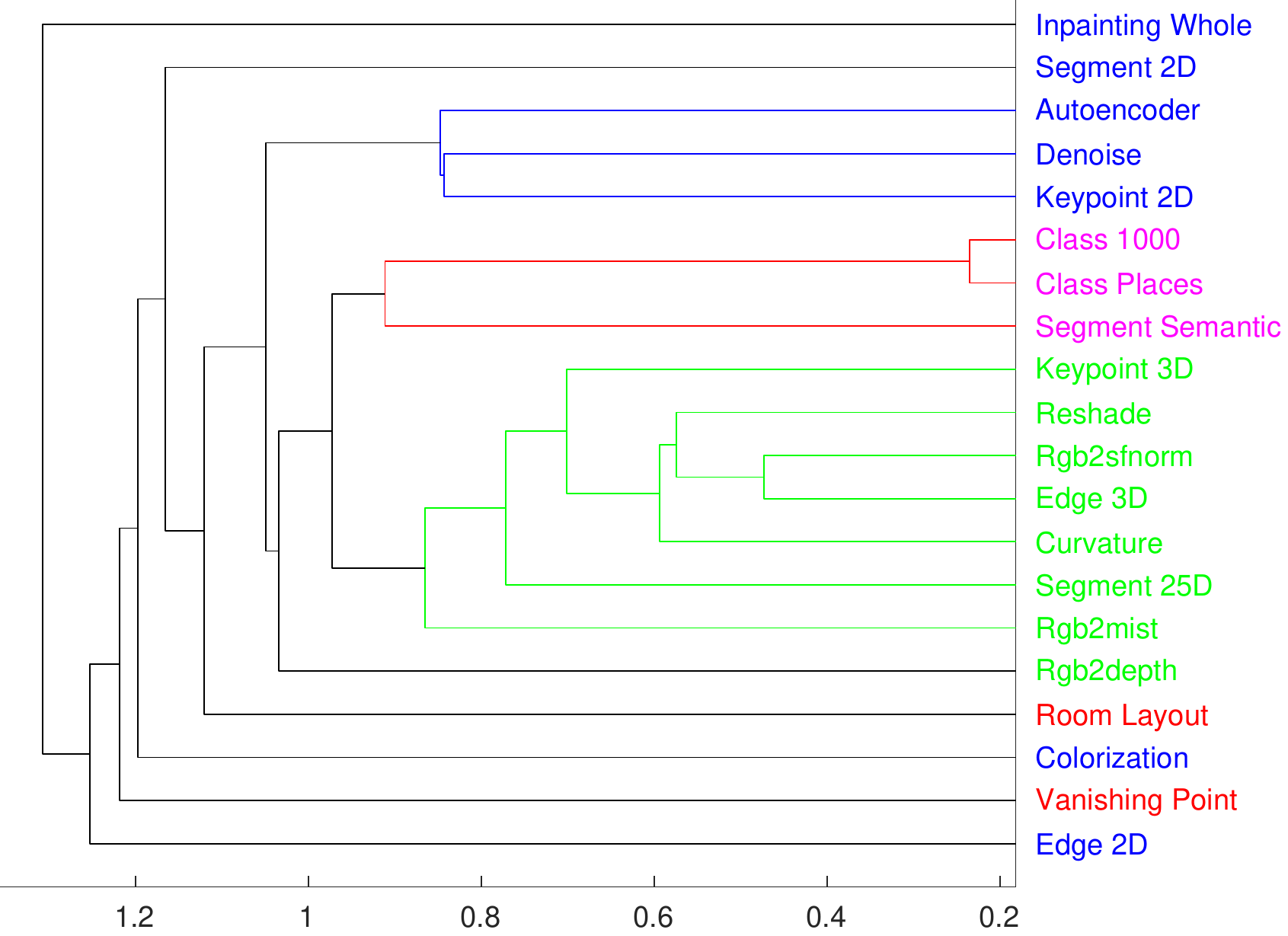}
\includegraphics[scale=0.25]{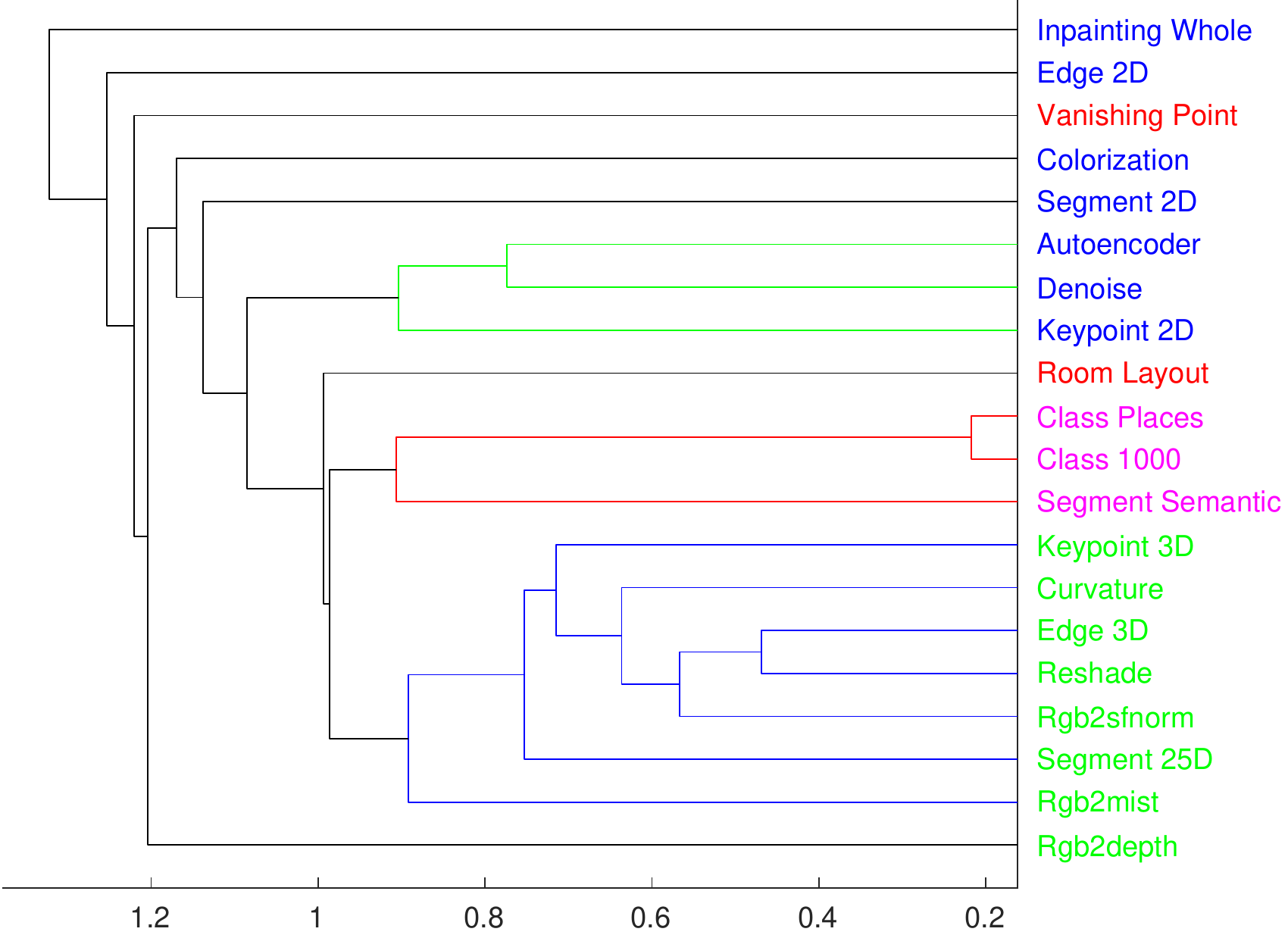}
\end{minipage}%
}%

\subfigure{
\begin{picture}(0,0)
\put(1,3){\rotatebox{90}{\textbf{Gradient*Input}~\cite{Shrikumar2016NotJA}}}
\end{picture}
\begin{minipage}[t]{\linewidth}
\centering
\includegraphics[scale=0.25]{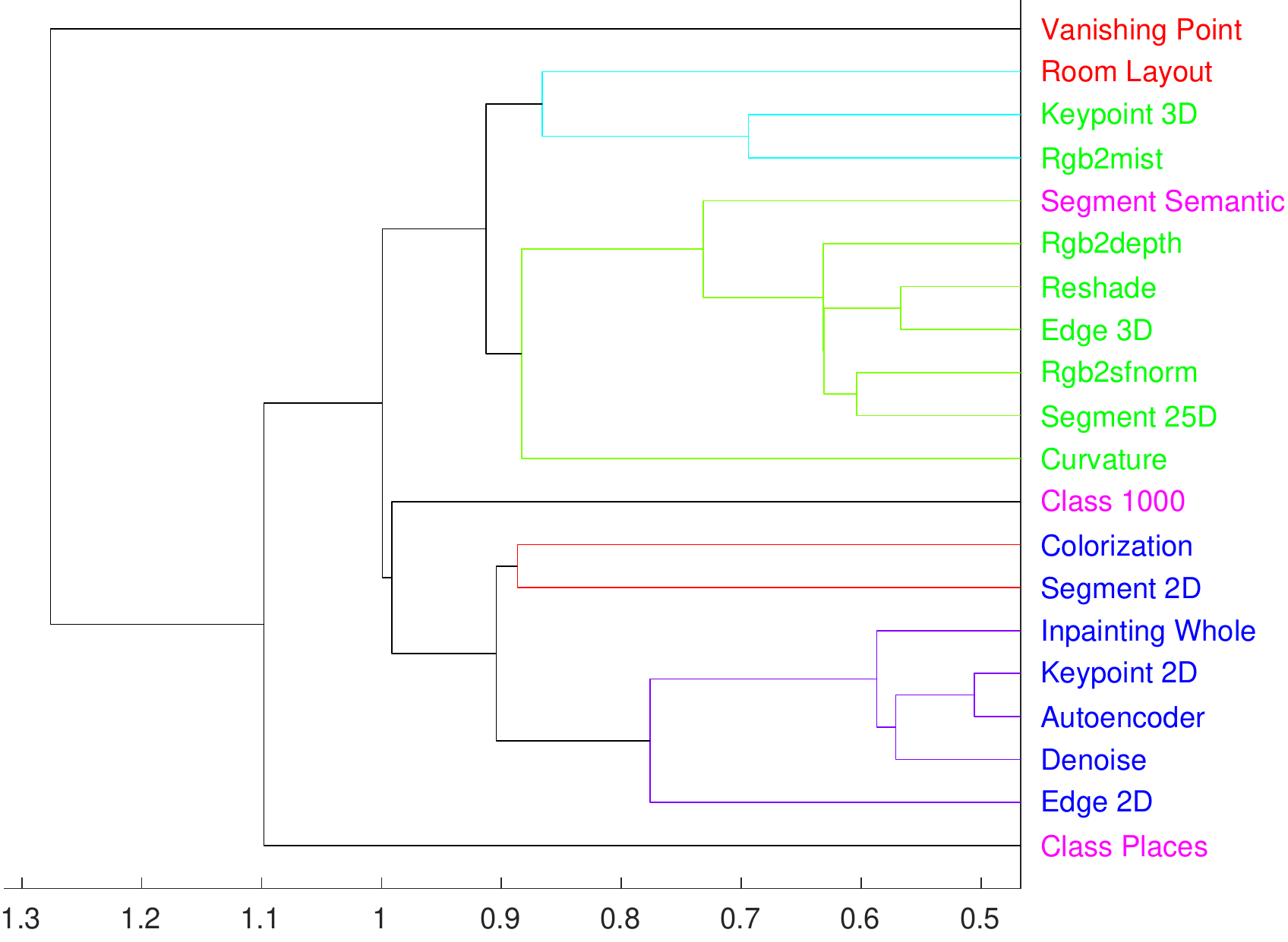}
\includegraphics[scale=0.25]{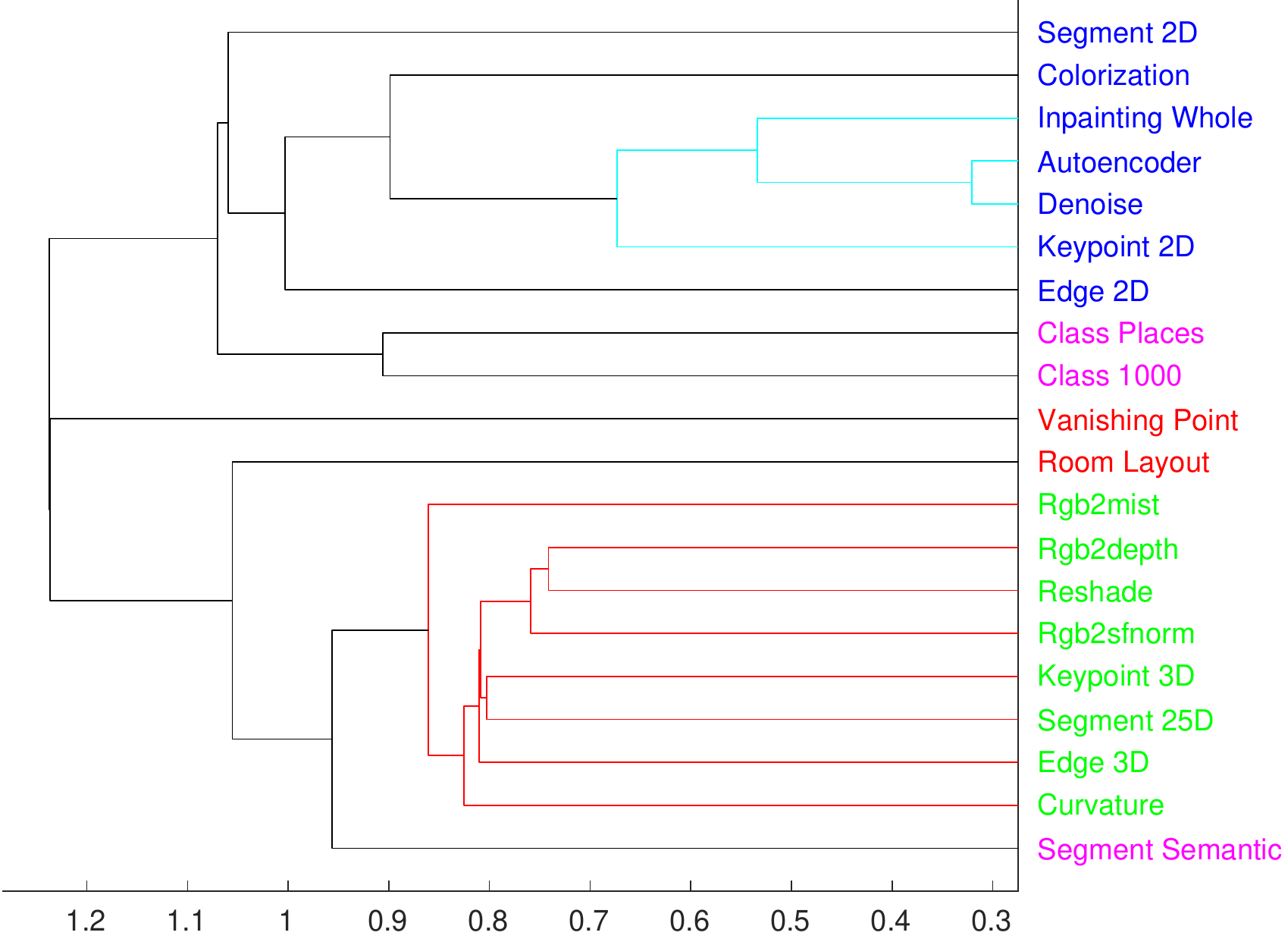}
\includegraphics[scale=0.25]{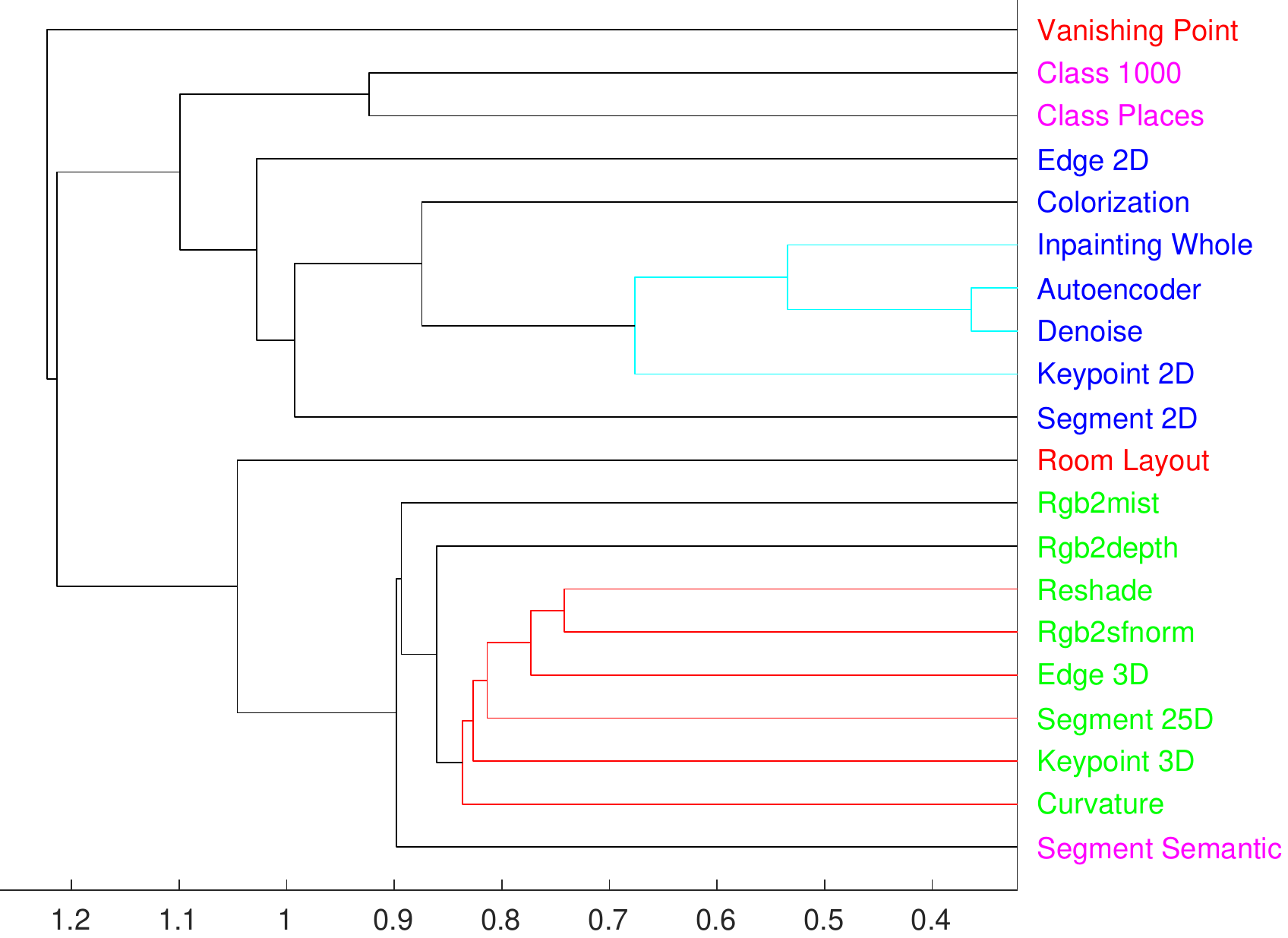}
\end{minipage}%
}%

\subfigure{
\begin{picture}(0,0)
\put(1,20){\rotatebox{90}{$\epsilon$\textbf{-LRP}~\cite{Bach2015OnPE}}}
\end{picture}
\begin{minipage}[t]{\linewidth}
\centering
\includegraphics[scale=0.25]{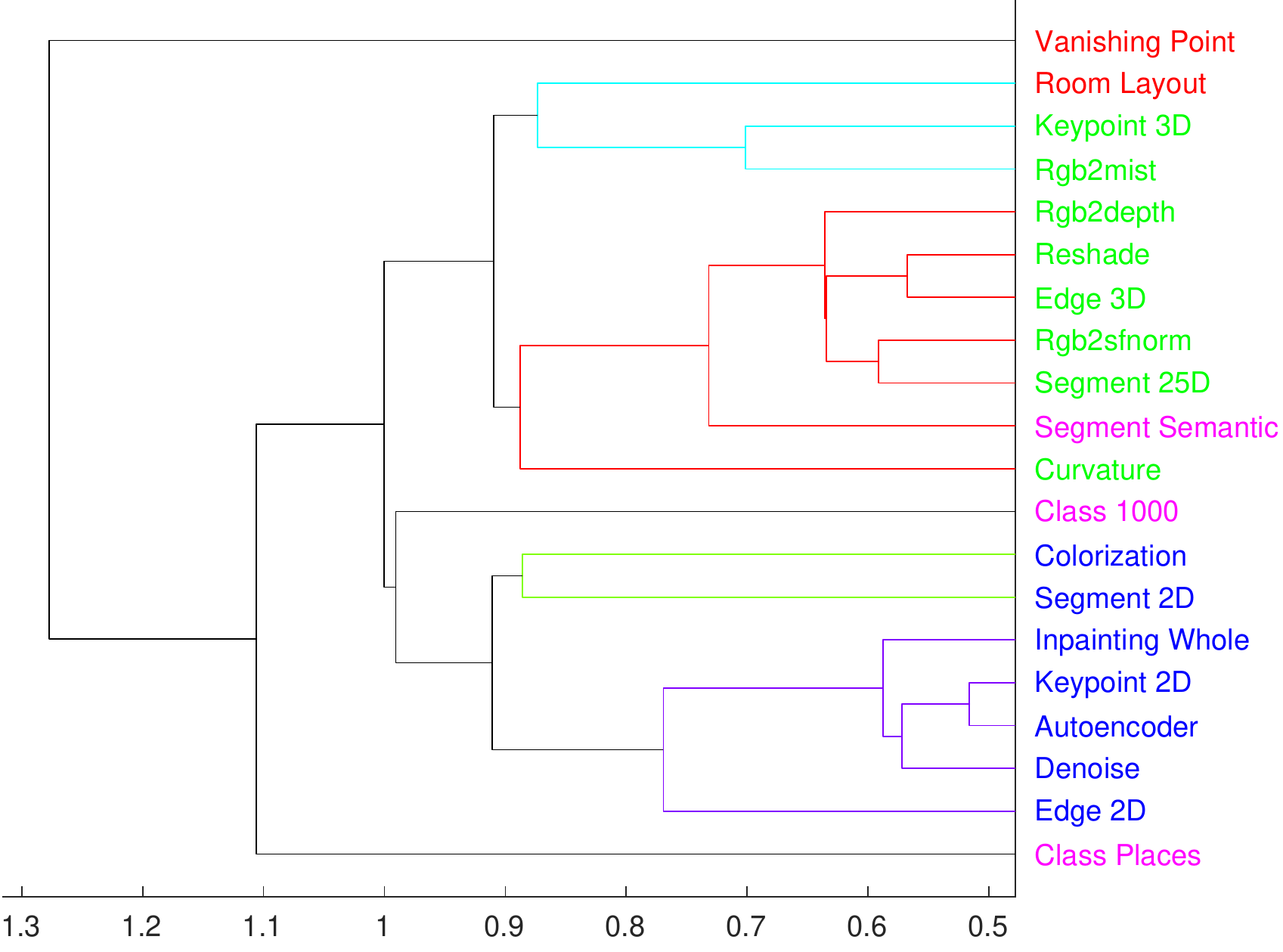}
\includegraphics[scale=0.25]{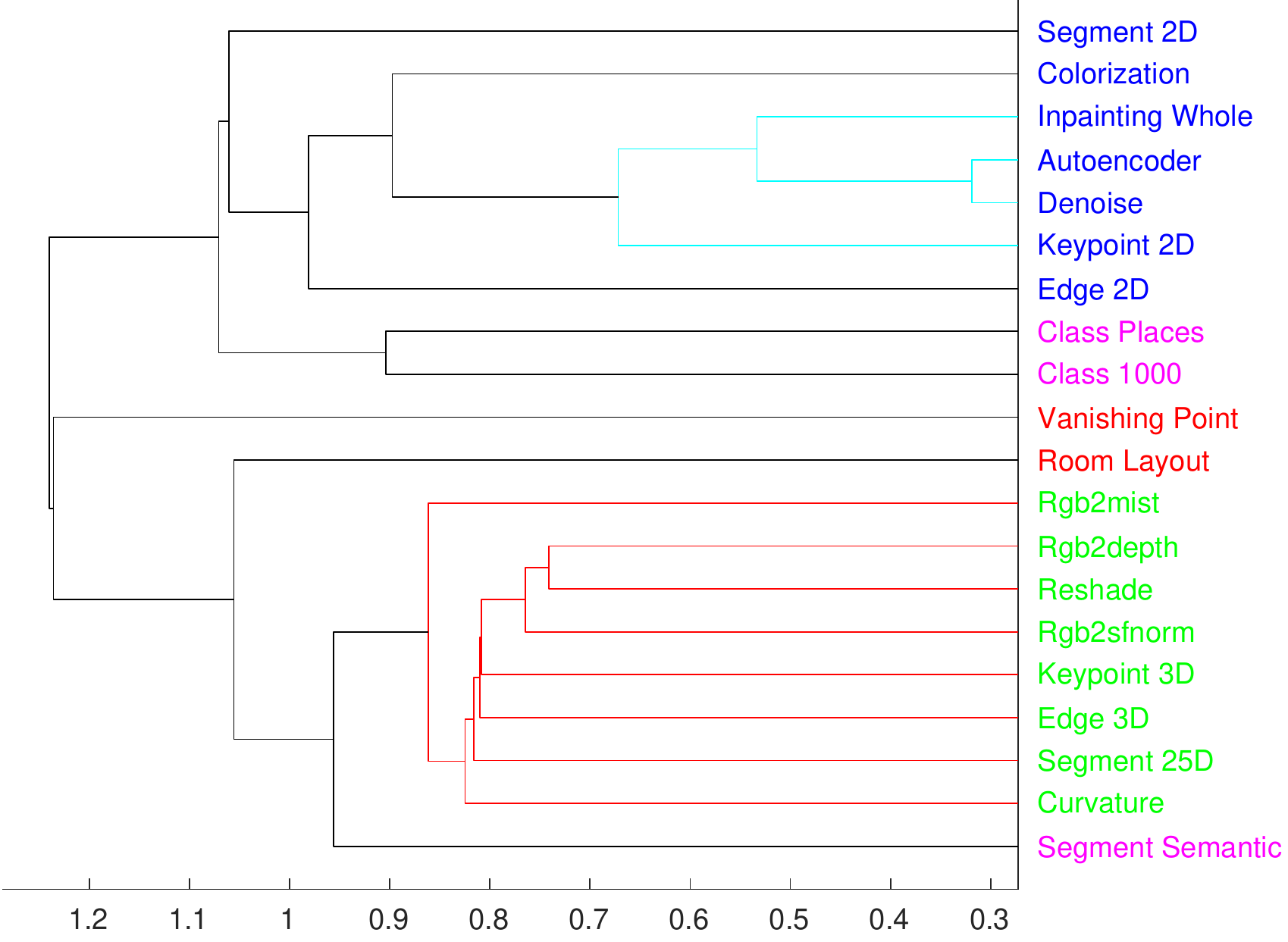}
\includegraphics[scale=0.25]{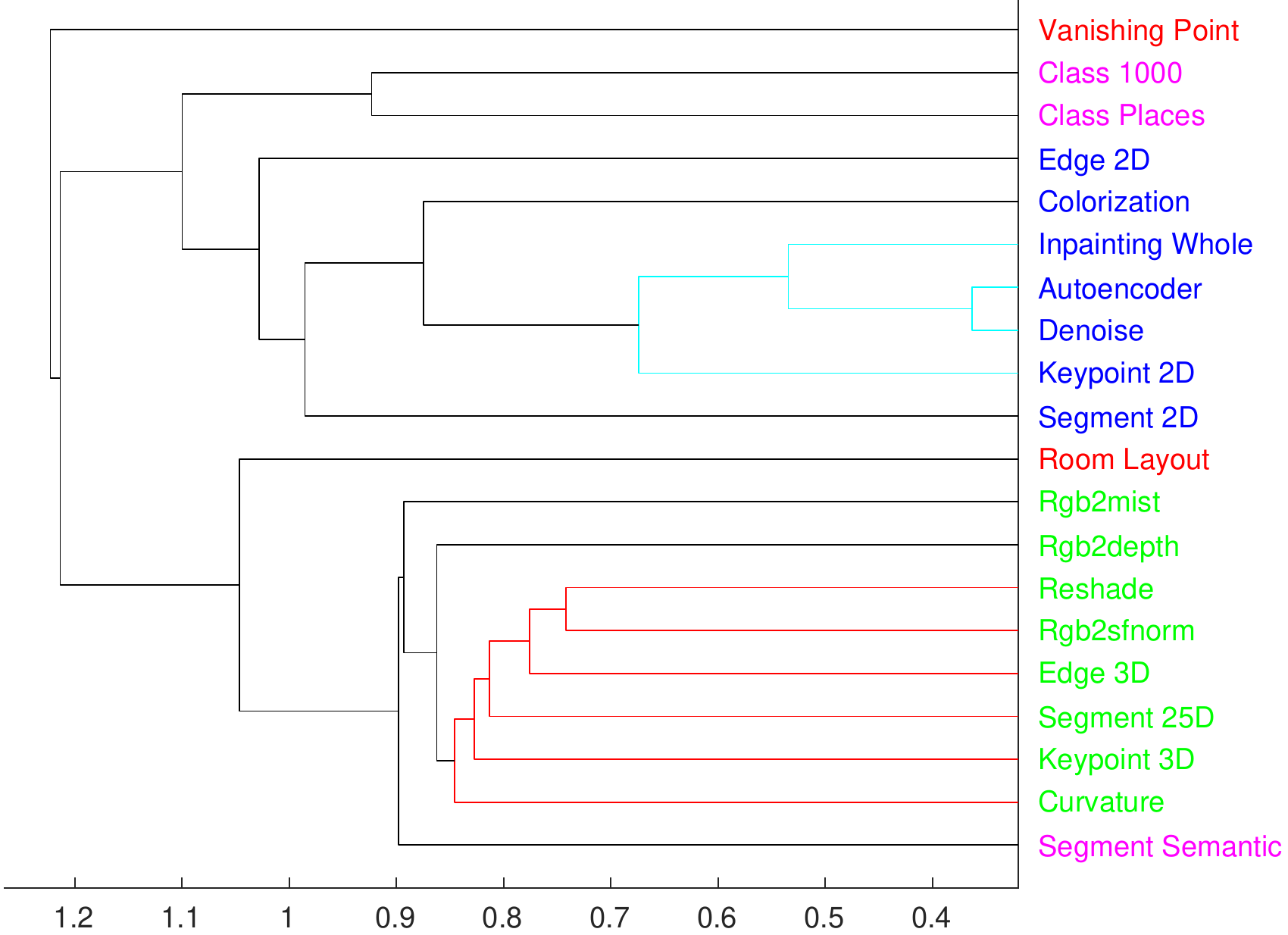}
\end{minipage}%
}%
\centering
\caption{Task similar trees produced by our method with saliency~\cite{Simonyan2013DeepIC}, gradient*input~\cite{Shrikumar2016NotJA} and $\epsilon$-LRP~\cite{Bach2015OnPE} on images randomly selected from taxonomy data~\cite{Zamir_2018_CVPR}, indoor scene~\cite{Quattoni2009RecognizingIS}, COCO data~\cite{Lin2014MicrosoftCC}.}
\label{fig:vis_tree}
\end{figure}

In this section, we provide the results of deep model transferability found by the proposed method. Here we show the model transferability in two way: visualization of the affinity matrix and the task similarity tree. The visualization of attribution maps is provided in Figure~\ref{fig:vis_heatmap}. In this figure, the affinity matrices in each row are produced by the same attribution method which is listed on the left. The probe data used for matrices in each column are listed on the top. The results demonstrate that:
\begin{itemize}[leftmargin=15pt]
\item For each attribution method, no matter which probe data is adopted, the produced affinity matrixes (in the same row) are highly similar. It implies that the proposed method is insensitive to the choice of probe dataset to some degree, which renders our method robust and flexible.
\item On each probe dataset, the transferability matrices produced by $\epsilon$-LRP are visually similar to gradient * input. However, the saliency seems to produce visually more dissimilar results. Actually, the transferability matrices produced by $\epsilon$-LRP and gradient*input are more consistent with that produced by taskonomy than that produced by saliency. The fact accounting for this phenomenon may be that saliency generate attributions entirely based on gradients which denote the direction for optimization. However, the gradients can't fully reflect the relevance between the inputs and the outputs of the deep model, thus leading to inferior results in our method.
    %It implies the attribution method can affect the performance of our method. Devising more suitable attribution methods can further promote the accuracy of our method.
\end{itemize}
To better understand the results produced by different attribution methods and probe data, we construct the task similar trees by agglomerative hierarchical clustering. The constructed task similar trees are depicted in Figure~\ref{fig:vis_tree}. To be consistent with taskonomy~\cite{Zamir_2018_CVPR}, {\color{green}3D}, {\color{blue}2D}, {\color{red}geometric} and {\color{magenta}semantic} tasks are listed in different fonts. Taskonomy has shown that the tasks in the same font plays similar roles in transferring to other tasks, thus they should be clustered together. The results shown in Figure~\ref{fig:vis_tree} indicate that:
\begin{itemize}[leftmargin=15pt]
\item For each attribution method, the task similar trees obtained on the three probe datasets are highly similar. It again verifies that our method is insensitive to the choice of the probe data.
\item For gradient * input~\cite{Shrikumar2016NotJA} and $\epsilon$-LRP~\cite{Bach2015OnPE}, the tasks of the same type are grouped into the same cluster with few exceptions. These results are highly similar to that produced by taskonomy. Considering that our method is much more computation efficient, we argue that our method is much more scalable and practical.
\item If we adopt saliency for attribution in our method, the constructed similar trees are a little different, where the {\color{blue}2D} tasks tend to be clustered into two groups: <Inpainting, Edge2D, Colorization> and <Keypoint2D, Autoencoder, Segment2D>.
\end{itemize}
\begin{figure}[t]
\centering
\subfigure[100]{
\begin{minipage}[t]{0.30\linewidth}
\centering
\includegraphics[scale=0.16]{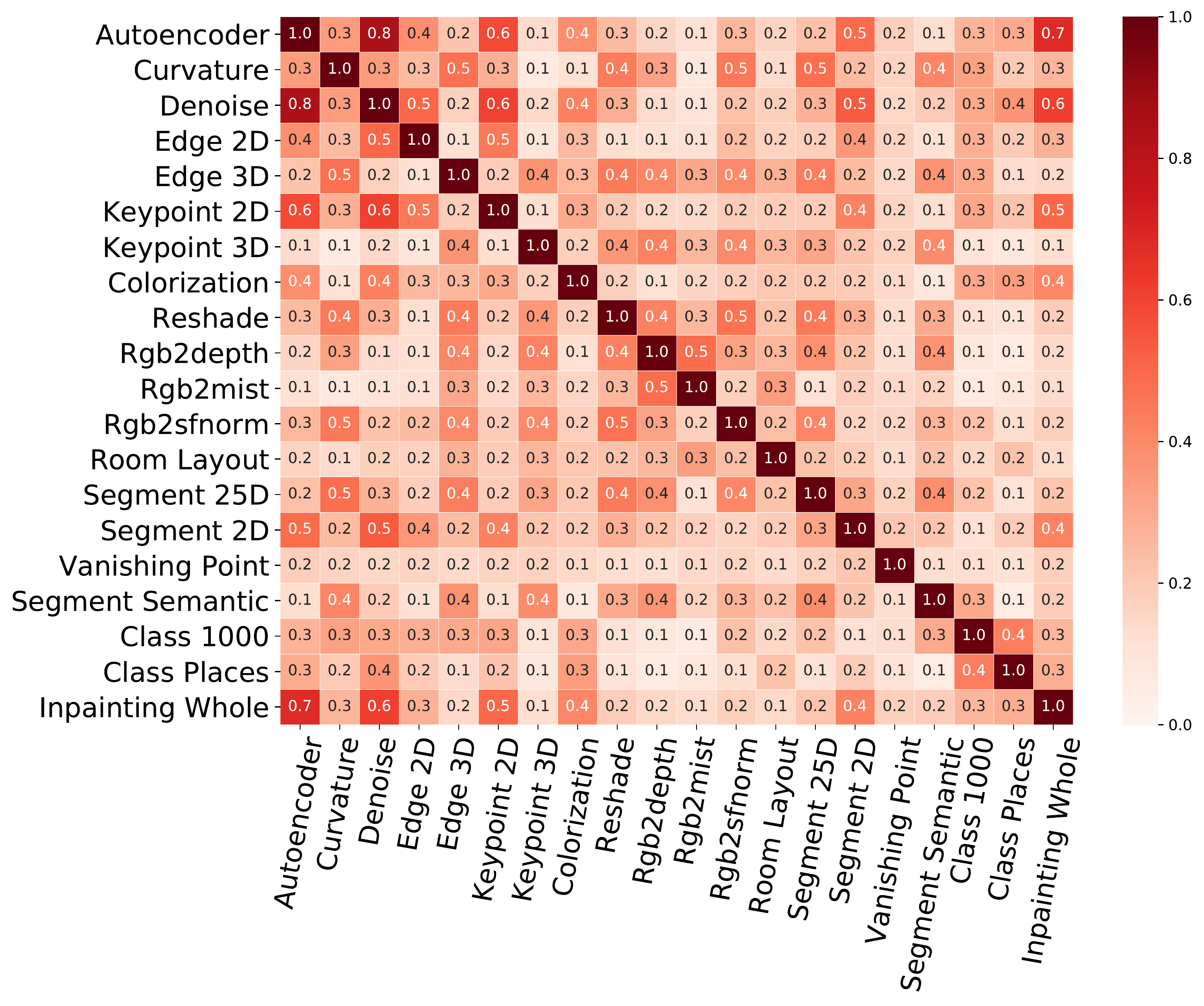}\\
\includegraphics[scale=0.25]{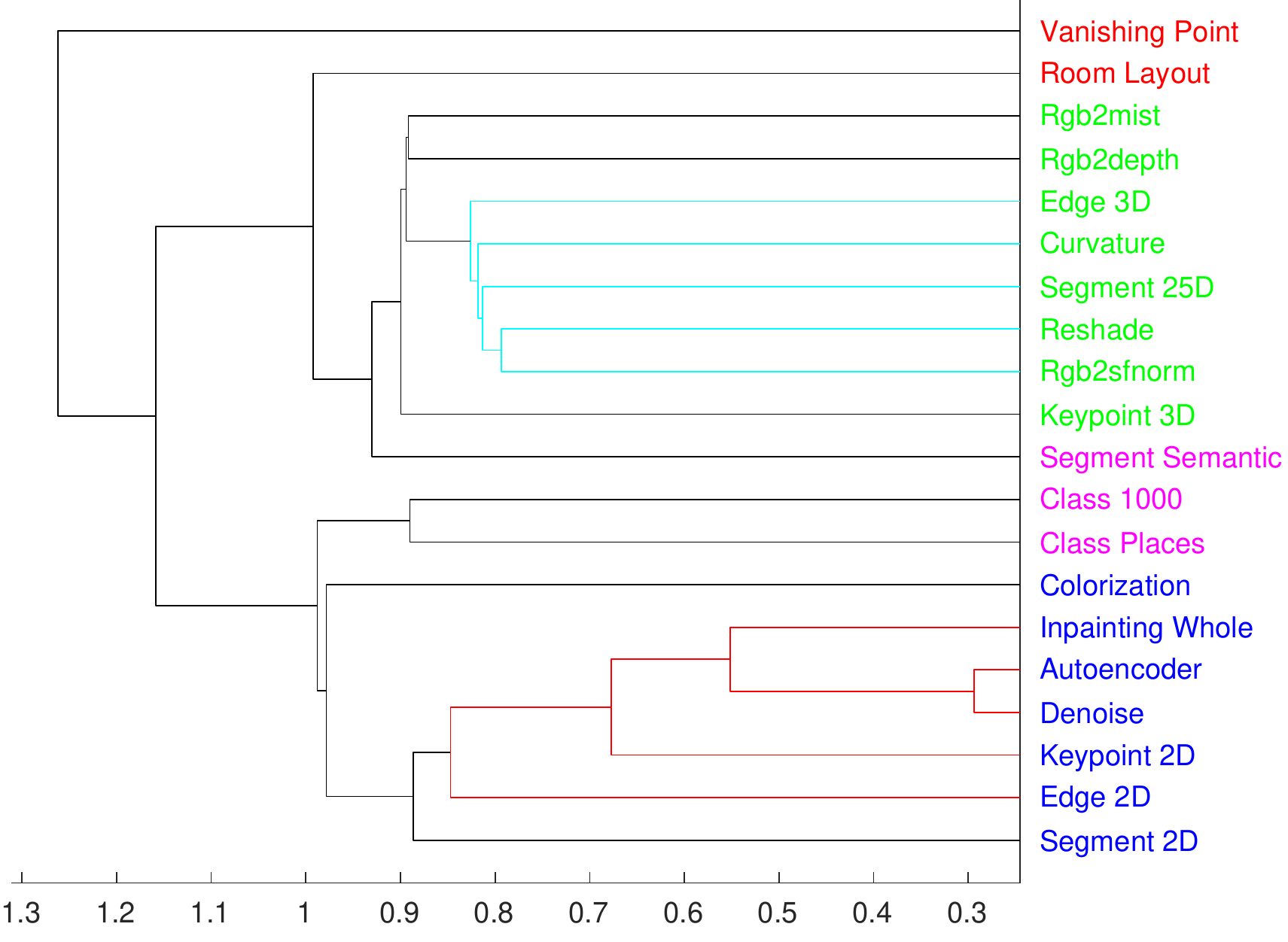}
\end{minipage}%
}%
\subfigure[400]{
\begin{minipage}[t]{0.30\linewidth}
\centering
\includegraphics[scale=0.16]{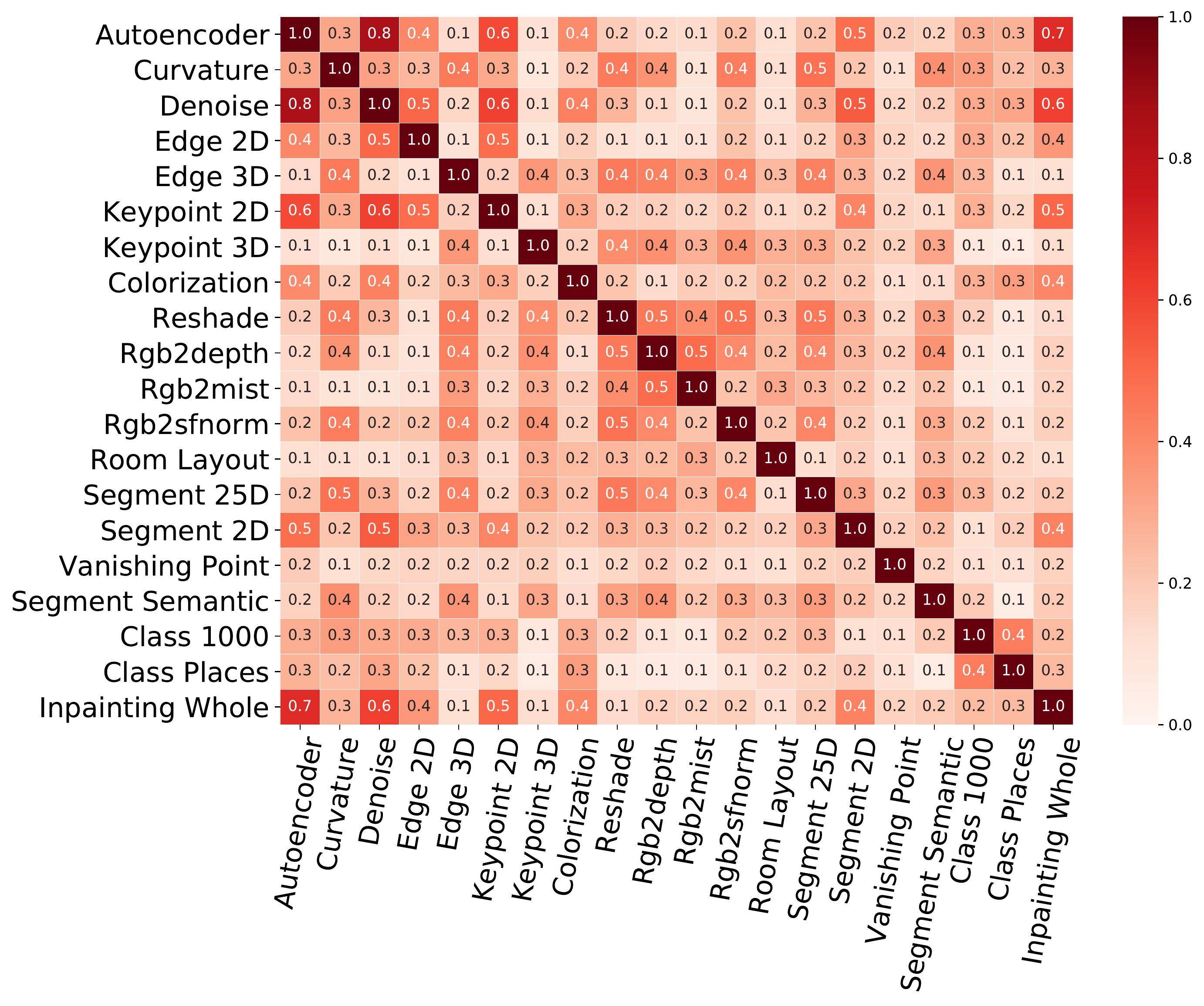}\\
\includegraphics[scale=0.25]{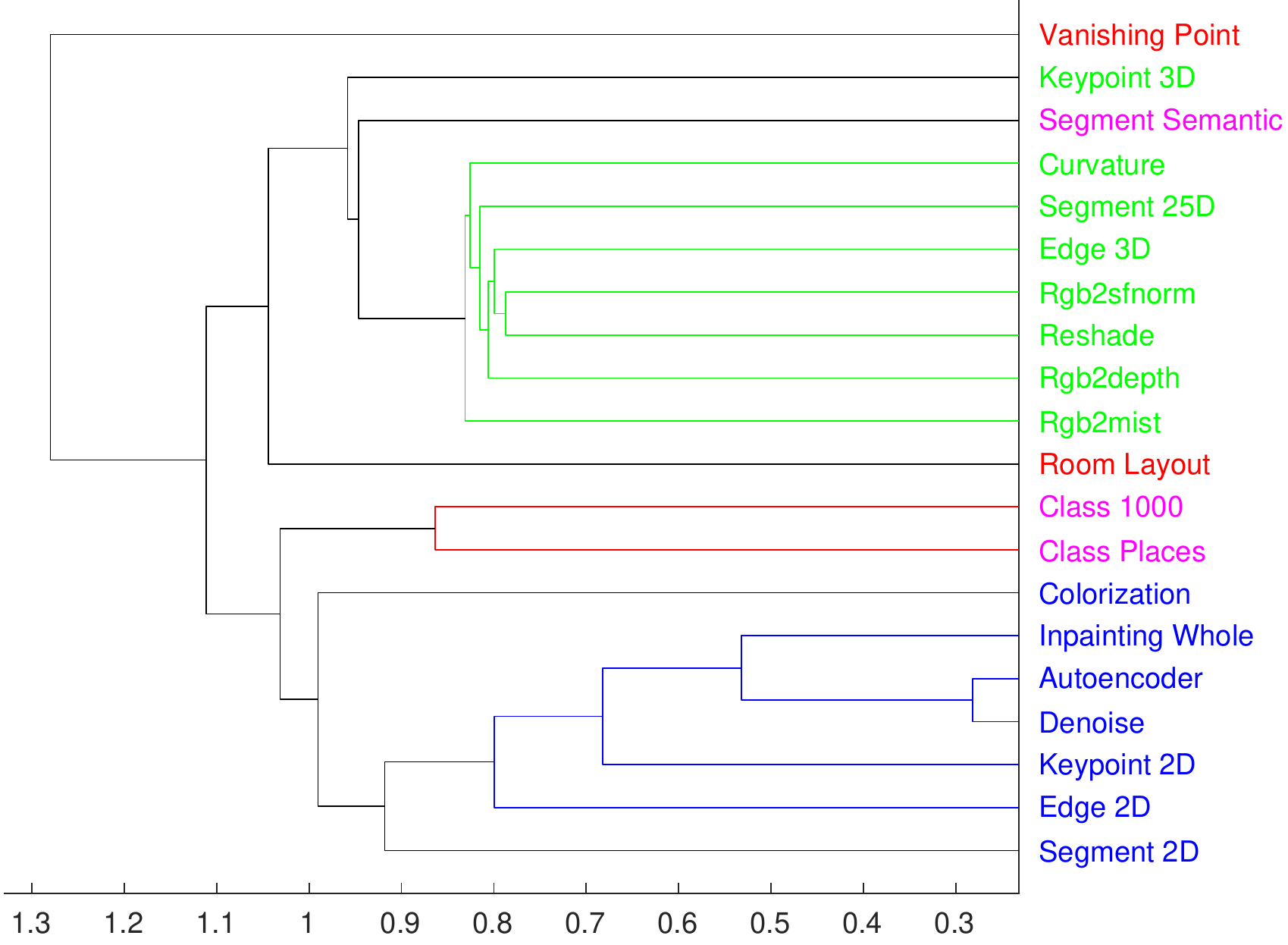}
\end{minipage}%
}%
\subfigure[800]{
\begin{minipage}[t]{0.30\linewidth}
\centering
\includegraphics[scale=0.16]{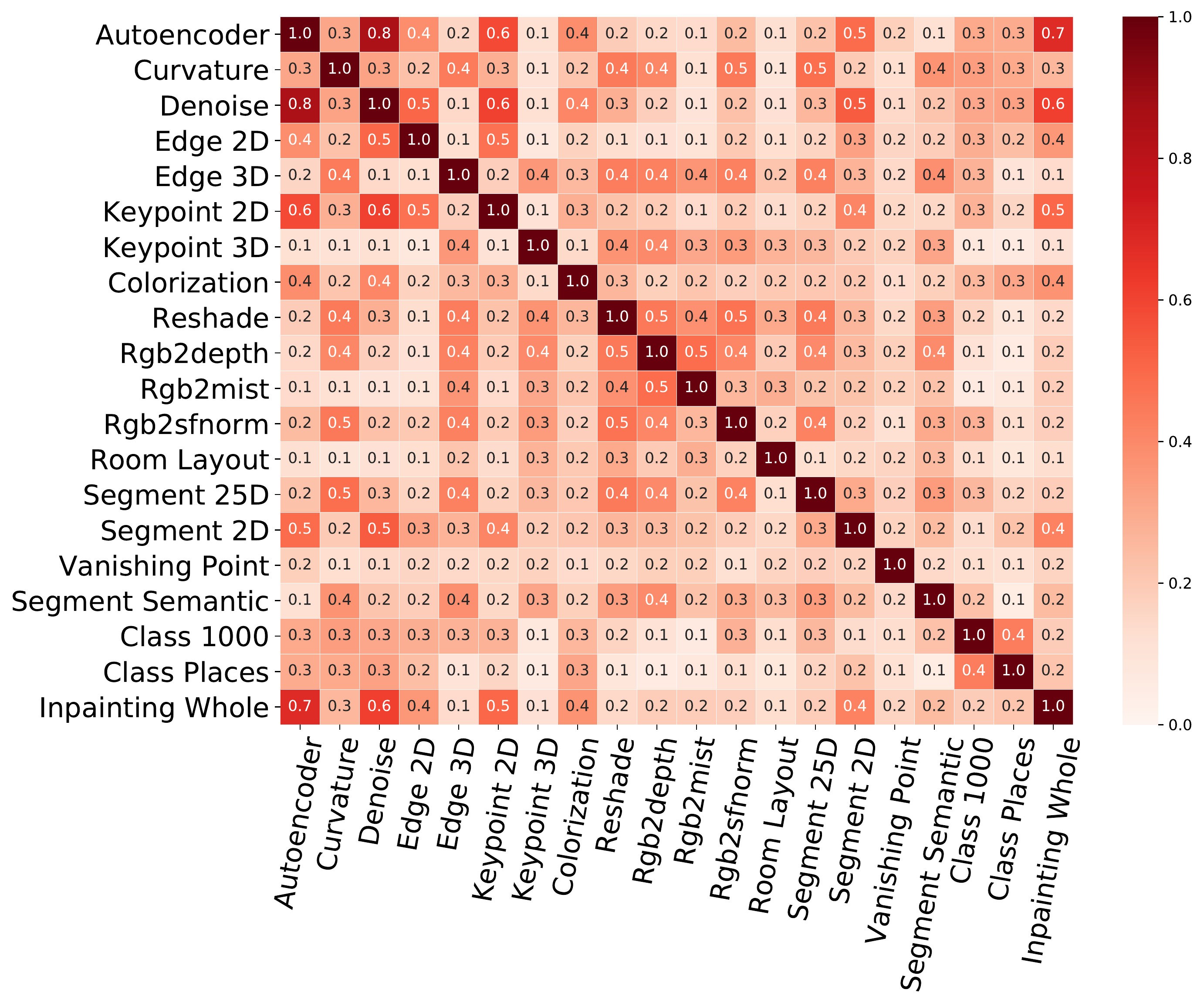}\\
\includegraphics[scale=0.25]{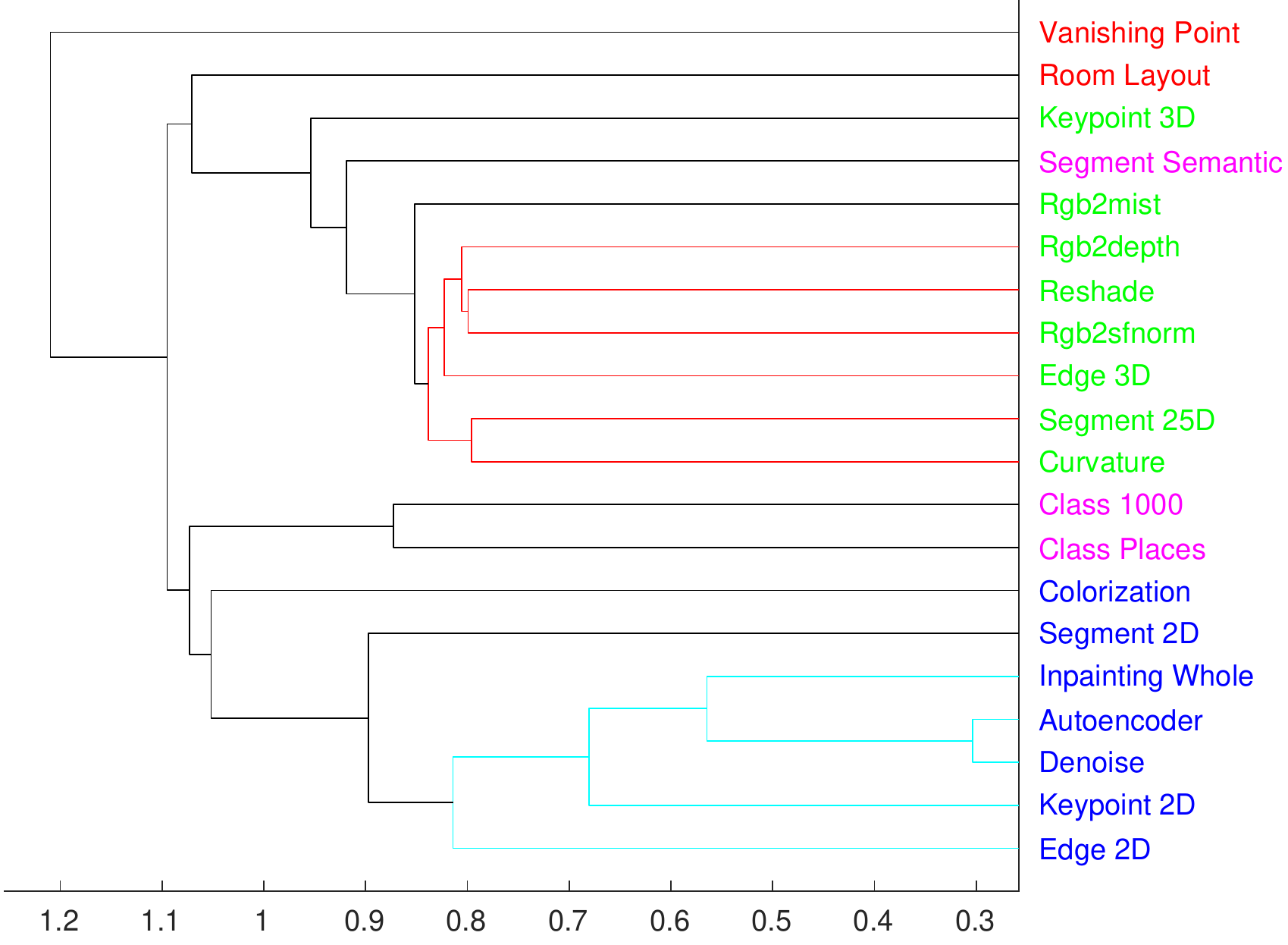}
\end{minipage}%
}%

\subfigure[1200]{
\begin{minipage}[t]{0.30\linewidth}
\centering
\includegraphics[scale=0.16]{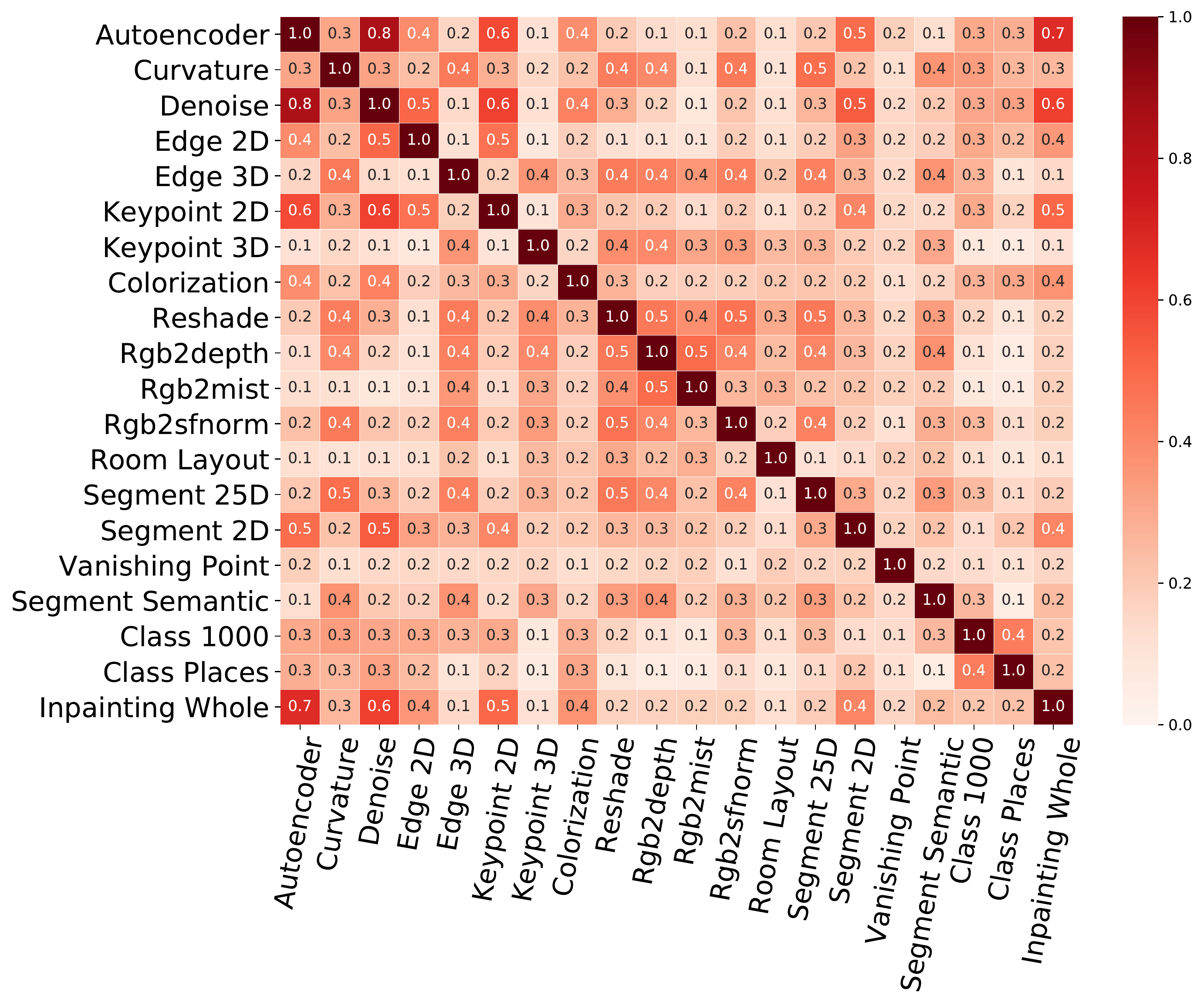}\\
\includegraphics[scale=0.25]{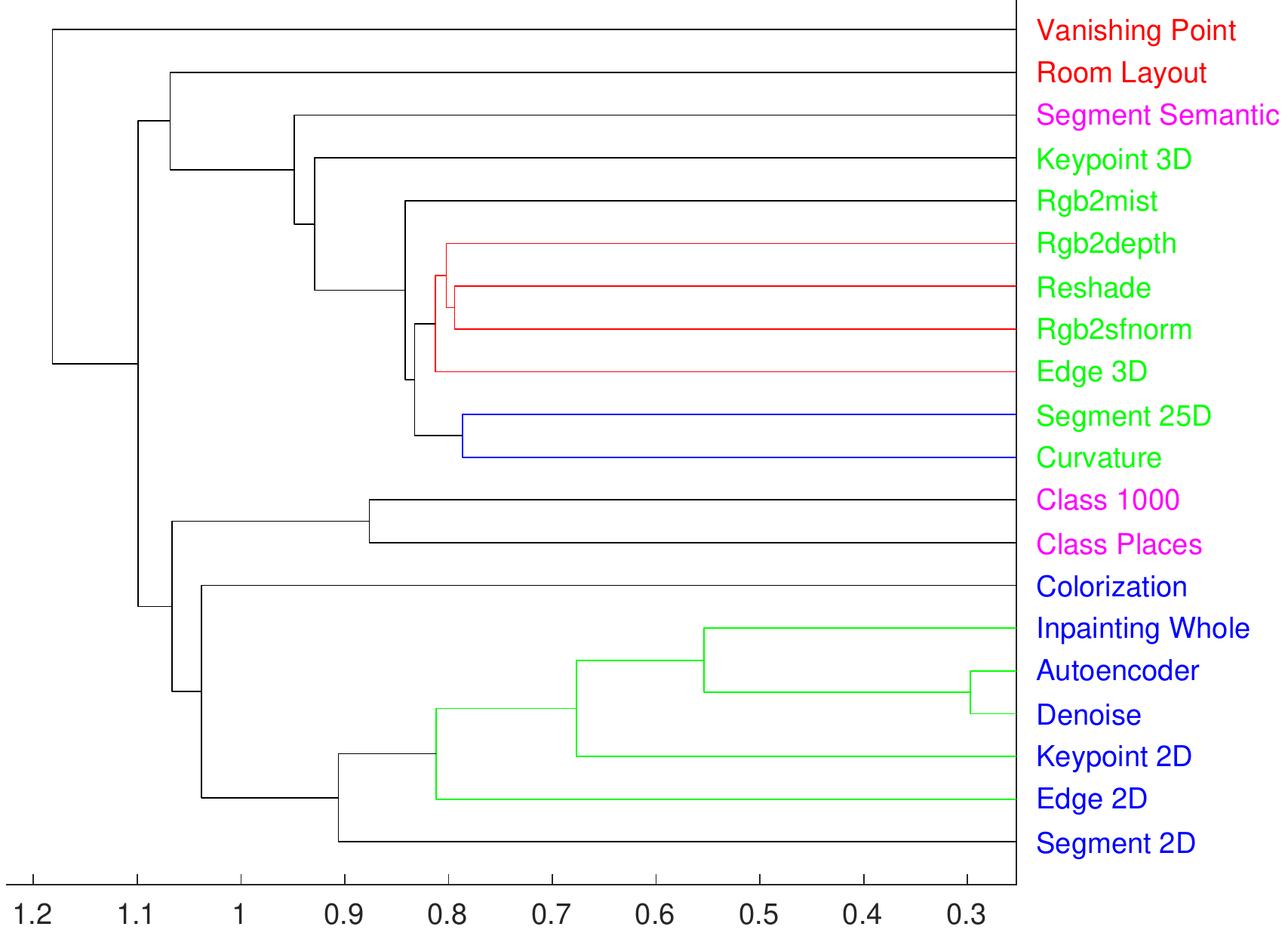}
\end{minipage}%
}%
\subfigure[1600]{
\begin{minipage}[t]{0.30\linewidth}
\centering
\includegraphics[scale=0.16]{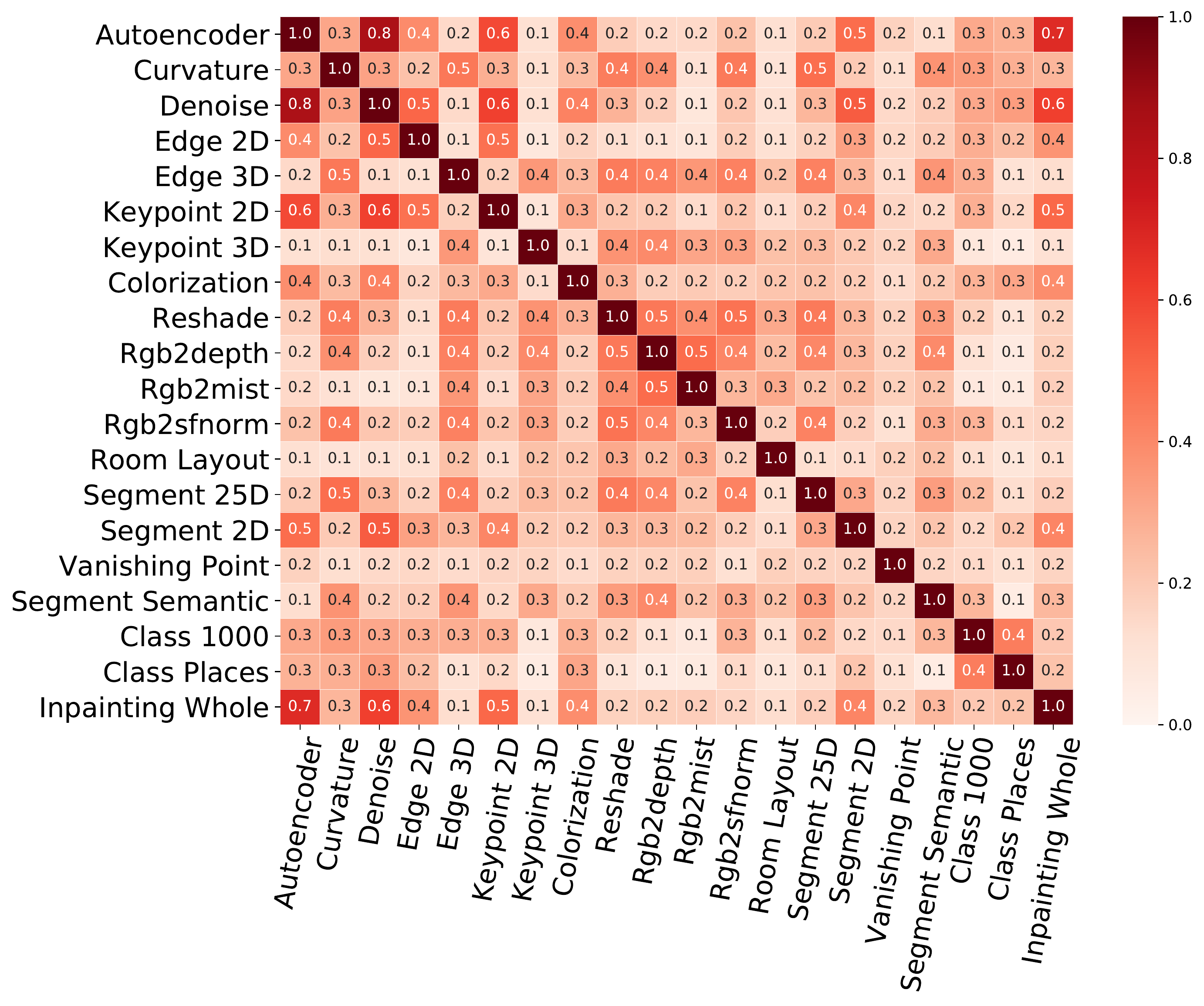}\\
\includegraphics[scale=0.25]{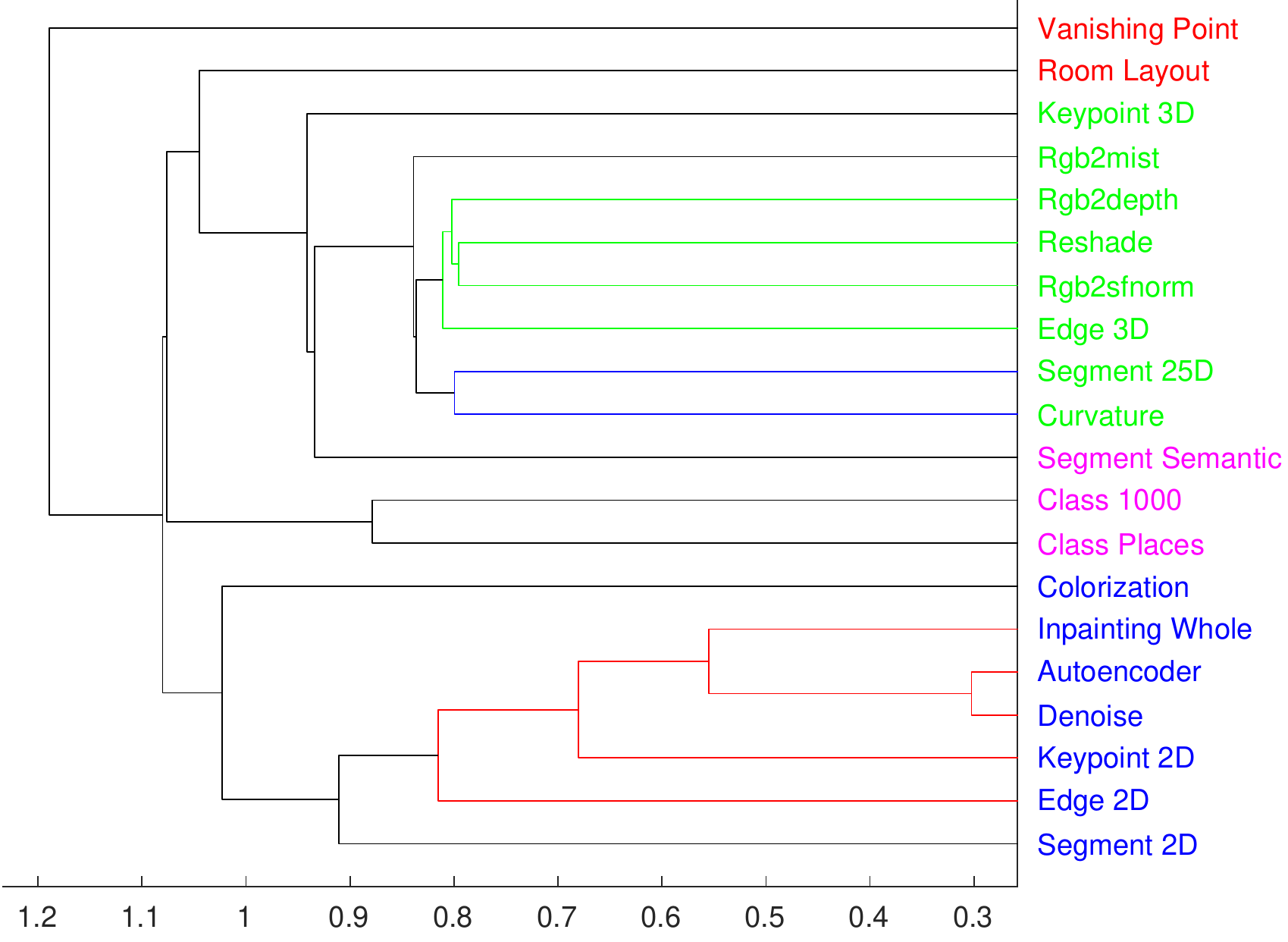}
\end{minipage}%
}%
\subfigure[2000]{
\begin{minipage}[t]{0.30\linewidth}
\centering
\includegraphics[scale=0.16]{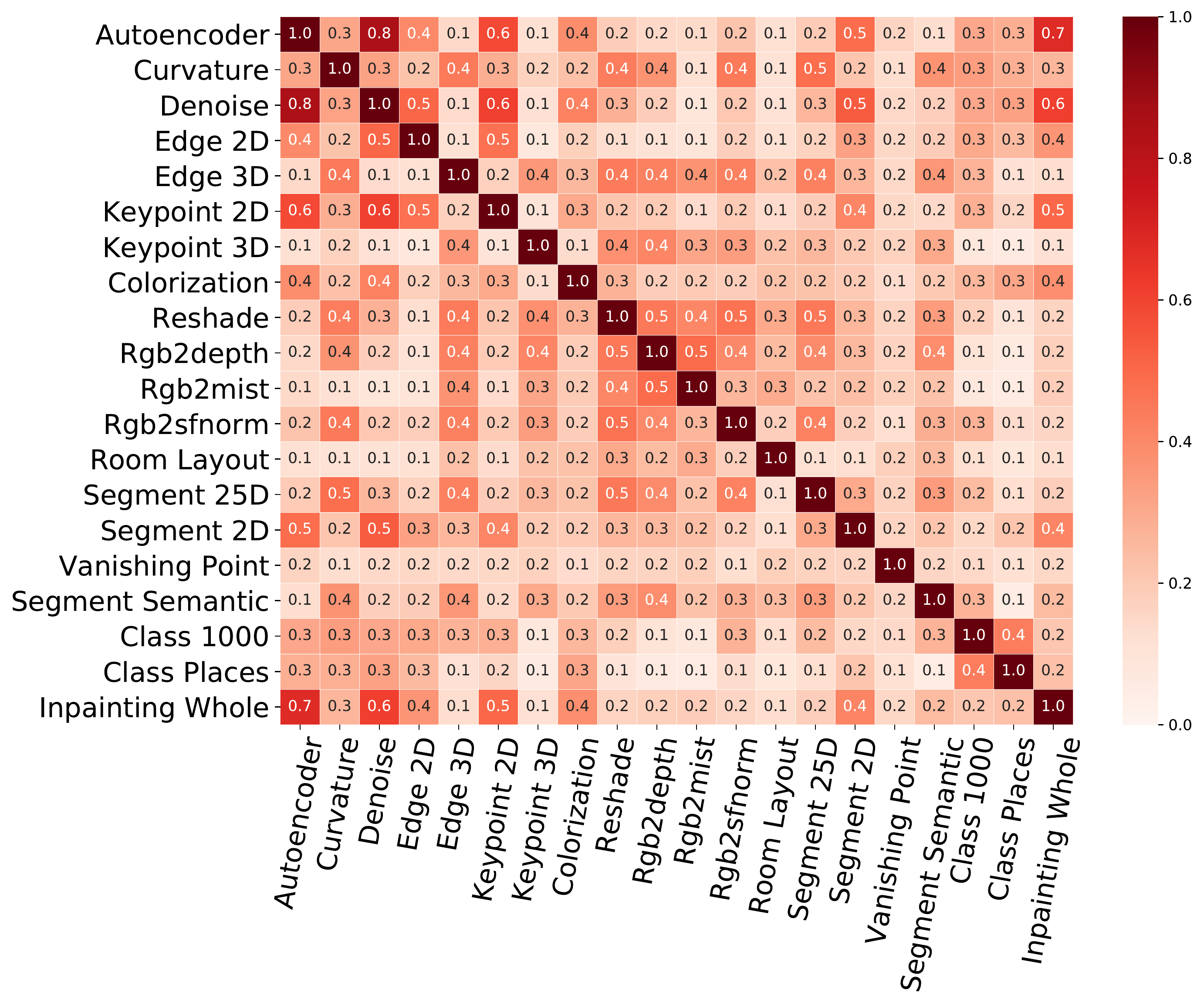}\\
\includegraphics[scale=0.25]{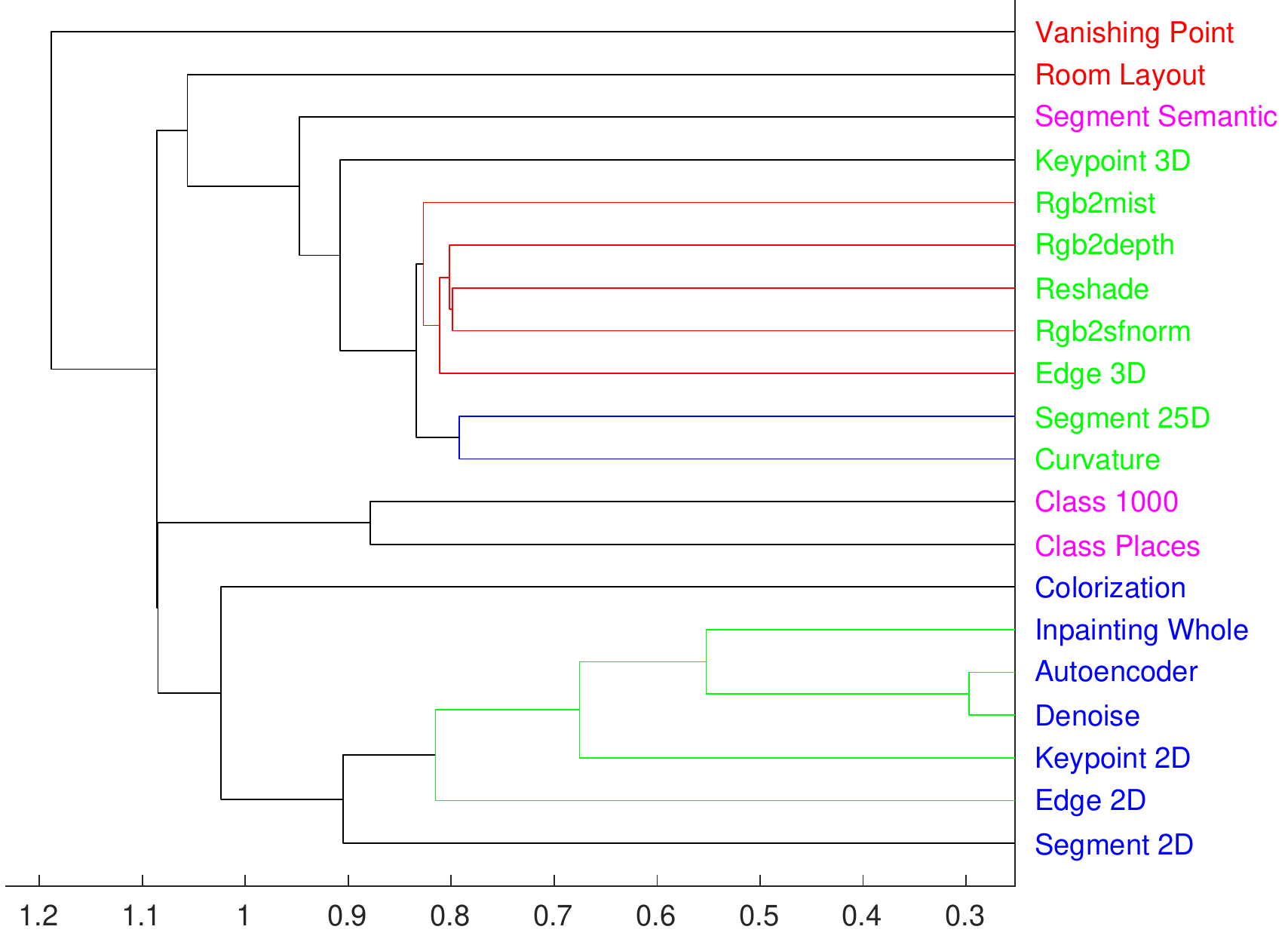}
\end{minipage}%
}%

\centering
\caption{Model transferability matrices and the task similar trees produced by our method with $\epsilon$-LRP on various size (listed below the subfigures) of probe data randomly selected from COCO. }
\label{fig:vis_increasing_data}
\end{figure}
We also investigate how the size of the probe data affects the results. Towards this end, we randomly sample $\{100, 400, 800, 1200, 1600, 2000\}$ images from taskonomy to produce the model transferability. Results are provided in Figure~\ref{fig:vis_increasing_data}. It can be seen that with the increasing probe data, the constructed relatedness (shown in task similar trees and the visualization of transferability matrices) keeps almost unchanged. This result indicates that the proposed method is also insensitive to the size of the probe data. A few hundreds of images are sufficient for the proposed method.

\begin{figure}[t]
  \centering
  \includegraphics[scale=0.25]{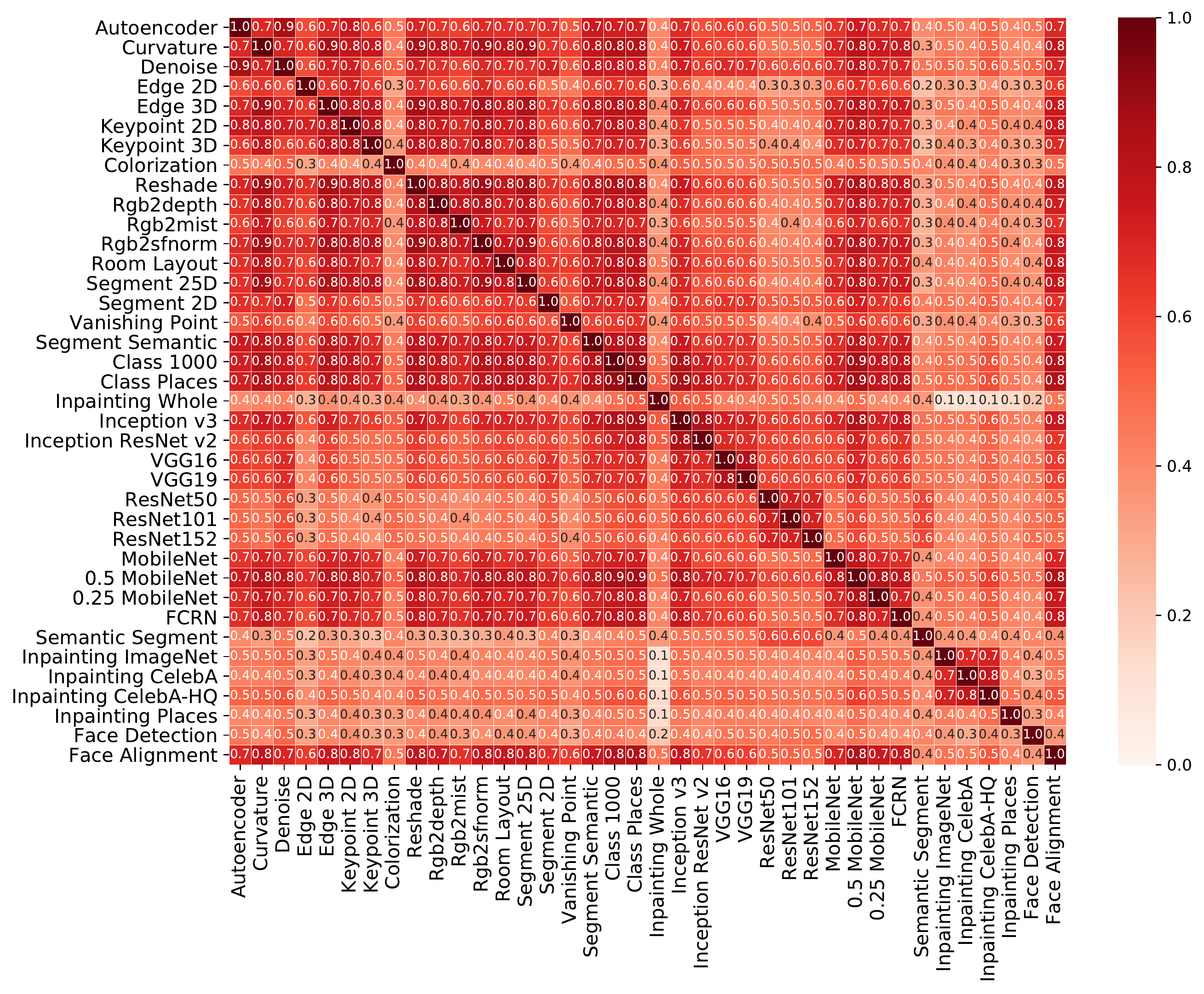}
  \includegraphics[scale=0.4]{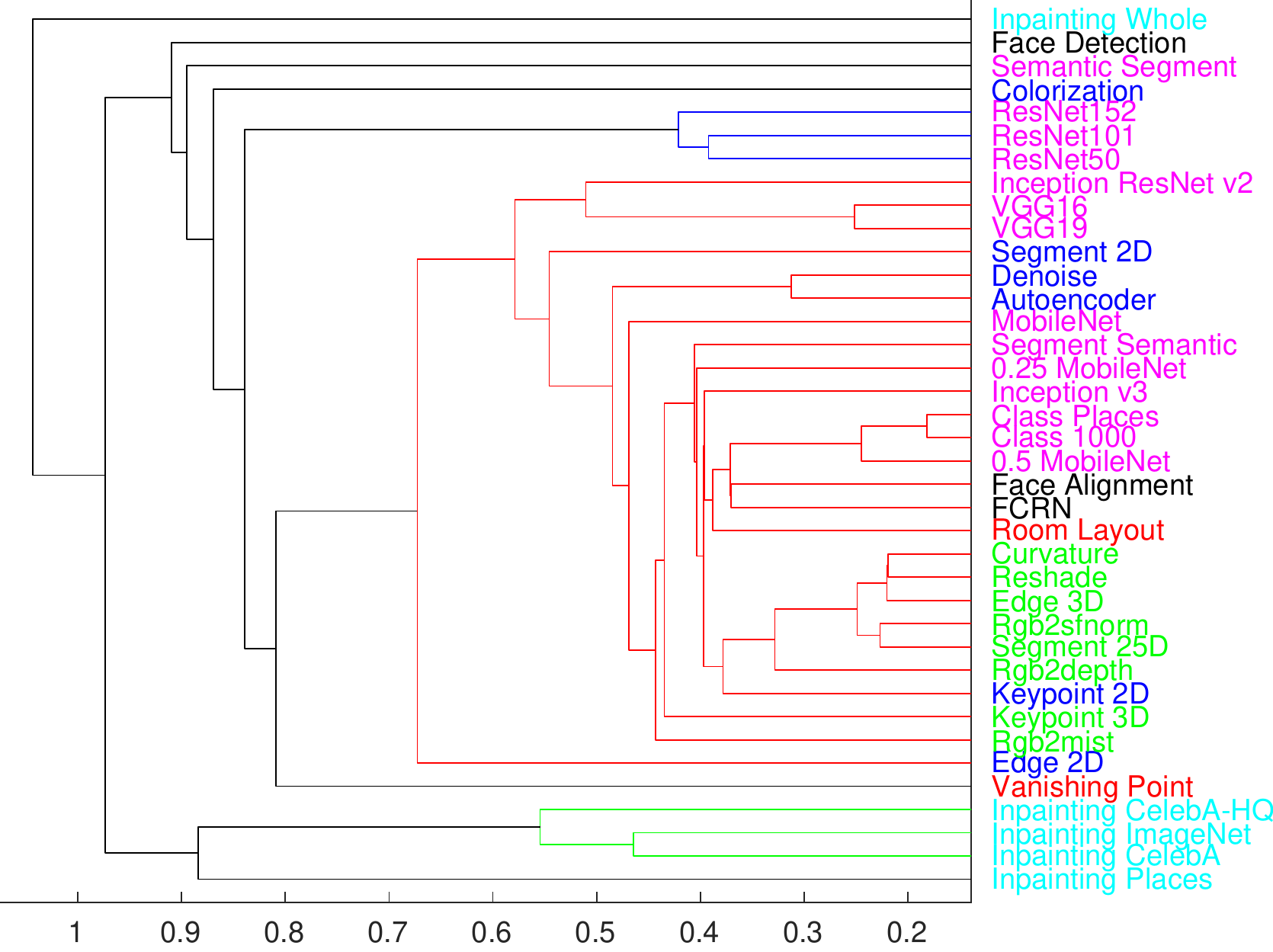}
  \caption{The model transferability matrix (left) and the task similar tree (right) produced by saliency.}
  \label{fig:all_saliency}
\end{figure}

\begin{figure}[t]
  \centering
  \includegraphics[scale=0.25]{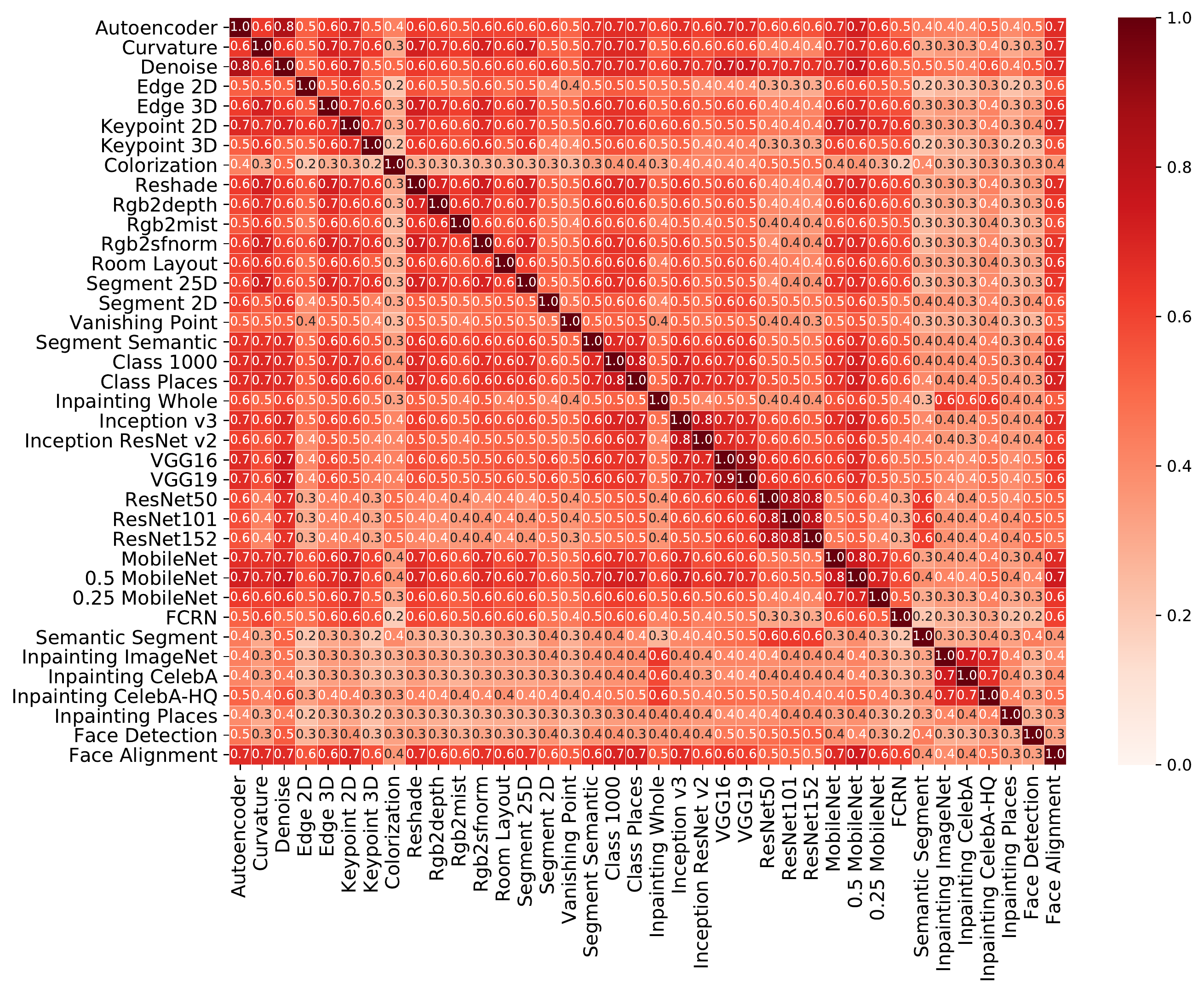}
  \includegraphics[scale=0.4]{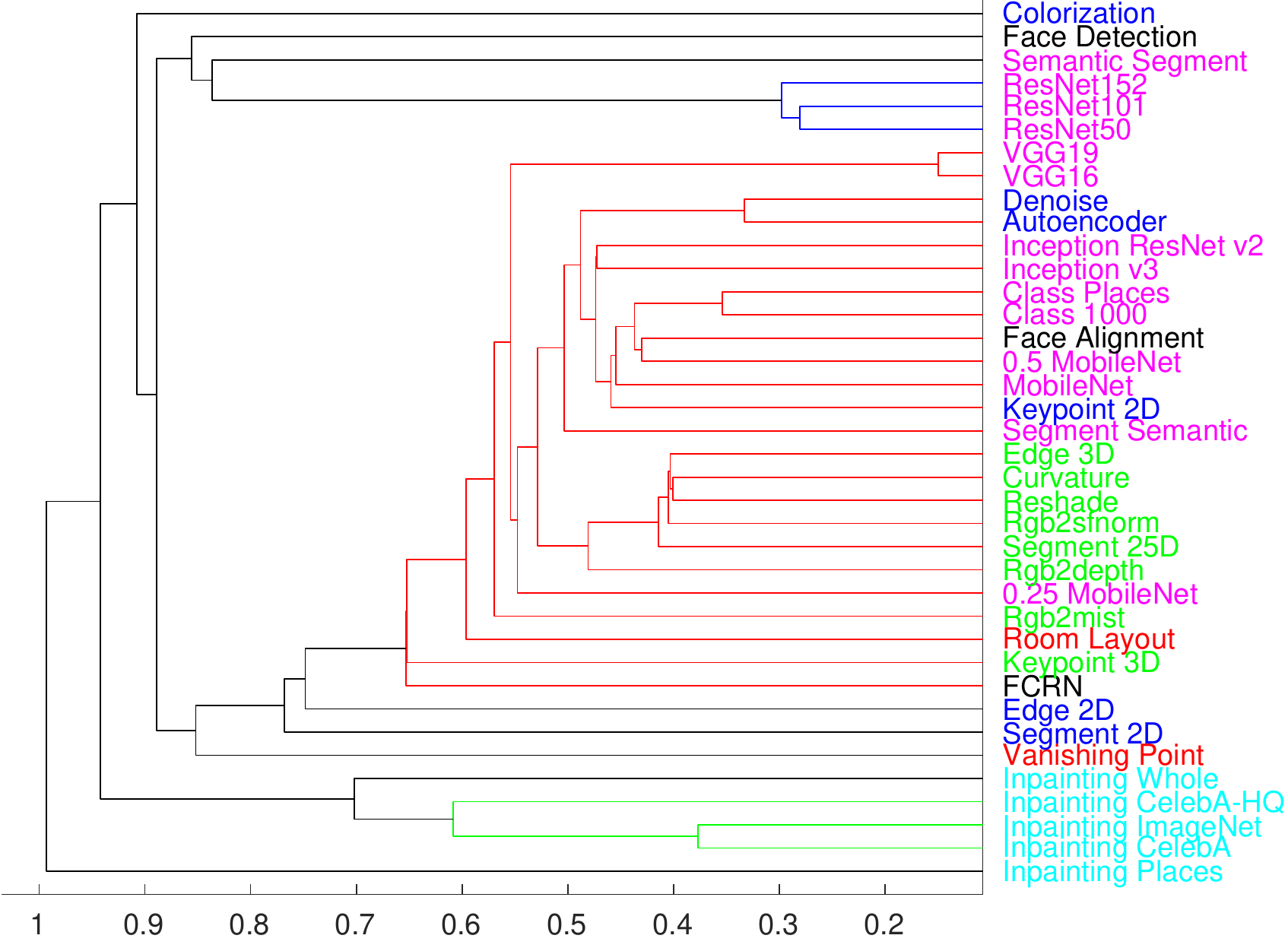}
  \caption{The model transferability matrix (left) and the task similar tree (right) produced by gradient*input.}
  \label{fig:all_gradXinput}
\end{figure}

\begin{figure}[t]
  \centering
  \includegraphics[scale=0.25]{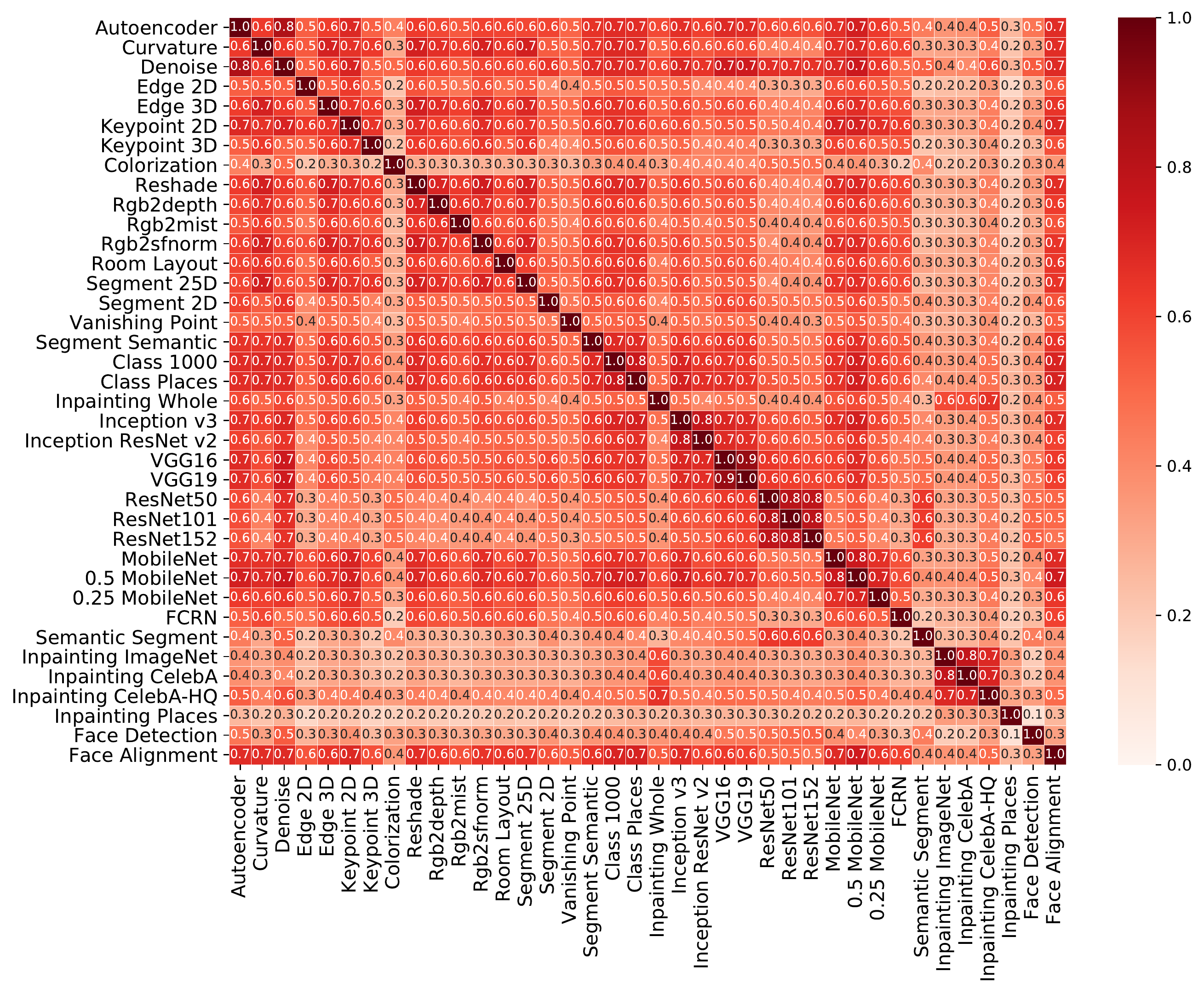}
  \includegraphics[scale=0.4]{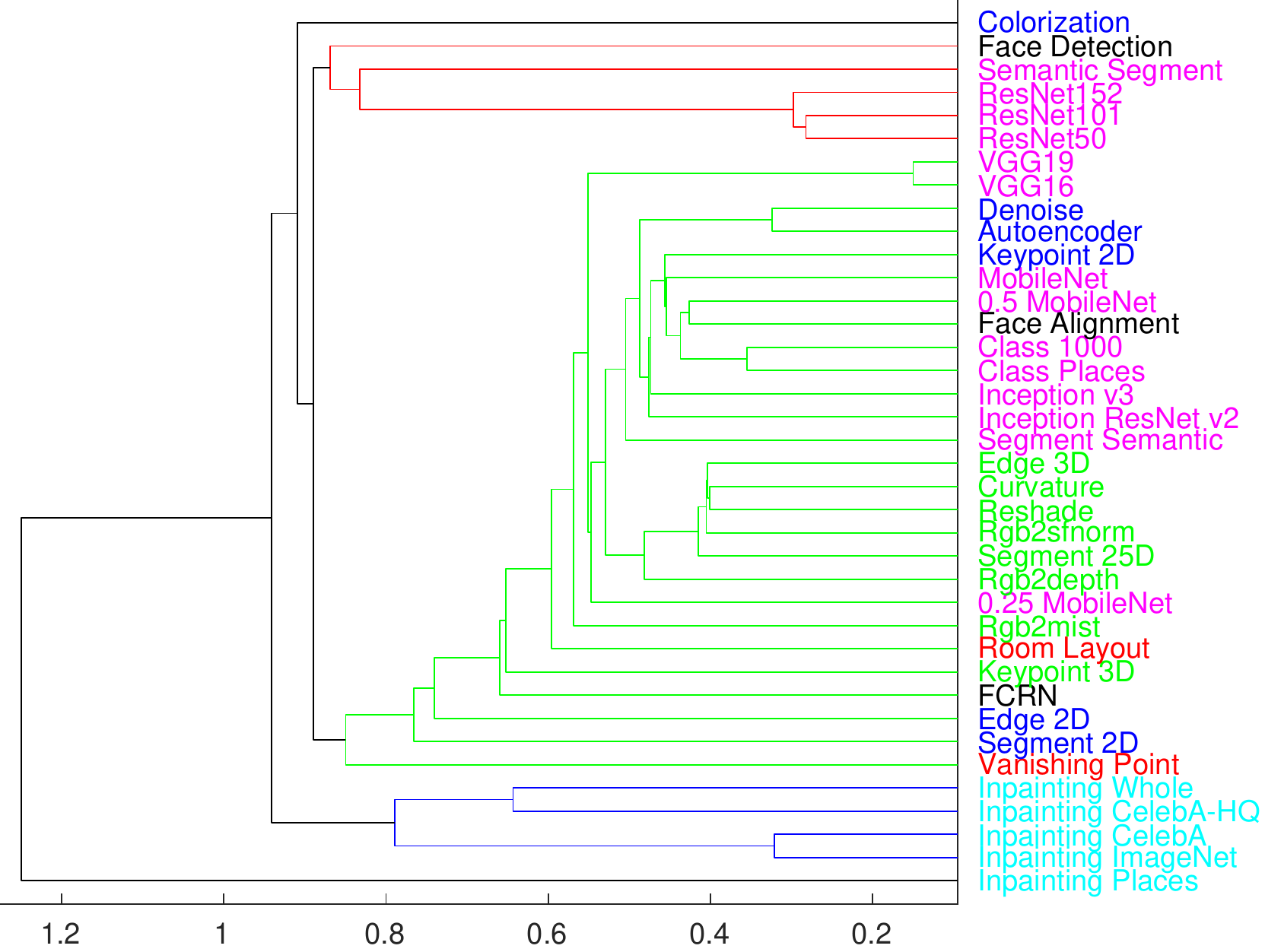}
  \caption{The model transferability matrix (left) and the task similar tree (right) produced by $\epsilon$-LRP.}
  \label{fig:all_elrp}
\end{figure}
\section{Experiments on the Merged Group}
To better understand our method, we also merge the two groups (taskonomy models and the pre-trained models collected outside taskonomy) to form a comprehensive group, which consists of totally 38 models. Experiments are conducted on the taskonomy data with saliency, gradient*input and $\epsilon$-LRP, of which results are shown in Figure~\ref{fig:all_saliency},~~\ref{fig:all_gradXinput} and~\ref{fig:all_elrp}, respectively. By comparing the results of the three attribution methods, we make the following three observations or analysis:
\begin{itemize}[leftmargin=15pt]
\item $\epsilon$-LRP and saliency produce highly similar results. However, the results of saliency are a little different. These results are consistent with results on taskonomy models or collected models alone, which implies that the proposed method can be scalable to model library of larger size.
\item The same-task models, although trained on data from different domains, tend to cluster together. These can be verified by the inpainting models which cluster together in our experiments. Furthermore, we additionally conduct an experiment (not shown in the figure) on another colorization model~\cite{nazeri2018image}. The model affinity obtained by our method ranks the taskonomy colorization model first among all others, which again verifies our conclusion.
\item In most cases, the global task structure (all models are considered) preserves the local task structure (only a fraction of models are considered). For example, when removing the taskonomy models from the task similar tree, the remaining tree structure is highly similar with that of collected models.
\end{itemize}
We argue all these observations are not trivial and providing us more insights into deep models and transfer learning.

{\small
\bibliographystyle{plain}
\bibliography{supp}
}